\pdfoutput=1
\documentclass[12pt]{utils/wlscirep}

\usepackage{utils/utils}

\title{Spectral symmetry fingerprints in equivariant representations}

\author[1,2,3]{Jiacheng Cen}
\author[1,2,3]{Anyi Li}
\author[1,2,3]{Ning Lin}
\author[1,2,3]{Zihe Wang}
\author[1,2,3]{Ji-Rong Wen}
\author[1,2,3,4, \Letter]{Wenbing~Huang}
\affil[1]{Gaoling School of Artificial Intelligence, Renmin University of China, Beijing, China}
\affil[2]{Beijing Key Laboratory of Research on Large Models and Intelligent Governance, Beijing, China}
\affil[3]{Engineering Research Center of Next-Generation Intelligent Search and Recommendation, MOE, Beijing, China}
\affil[4]{Beijing Academy of Artificial Intelligence, Beijing, China}
\affil[\Letter]{Correspondence should be addressed to: hwenbing@ruc.edu.cn}

\begin{document}

\begin{abstract}
Symmetry governs the organization of physical structures, while symmetry breaking reveals their transformations. In geometric deep learning, symmetry imposes rank constraints on equivariant representations, limiting the expressive capacity of equivariant graph neural networks (GNNs). Here we show that these constraints can instead be exploited: the singular-value spectra of equivariant representations provide a unified fingerprint of structural symmetry. We introduce spectral symmetry fingerprints (SymFP), which enable discrete point-group identification through rank-template matching and continuous symmetry analysis by quantifying spectral energy beyond target rank constraints. SymFP is orientation invariant and architecture agnostic, enabling a parameter-free GNN implementation. Its spectra can be reused across multiple target groups, and the resulting leakage scores satisfy subgroup monotonicity. Across synthetic structures, molecules, coordination complexes and large biological assemblies, SymFP accurately identifies point groups, tracks controlled symmetry breaking and reveals local and degree-resolved symmetry changes that can be obscured by global geometric measures. These results establish the spectral structure of equivariant representations as a general framework for characterizing symmetry across physical systems.
\end{abstract}

\keywords{}

\settitle

\setcounter{secnumdepth}{3}
\begin{bibunit}[unsrtnat]

\section{Introduction}

Symmetry, defined as the invariance of a structure under specific transformations, provides a fundamental principle for understanding structural organization in the physical world. In chemistry and structural biology, characterizing symmetry offers important insights into structural states, functional properties, and state transformations. Real-world structures, however, often exhibit only approximate symmetry due to thermal fluctuations, structural distortions, and external perturbations. Structural symmetry characterization therefore involves two complementary problems: discrete point-group identification, which assigns exactly symmetric structures to predefined point groups, and continuous symmetry analysis, which quantifies the extent to which a structure deviates from a target symmetry \citep{zabrodsky1992continuous,zabrodsky1993continuous,nielsen2024evaluation,nielsen2025determining,engel2021point}.

Equivariant representations offer a natural starting point for addressing these two tasks. These representations transform consistently with their inputs under rotations and reflections \citep{bronstein2021geometric} and form the basis of equivariant graph neural networks (GNNs), which have achieved considerable success across scientific applications \citep{jumper2021highly,baek2021accurate,zeni2025generative}. Specifically, equivariant representations of degree $l$ are collected into a matrix $\smVl\in\mathbb{R}^{(2l+1)\times C}$, whose $C$ feature channels transform under the same degree-$l$ representation of rotations and reflections. For symmetric structures, however, equivariance restricts these feature channels to subspaces fixed by the input point group, leading to a reduction in expressive power \citep{smidt2021finding,kaba2023symmetry,joshi2023expressive,cen2024are,lin2026reducing}. The dimensions of these subspaces determine the rank constraints on $\smVl$, which vary across point groups and degrees \citep{cen2024are,lin2026reducing}. Here we turn these rank constraints into a tool for characterizing symmetry.

To this end, we construct spectral symmetry fingerprints (SymFP) from the singular-value spectra of equivariant representations across degrees, providing a common basis for discrete point-group identification and continuous symmetry analysis. For the former, we define a rank template for each point group based on its fixed-subspace dimensions across degrees. The numerical rank sequence of an input structure is obtained from SymFP by counting singular values above a numerical threshold at each degree. Matching this numerical rank sequence against the candidate rank templates allows us to identify the point group. Degrees $l=0,\ldots,9$ are sufficient to distinguish the rank templates of all 32 crystallographic point groups. For the latter, deviations from a target symmetry can cause singular values that would otherwise be zero to become nonzero. To quantify these deviations, we introduce leakage scores (LS), which measure the fraction of spectral energy associated with singular values beyond the target fixed-subspace dimension at each degree. Degree-wise leakage scores further reveal how symmetry deviations are distributed across degrees. By averaging LS across selected degrees and applying a regularized negative-log transformation, we obtain pLS, with higher values indicating stronger agreement with the target rank constraints. As SymFP relies only on the algebraic structure of equivariant representations, it can be applied across different equivariant GNN architectures. To facilitate reuse and evaluation, we also develop a parameter-free (PF) equivariant GNN that generates representations directly from input structures using fixed radial bases and spherical harmonics.

By comparison, traditional approaches to structural symmetry characterization primarily rely on geometric distances or symmetry operations. Continuous symmetry measures (CSMs) quantify deviations from symmetry through the distance to the nearest symmetric configuration \citep{zabrodsky1992continuous,zabrodsky1993continuous}. The continuous symmetry operation measure (CSoM) evaluates deviations under individual symmetry operations and combines them into a continuous score \citep{nielsen2024evaluation,nielsen2025determining}. Engel's method, here termed Engel score, uses spherical harmonics to represent structures and searches over target-group orientations to maximize the energy retained by projection onto the corresponding fixed subspaces \citep{engel2021point,mbah2023early}. In contrast, SymFP characterizes structural symmetry through the intrinsic spectral structure of equivariant representations. Its construction is independent of any target point group, which enters subsequent analysis only through its fixed-subspace dimensions. This avoids constructing symmetric reference structures or searching over target orientations. Moreover, the same spectra can be reused across different target groups without recomputing the representations or their singular values, enabling efficient analysis of multiple target symmetries. Beyond computational efficiency, SymFP also satisfies subgroup monotonicity: for a given structure, relaxing the symmetry requirements from a group to one of its subgroups cannot increase the leakage score. This property supports consistent analysis of symmetry breaking along group--subgroup hierarchies. We further connect CSoM-style operation-wise deviations with Engel's retained projection energy in representation space and establish their relationships with the spectral energy underlying SymFP through variational analysis.

We evaluate SymFP on synthetic structures and real-world systems across scales. Experiments on synthetic symmetric structures show that numerical rank sequences match the corresponding rank templates and demonstrate accurate discrete point-group identification across GNN architectures and training states, even with the parameter-free GNN. Controlled deformations further show that SymFP tracks symmetry breaking and distinguishes deformation modes even at matched displacement amplitudes. We then extend the analysis to three classes of real-world systems. In aromatic molecules, torsion scans show that SymFP retrieves energy extrema from geometric symmetry, outperforming CSoM and Engel score in overall retrieval performance, while pretraining further enhances spectral contrasts associated with target symmetries. In DOTA complexes, fixed-subspace dimensions correspond to crystal-field parameter counts, and spectral responses to capping-water rotation exhibit periodicity consistent with electronic-structure calculations. For Dy-DOTA-water, SymFP evaluates seven target groups approximately two orders of magnitude faster than the fastest tested Engel implementation and four orders of magnitude faster than CSoM. Finally, across four large biological assemblies, SymFP reveals opposing local symmetry changes masked by whole-assembly scores, regional differences during conformational transitions, spatial variations in symmetry and the effects of missing assembly units. Together, these results show how SymFP brings discrete point-group identification and continuous symmetry analysis into a common framework, revealing structural symmetry across degrees and spatial regions.
\section{Results}
\label{sec:results}

\subsection{Spectral symmetry fingerprints}
\label{sec:exp-symfp}

\begin{figure}[!t]
\centering
\includegraphics[width=\textwidth]{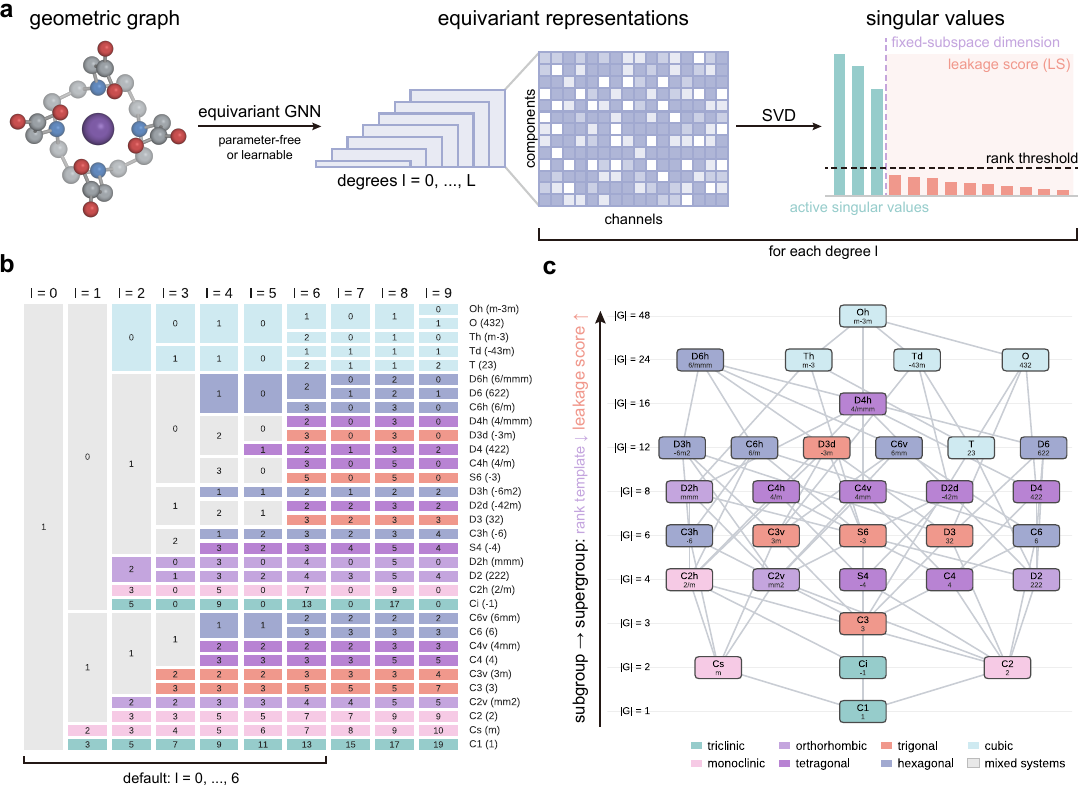}
\caption[Overview of SymFP.]{
\figtitle{Overview of SymFP.}
\textbf{a}, Pipeline for constructing SymFP from a geometric graph. The spectra support discrete point-group identification through rank-template matching and continuous symmetry analysis through leakage scores.
\textbf{b}, Rank templates of the 32 crystallographic point groups.
\textbf{c}, Subgroup relations and monotonicity of rank templates and leakage scores. In \textbf{b,c}, fill colour denotes the crystal system, with grey for blocks spanning multiple systems.
}
\label{fig:pipeline}
\end{figure}

We construct spectral symmetry fingerprints (SymFP) from the singular-value spectra of degree-wise representations produced by an equivariant GNN. Specifically, let $f^{(l)}$ denote the degree-$l$ output map of an equivariant GNN \citep{han2025survey}. For a geometric graph $\gG$ (\emph{e.g.}, a 3D molecule or protein structure), the $C$ representation channels produced at each degree $l=0,\ldots,L$ are collected as
$f^{(l)}(\gG)=\smVlgG=[\svv^{(l)}_1(\gG),\ldots,\svv^{(l)}_C(\gG)]\in\mathbb{R}^{(2l+1)\times C}$,
where each $\svv^{(l)}_c(\gG)\in\mathbb{R}^{2l+1}$ denotes the degree-$l$ representation vector of channel $c$ (\cref{fig:pipeline}a). By default, we use the same channel count $C>2L+1$ across all degrees. In addition to being permutation-invariant with respect to nodes and edges, these representations are equivariant to transformations in $\mathrm{O}(3)$:
\begin{equation}
\label{eq:results-equivariance}
    \smVl(\Gh\cdot\gG)=\rho^{(l)}(\Gh)\smVlgG,\qquad \forall\,\Gh\in\mathrm{O}(3),
\end{equation}
where $\Gh\in\mathrm{O}(3)$ denotes a rotation or reflection, and $\rho^{(l)}$ is the corresponding degree-$l$ representation \citep{geiger2022e3nn}. We apply singular value decomposition (SVD) to each degree-$l$ representation $\smVlgG$ and collect the ordered singular values $\sigma_1^{(l)}\geq\cdots\geq\sigma_{2l+1}^{(l)}\geq0$ across degrees to define the spectral symmetry fingerprint (SymFP):
\begin{equation}
\label{eq:results-fingerprint}
    \mathrm{SymFP}(\gG)
    \coloneqq(\sigma_i^{(l)})_{0\leq l\leq L,\ 1\leq i\leq 2l+1}.
\end{equation}
By construction, SymFP preserves $\mathrm{O}(3)$-invariance, enabling symmetry analysis without explicit orientation optimization or search. We consider two complementary tasks: discrete point-group identification for exactly symmetric structures and continuous symmetry analysis to quantify departures from target symmetry.

\subsubsection*{Discrete point-group identification}

If a graph $\gG$ is symmetric under a group $\GH$, i.e., $\Gh\cdot\gG=\gG$ for every $\Gh\in\GH$, equivariance constrains its degree-$l$ representations to the corresponding fixed subspace of dimension $m_l(\GH)$, implying $\operatorname{rank}(\smVlgG)\leq m_l(\GH)$. The ranks of equivariant representations across degrees define the rank sequence:
\begin{equation}
\label{eq:results-rank-sequence-template}
    \mathrm{RankSeq}(\gG)
    \coloneqq \bigl(\operatorname{rank}(\smVl(\gG))\bigr)_{l=0}^{L}
    \leq \bigl(m_l(\GH)\bigr)_{l=0}^{L}
    \eqqcolon \mathrm{RankTemplate}(\GH).
\end{equation}
{\looseness=-1
Here, $\leq$ is understood componentwise, and $\mathrm{RankTemplate}(\GH)$ denotes the rank template associated with $\GH$ (\cref{fig:pipeline}b). In practice, we estimate $\mathrm{RankSeq}(\gG)$ by counting the singular values in SymFP above an appropriate numerical threshold, obtaining a numerical rank sequence (\cref{eq:numerical-rank-sequence}). When the representations $\smVlgG$ span the fixed subspaces, $\mathrm{RankSeq}(\gG)=\mathrm{RankTemplate}(\GH)$, so a unique match among the candidate rank templates then identifies the point group directly (\cref{eq:rank-template-matching}). Degrees $l=0,\ldots,6$ already yield 30 distinct rank templates across the 32 crystallographic point groups, with only $\mathrm{D}_6/\mathrm{D}_{6h}$ and $\mathrm{O}/\mathrm{O}_h$ sharing the same template. Extending the sequence to $l=0,\ldots,9$ distinguishes all 32 groups (\cref{fig:pipeline}b; \cref{cor:crystallographic-32}). Empirically, we find that the representations span the full fixed subspaces in most tested cases (\cref{tab:synthetic-training-regimes}). Moreover, for a fixed input and under suitable regularity conditions, our theoretical analysis shows that if equality in \cref{eq:results-rank-sequence-template} holds at one parameter setting, it holds for almost every parameter choice.\par}

\subsubsection*{Continuous symmetry analysis}

When a graph $\gG$ departs from exact symmetry, its degree-wise representations may violate the corresponding rank-template constraints. Specifically, singular directions beyond the fixed-subspace dimension $m_l(\GH)$ can become active, resulting in $\sigma_i^{(l)}>0$ for some $i>m_l(\GH)$. The leakage score (LS) with respect to $\GH$ quantifies the fraction of spectral energy contained in these spectral tail directions:
\begin{equation}
\label{eq:results-leakage}
    \mathrm{LS}^{(l)}_{\GH}(\gG)
    =\frac{1}{Z_l(\gG)}\sum_{i>m_l(\GH)}(\sigma_i^{(l)})^2,
    \qquad
    Z_l(\gG)=\sum_i(\sigma_i^{(l)})^2+\varepsilon_{\mathrm{norm}}.
\end{equation}
The numerator measures the spectral tail energy beyond the target fixed-subspace dimension. By default, $Z_l(\gG)$ normalizes this quantity by the total spectral energy at degree $l$, enabling comparison across degrees. Setting $Z_l(\gG)=1$ instead recovers the unnormalized spectral tail energy. To aggregate responses over a selected set $\Lambda\subseteq\{0,\ldots,L\}$, we define the pooled leakage score and its negative-log transformation (pLS):
\begin{equation}
\label{eq:results-pooled-leakage}
    \mathrm{LS}_{\Lambda,\GH}(\gG)
    =\frac{1}{|\Lambda|}\sum_{l\in\Lambda}\mathrm{LS}^{(l)}_{\GH}(\gG),
    \qquad
    \mathrm{pLS}_{\Lambda,\GH}(\gG)
    =-\log_{10}\!\left(\mathrm{LS}_{\Lambda,\GH}(\gG)+\varepsilon_{\log}\right).
\end{equation}
We set $\varepsilon_{\mathrm{norm}}=\varepsilon_{\log}=10^{-10}$. The logarithmic transformation improves sensitivity to small leakage values and yields a maximum pLS value of $10$ when $\mathrm{LS}_{\Lambda,\GH}(\gG)=0$. Therefore, lower LS and higher pLS indicate stronger agreement with the rank-template constraints imposed by the target symmetry.

These target-dependent scores are derived from the same SymFP, which is intrinsic to the equivariant representation and constructed independently of any target point group. The target group affects the leakage score only through the fixed-subspace dimensions $m_l(\GH)$, which determine the spectral tail at each degree. Consequently, a single SymFP supports symmetry analysis across multiple target groups without recomputing the underlying representations or singular values.

\subsubsection*{Subgroup monotonicity}

The SymFP framework also satisfies subgroup monotonicity in both tasks (\cref{fig:pipeline}c). A subgroup admits a fixed subspace of equal or larger dimension, yielding a componentwise equal or larger rank template. For the same input, degree set and pooling weights, it therefore yields an equal or lower leakage score and an equal or higher negative-log leakage score. Thus, whenever $\GH\leq\GH'$, the following three monotonicity relations hold:
\begin{equation}
\label{eq:results-subgroup-monotonicity}
    \mathrm{RankTemplate}(\GH)\geq\mathrm{RankTemplate}(\GH'),
    \quad
    \mathrm{LS}_{\Lambda,\GH}(\gG)\leq\mathrm{LS}_{\Lambda,\GH'}(\gG),
    \quad
    \mathrm{pLS}_{\Lambda,\GH}(\gG)\geq\mathrm{pLS}_{\Lambda,\GH'}(\gG),
\end{equation}
where the rank-template inequality is understood componentwise.

\subsubsection*{Parameter-free equivariant GNN}

SymFP relies only on the algebraic structure of equivariant representations and is therefore applicable to any equivariant GNN architecture. To facilitate reuse and evaluation, we provide a parameter-free (PF) equivariant GNN (\cref{alg:pf}). It combines node and edge attributes, fixed radial bases and real spherical harmonics into node, edge and hybrid representations.

\begin{algorithm}[!ht]
\DontPrintSemicolon
\caption{PF Construction at Fixed Degree $l$}
\label{alg:pf}
\KwData{Geometric graph $\gG(\mH,\gmX;\mA)=\gG(\gV;\gE)$ with node attributes $\vh_i\in\sR^{d_v}$, node coordinates $\gvx_i\in\sR^3$ and edge attributes $\ve_{ij}\in\sR^{d_e}$ on $\gE$. Predefined RBFs $\{\rho_n\}_{n=1}^{N_{\gV}}$ and $\{\eta_m\}_{m=1}^{N_{\gE}}$, and radial normalization scales $Z_{\gV}$ and $Z_{\gE}$.}
\KwResult{Node block $\smVl_{\gV}\in\sR^{(2l+1)\times C_{\gV}}$, edge block $\smVl_{\gE}\in\sR^{(2l+1)\times C_{\gE}}$, and hybrid block $\smVl\in\sR^{(2l+1)\times (C_{\gV}+C_{\gE})}$.}

\BlankLine
\tcp*[l]{----- Node Representations -----}
$C_{\gV} = d_vN_{\gV}$\;
\For{$i\in\gV$}{
    $\va_i = [\vh_{ia}\rho_n(\|\gvx_{ic}\|/(Z_{\gV}+\varepsilon_0))]_{(a,n)} \in \sR^{C_{\gV}}$\tcp*[r]{\small\frenchspacing $Z_{\gV}$: mean radius (default)}
}
$\smVl_{\gV} = \sum_{i\in\gV}Y^{(l)}(\hat{\vx}_{ic})\otimes \va_i \in \sR^{(2l+1)\times C_{\gV}}$\tcp*[r]{\small\frenchspacing $Y^{(l)}$: real spherical harmonics}

\BlankLine
\tcp*[l]{----- Edge Representations -----}
$C_{\gE} = (2d_v+d_e)N_{\gE}$\;
\For{$(i,j)\in\gE$}{
    $\vq_{ij} = [\vh_i,\vh_j,\ve_{ij}] \in \sR^{2d_v+d_e}$\;
    $\vb_{ij} = [\vq_{ija}\eta_m(\|\gvx_{ij}\|/(Z_{\gE}+\varepsilon_0))]_{(a,m)} \in \sR^{C_{\gE}}$\tcp*[r]{\small\frenchspacing $Z_{\gE}$: mean edge length (default)}
}
$\smVl_{\gE} = \sum_{(i,j)\in\gE}Y^{(l)}(\hat{\vx}_{ij})\otimes \vb_{ij} \in \sR^{(2l+1)\times C_{\gE}}$\tcp*[r]{\small\frenchspacing $Y^{(l)}$: real spherical harmonics}

\BlankLine
\tcp*[l]{----- Hybrid Representations -----}
$\smVl = [\smVl_{\gV},\smVl_{\gE}] \in \sR^{(2l+1)\times (C_{\gV}+C_{\gE})}$\;
\end{algorithm}

\FloatBarrier

\subsection{Discrete point-group identification across equivariant GNNs}
\label{sec:exp-synthetic}

We first evaluate SymFP for discrete point-group identification across different equivariant GNN architectures, focusing on whether the observed rank sequences match the theoretical rank templates. Across architectures and training regimes, rank-template matching yields accurate discrete point-group identification, including at random initialization.

Specifically, we construct a synthetic dataset covering all $32$ crystallographic point groups and compare PF with four learnable equivariant GNNs: HEGNN\citep{cen2024are}, TFN\citep{thomas2018tensor}, MACE\citep{batatia2022mace} and Equiformer\citep{liao2023equiformer}, across different representation degrees. For the learnable models, we consider three parameter settings: random initialization, coordinate-denoising pretraining inspired by Uni-Mol\citep{zhou2023unimol}, and supervised training with point-group labels. We evaluate performance using two metrics. Point-group identification accuracy measures the fraction of structures assigned to their true point group by nearest-template matching under the $\ell_1$ distance. Rank-template match rate measures the fraction of samples whose numerical rank sequences exactly match the true group's rank template at every selected degree.

\begin{figure}[!t]
    \centering
    \includegraphics[width=\textwidth]{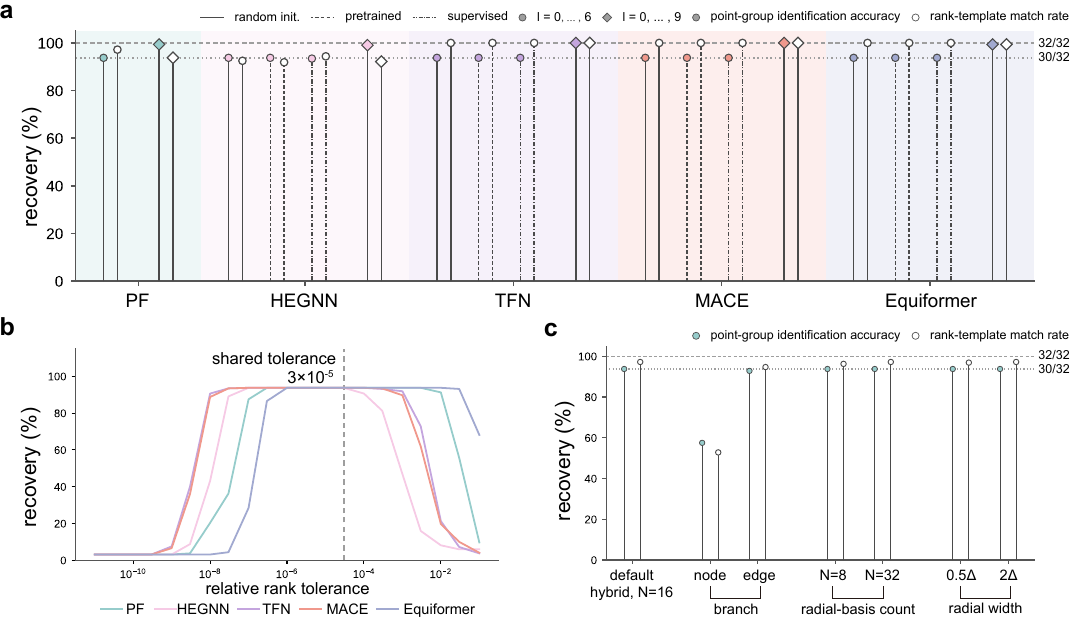}
    \caption[Discrete point-group identification across equivariant GNNs.]{\figtitle{Discrete point-group identification across equivariant GNNs.}
\textbf{a}, Point-group identification accuracy and rank-template match rate across GNN architectures and training states. The $30/32$ reference line marks the identification accuracy limit at $l=0,\ldots,6$ on this balanced benchmark. Owing to computational constraints, learnable GNNs are evaluated only at random initialization for $l=0,\ldots,9$.
\textbf{b}, Sensitivity of point-group identification accuracy to the relative rank tolerance for randomly initialized models at $l=0,\ldots,6$; the shared tolerance $3\times10^{-5}$ used across models is indicated.
\textbf{c}, Ablations of the PF node and edge branches and radial-basis count and width at $l=0,\ldots,6$. $N$ denotes the number of radial bases per branch, and $\Delta$ their centre spacing.}
    \label{fig:synthetic-benchmark}
\end{figure}

In \cref{fig:synthetic-benchmark}a, point-group identification accuracy reaches or closely approaches the $30/32$ template-based limit at $l=0,\ldots,6$, with consistently high rank-template match rates across models and training states. Extending to $l=0,\ldots,9$ raises accuracy to $99.06$--$100\%$ across all models, consistent with the distinct rank templates of all $32$ crystallographic point groups at these degrees (\cref{tab:synthetic-degree-nine}). The observed rank patterns are also reproducible across random seeds (\cref{tab:synthetic-initialization-seeds}). In \cref{fig:synthetic-benchmark}b, PF and all randomly initialized GNNs attain the same identification accuracy at a shared relative rank tolerance, showing that a common numerical threshold suffices across models (\cref{tab:rank-tolerance-sweep}). In the PF ablation (\cref{fig:synthetic-benchmark}c), combining node and edge representations reaches the identification limit at $l=0,\ldots,6$, and varying the number or width of the radial bases leaves identification accuracy unchanged (\cref{tab:pf-ablation}). We therefore use the combined representation with $16$ radial bases per branch in subsequent experiments.

These empirical patterns align with the theoretical analysis presented later. For a fixed input and an architecture satisfying the stated regularity conditions, if the rank sequence matches the theoretical rank template at one parameter setting, then it does so for almost every parameter choice (\cref{cor:generic-full-span,cor:generic-finite-collection}). This result explains the consistency across random initializations and training states, while experiments show that accurate discrete point-group identification is also achieved across GNNs.

\FloatBarrier

\clearpage
\subsection{Spectral responses track symmetry breaking and distinguish deformation modes}
\label{sec:exp-sym-breaking}

\begingroup
{\centering
    \includegraphics[width=\textwidth]{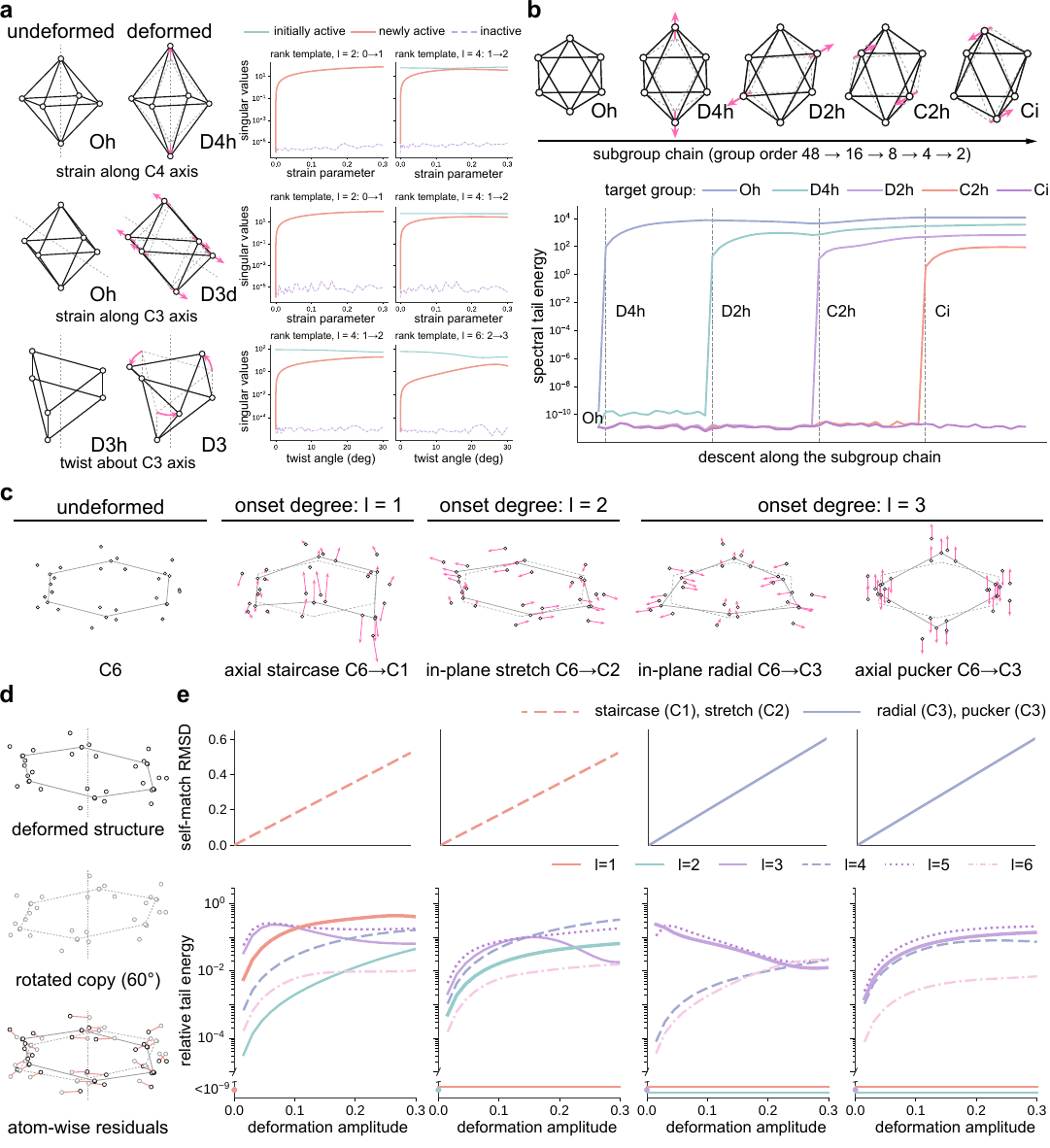}
\par}
\captionof{figure}[SymFP tracks symmetry breaking and distinguishes deformation modes.]{\figtitle{SymFP tracks symmetry breaking and distinguishes deformation modes.}
All panels use PF at $l=0,\ldots,6$.
\textbf{a}, Symmetry breaking immediately activates new singular values, whose number matches the change in the rank template induced by the deformation. Structural illustrations show the three deformations, and curves show the corresponding singular-value trajectories at selected degrees.
\textbf{b}, Along the subgroup chain, spectral tail energies relative to the target point groups rise in sequence and follow the theoretically predicted subgroup monotonicity. Vertical dashed lines mark the symmetry-breaking transitions.
\textbf{c}, Four deformation modes of the same $\mathrm{C}_6$ structure with equal RMSD.
\textbf{d}, Self-match RMSD between a deformed structure and its $60^\circ$-rotated copy, computed using corresponding points and illustrated for the in-plane stretch.
\textbf{e}, Self-match RMSD (top) and degree-resolved relative tail energy (bottom) as functions of deformation amplitude for the four modes in \textbf{c}. The self-match RMSD curves for the four modes largely overlap in pairs, whereas spectral onset degrees further distinguish their angular differences. Relative tail energies below $10^{-9}$ are displayed in a separate numerical-floor strip.}
\label{fig:sym-breaking}
\par
\endgroup

{\looseness=-1
We further evaluate SymFP for continuous symmetry analysis using three controlled deformation experiments: single-step symmetry breaking, progressive symmetry breaking along a subgroup chain, and distinct deformation modes applied to the same structure. Together, these experiments demonstrate the sensitivity of SymFP to symmetry breaking, confirm subgroup monotonicity, and show how degree-resolved spectra distinguish different deformation modes.\par}

{\looseness=-1
In the single-step symmetry-breaking experiment (\cref{fig:sym-breaking}a), the number of activated singular values agrees exactly with the rank template predictions as symmetry is reduced. These singular values rise rapidly at the onset of symmetry breaking and then level off, demonstrating that the singular-value spectrum is highly sensitive to symmetry breaking and can continuously track its evolution. A similar response is observed in the progressive symmetry-breaking experiment along a subgroup chain (\cref{fig:sym-breaking}b), where the spectral tail energy for each target group rises sharply upon symmetry breaking before levelling off, while the relative magnitudes across groups verify subgroup monotonicity.\par}

{\looseness=-1
In the deformation-mode comparison (\cref{fig:sym-breaking}c), we apply four deformations with equal RMSD to the same structure: axial staircase, in-plane stretch, in-plane radial, and axial pucker. Although these deformations contain different Fourier components, their self-match RMSD responses largely collapse onto two curves (\cref{fig:sym-breaking}e). By contrast, the degree-resolved spectra reveal distinct patterns, with onset degrees of $l=1$, $2$, $3$, and $3$, respectively. These onset degrees agree with the lowest nonzero azimuthal orders of the corresponding deformation fields, consistent with the spherical-harmonic constraint $l\geq |m|$. Thus, degree-resolved spectra can distinguish deformation modes that are difficult to distinguish using a single RMSD measure.\par}

{\looseness=-1
We repeat the subgroup-chain and deformation-mode experiments using four randomly initialized learnable equivariant GNNs (\cref{edfig:cross-gnn}). All models reproduce the subgroup-transition responses observed with PF and share the same onset degrees across deformation modes. The only exception is the axial-pucker response of HEGNN, which begins at $l=4$ rather than $l=3$. Introducing node types that preserve the original symmetry restores the onset to $l=3$ (\cref{tab:mode-onsets,sec:appendix-gnn-responses}). These results verify the generality of the symmetry constraints across different GNN architectures, while showing that the specific spectral responses still depend on the model architecture and graph input.\par}

\subsection{Spectral symmetry scores retrieve molecular energy extrema}
\label{sec:exp-torsion}

{\looseness=-1
Having validated the theoretical properties of SymFP, we next explore its practical applications, starting with the retrieval of energy extrema among molecular conformations. This task is motivated by the observation that symmetric conformations often correspond to energy extrema \citep{pophristic2001hyperconjugation,popelier2019relative}. We use a torsion-scan dataset of $86$ aromatic molecules, each containing two planar groups connected by a rotatable bond \citep{smith2021molssi,qiu2020driving}. Each molecule has $24$ conformations sampled at different torsion angles, yielding $2{,}064$ conformations with energy labels (\cref{fig:torsion}a).\par}

\clearpage
\begingroup
{\centering
    \includegraphics[width=\textwidth]{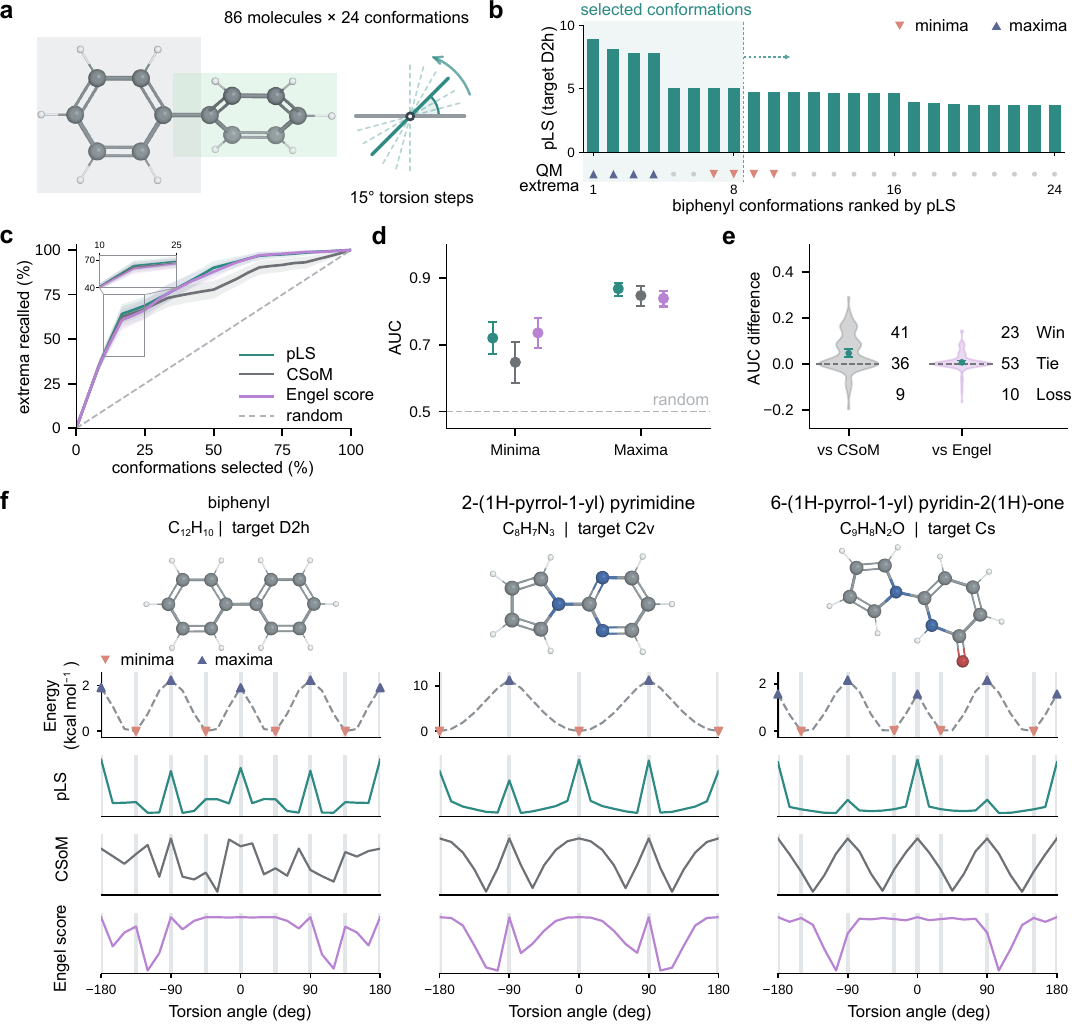}
\par}
\captionof{figure}[Symmetry and energy-extremum retrieval along biaryl torsions.]{\figtitle{Symmetry and energy-extremum retrieval along biaryl torsions.} Throughout, SymFP uses PF representations with uniform pooling over natural-parity degrees $l=0,2,4,6$.
\textbf{a}, Biphenyl torsion schematic and evaluated cohort: 86 molecules, 24 conformations each at $15^\circ$ increments, with deposited coordinates and quantum-mechanical (QM) energies.
\textbf{b}, Biphenyl conformations ranked by descending even-degree negative-log leakage score (pLS) for the $\mathrm{D_{2h}}$ target. Coral downward and blue-grey upward triangles mark minima and maxima in the QM energy profile; grey dots mark other conformations. Shading marks the first eight conformations, and the arrow indicates increasing the selection budget. The four maxima rank $1$--$4$ and the four minima rank $7$--$10$, so the first ten of the $24$ conformations recover all eight sampled energy extrema.
\textbf{c}, Energy-extremum recall against the selected fraction of conformations over 94 molecule--target pairs from 86 molecules. Extrema are defined relative to both periodic neighbours in the deposited energy profile. The inset enlarges the 10--25\% selection range. Mean areas under the recall--budget curve (AUC) are $0.794$, $0.747$ and $0.787$ for SymFP, CSoM and Engel score, respectively, compared with a random-ranking expectation of $0.5$. At a $25\%$ selection budget, the corresponding recall values are $68.8\%$, $66.9\%$ and $66.3\%$.
\textbf{d}, Mean AUC for minima and maxima separately. For SymFP, CSoM and Engel score, respectively, AUCs are $0.720$, $0.647$ and $0.735$ for minima and $0.867$, $0.847$ and $0.839$ for maxima.
\textbf{e}, Per-molecule AUC differences (SymFP minus baseline). Counts give molecule-level wins, numerical ties and losses; SymFP records more wins than losses against either baseline. Violins show kernel-density estimates restricted to the observed range; green points show means. In \textbf{c--e}, targets are averaged within molecules. Shading in \textbf{c} and error bars in \textbf{d,e} give 95\% intervals from 5,000 molecule-bootstrap resamples, pointwise along the curves in \textbf{c}.
\textbf{f}, Energy and score profiles for three molecules illustrate the relationships between symmetry scores and energy extrema across different methods. Triangles mark energy extrema as in \textbf{b}; grey lines align them across profiles. CSoM is sign-reversed, and each score profile is independently scaled to [0,1], so that higher values indicate greater target agreement.}
    \label{fig:torsion}
\par
\endgroup

Using PF representations, we compute SymFP and rank the conformations by the resulting pLS values in descending order for each nontrivial point group encountered along a molecule's torsion scan (\cref{fig:torsion}b). Retrieval performance is summarized by the area under the recall--budget curve (AUC), which plots the fraction of energy extrema recovered against the fraction of ranked conformations selected.

SymFP outperforms CSoM and Engel score \citep{engel2021point} in overall energy-extremum retrieval (\cref{fig:torsion}c), achieving the best performance for maxima while remaining competitive with Engel score for minima (\cref{fig:torsion}d). Molecule-level comparisons show more wins than losses against both baselines (\cref{fig:torsion}e), and individual energy and score profiles further illustrate how the methods respond differently to energy extrema (\cref{fig:torsion}f). To understand the retrieval difference between maxima and minima, we divide the extrema into four groups according to their type and whether their conformations possess the target symmetry. SymFP achieves the highest mean AUC in three of the four groups, falling slightly below Engel score only for minima without the target symmetry (\cref{edtab:torsion-symmetry-groups}a). Moreover, $81.1\%$ of maxima possess the target symmetry, compared with only $29.1\%$ of minima (\cref{edtab:torsion-symmetry-groups}b). This is consistent with biaryl systems such as biphenyl, where highly symmetric conformations can correspond to torsional barriers, while energy minima need not preserve the target symmetry due to competing interactions \citep{popelier2019relative}. The stronger association between symmetry and energy maxima may help explain their better retrieval performance.

We also evaluate SymFP with learnable equivariant GNNs, comparing the four architectures introduced above at random initialization and after total-energy or Uni-Mol-style pretraining. At random initialization, all four GNNs achieve energy-extremum retrieval AUCs close to that of PF, suggesting that equivariant representations already contain symmetry-related structural information useful for this task. Both pretraining objectives enhance spectral contrast by increasing the mean pLS gap between energy maxima and minima (\cref{edtab:torsion-pretraining}a). They also enhance the contrast between conformations with and without the target symmetry, mainly because the mean pLS decreases more strongly for those without the target symmetry (\cref{edtab:torsion-pretraining}b). These results show that while the symmetry constraints captured by SymFP remain consistent across architectures and training states, learning can modulate the spectral organization of representations and enhance symmetry-related contrasts.

\clearpage
\subsection{Degree-resolved spectra reveal crystal-field symmetry constraints}
\label{sec:exp-dota}

\begingroup
{\centering
    \includegraphics[width=\textwidth]{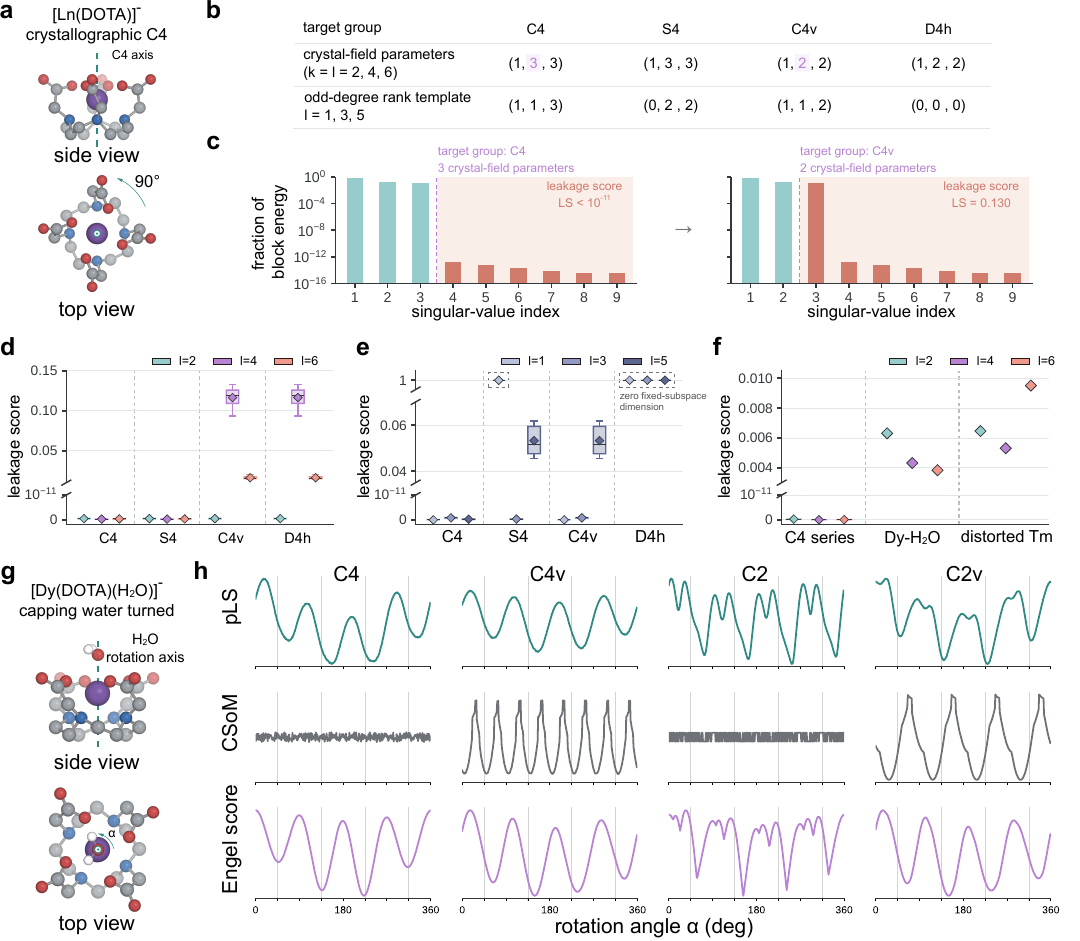}
\par}
\captionof{figure}[Crystal-field constraints and local orientation in lanthanide DOTA complexes.]{\figtitle{Crystal-field constraints and local orientation in lanthanide DOTA complexes (PF throughout).} Even degrees $l=2,4,6$ correspond to the crystal-field ranks, while odd degrees provide complementary geometric symmetry information.
\textbf{a}, Water-free \ce{[Dy(DOTA)]-} cage, representative of seven complexes with a crystallographic four-fold axis.
\textbf{b}, Crystal-field parameter counts at $k=l=2,4,6$ and odd-degree rank templates at $l=1,3,5$ for four tetragonal targets. Highlighted counts correspond to \textbf{c}.
\textbf{c}, The same $l=k=4$ Dy spectrum evaluated against $\mathrm{C}_4$ and $\mathrm{C_{4v}}$. Bars show normalized squared singular values; purple lines mark the target fixed-subspace dimensions. Green and red bars lie within and beyond the corresponding cutoffs, respectively; the red contributions sum to the leakage score (LS). Changing the target from $\mathrm{C}_4$ to $\mathrm{C_{4v}}$ moves the third component into the tail, increasing LS from below $10^{-11}$ to $0.130$.
\textbf{d}, Degree-resolved LS for the seven water-free structures at even degrees $l=2,4,6$. LS remains at the numerical floor for $\mathrm{C}_4$ and $\mathrm{S}_4$, while the five-parameter targets raise LS at $l=4,6$ but not at $l=2$. \textbf{e}, Odd-degree LS at $l=1,3,5$ distinguishes $\mathrm{C}_4$ from $\mathrm{S}_4$, despite their identical even-degree responses.
\textbf{f}, Heavy-atom LS for the $\mathrm{C}_4$ series, water-capped Dy complex and distorted Tm analogue, against $\mathrm{C}_4$ at $l=k=2,4,6$. Both the water-capped Dy complex and the distorted Tm analogue have higher LS than the crystallographic $\mathrm{C}_4$ series. In \textbf{d--f}, boxes span the interquartile range across seven structures, lines mark medians, whiskers span the minimum--maximum range and diamonds mark means; Dy and Tm diamonds in \textbf{f} show individual values.
\textbf{g}, Rigid rotation of the capping water in \ce{[Dy(DOTA)(H2O)]-} about the \ce{Dy-O} axis by $\alpha$, moving only its two protons.
\textbf{h}, Negative-log leakage score (pLS), CSoM and Engel score\citep{engel2021point} profiles for four targets along the water rotation. SymFP and Engel score show a dominant $90^\circ$ period for all four targets. CSoM is nearly constant for $\mathrm{C}_4$ and $\mathrm{C}_2$, and has dominant periods of $90^\circ$ and $45^\circ$ for $\mathrm{C_{2v}}$ and $\mathrm{C_{4v}}$, respectively. Each trace has its own vertical scale. The $\mathrm{C_s}$ result and harmonic content are given in \cref{tab:dota-periodicity}.}
\label{fig:lndota-cf}
\par
\endgroup

We next investigate the relationship between geometric symmetry and crystal-field constraints in DOTA metal complexes, where the arrangement of ligands strongly influences the electronic properties of the central metal ion \citep{gorllerwalrand1996rationalization,cucinotta2012magnetic,briganti2019covalency}.

We first consider seven water-free DOTA complexes with a crystallographic four-fold axis (\cref{fig:lndota-cf}a) \citep{manvell2024lndota}. For the $4f^n$ electronic configuration, the crystal-field terms responsible for splitting electronic energy levels occur at ranks $k=2,4,6$. At these ranks, the numbers of independent crystal-field parameters match the corresponding entries of the SymFP rank template, as both are determined by the same fixed-subspace dimensions (\cref{fig:lndota-cf}b). This correspondence is reflected in the degree-resolved spectra, where different target groups impose distinct constraints on the spectral tail (\cref{fig:lndota-cf}c). Across all seven complexes, LS remains at the numerical floor for $\mathrm{C}_4$ and $\mathrm{S}_4$, while the more restrictive targets $\mathrm{C_{4v}}$ and $\mathrm{D_{4h}}$ yield higher LS at $l=4,6$, where their parameter counts differ (\cref{fig:lndota-cf}d). Although $\mathrm{C}_4$ and $\mathrm{S}_4$ have identical even-degree responses, their odd-degree leakage scores distinguish them (\cref{fig:lndota-cf}e). The water-capped Dy complex and distorted Tm analogue also show higher LS relative to $\mathrm{C}_4$, reflecting departures from ideal four-fold symmetry (\cref{fig:lndota-cf}f). Together, these results show that degree-resolved spectra capture the symmetry constraints underlying crystal-field parameters while revealing geometric symmetry information beyond the even-degree crystal-field description.

We further investigate local orientational changes within the coordination cage. The orientation of the capping water is known to affect the low-lying electronic states and magnetic anisotropy of DyDOTA \citep{cucinotta2012magnetic,briganti2019covalency}. By rotating the capping water about the \ce{Dy-O} axis while keeping the cage fixed (\cref{fig:lndota-cf}g), we find that SymFP exhibits a dominant $90^\circ$ periodicity across the four targets $\mathrm{C}_4$, $\mathrm{C_{4v}}$, $\mathrm{C}_2$ and $\mathrm{C_{2v}}$, consistent with the electronic-property periodicity discussed by \citet{nielsen2025determining}. Engel score shows the same periodicity, whereas CSoM captures it only for the $\mathrm{C_{2v}}$ target \citep{nielsen2025determining} (\cref{fig:lndota-cf}h). This periodicity can be understood from the approximate four-fold symmetry of the coordination cage: a $90^\circ$ global rotation leaves the cage nearly unchanged while shifting the water orientation by $90^\circ$. Together with the global rotation invariance of SymFP, this explains the similar spectral responses at water orientations separated by $90^\circ$.

We also assess the computational efficiency of SymFP on the same system. By avoiding target-orientation searches and reusing spectra across targets, SymFP evaluates seven target groups for Dy-DOTA-water approximately two orders of magnitude faster than the fastest tested Engel implementation and four orders of magnitude faster than CSoM (\cref{edtab:computational-cost}a).

\clearpage
\subsection{SymFP reveals symmetry variation in large protein assemblies}
\label{sec:exp-protein-assemblies}

\begingroup
{\centering
    \includegraphics[width=\textwidth]{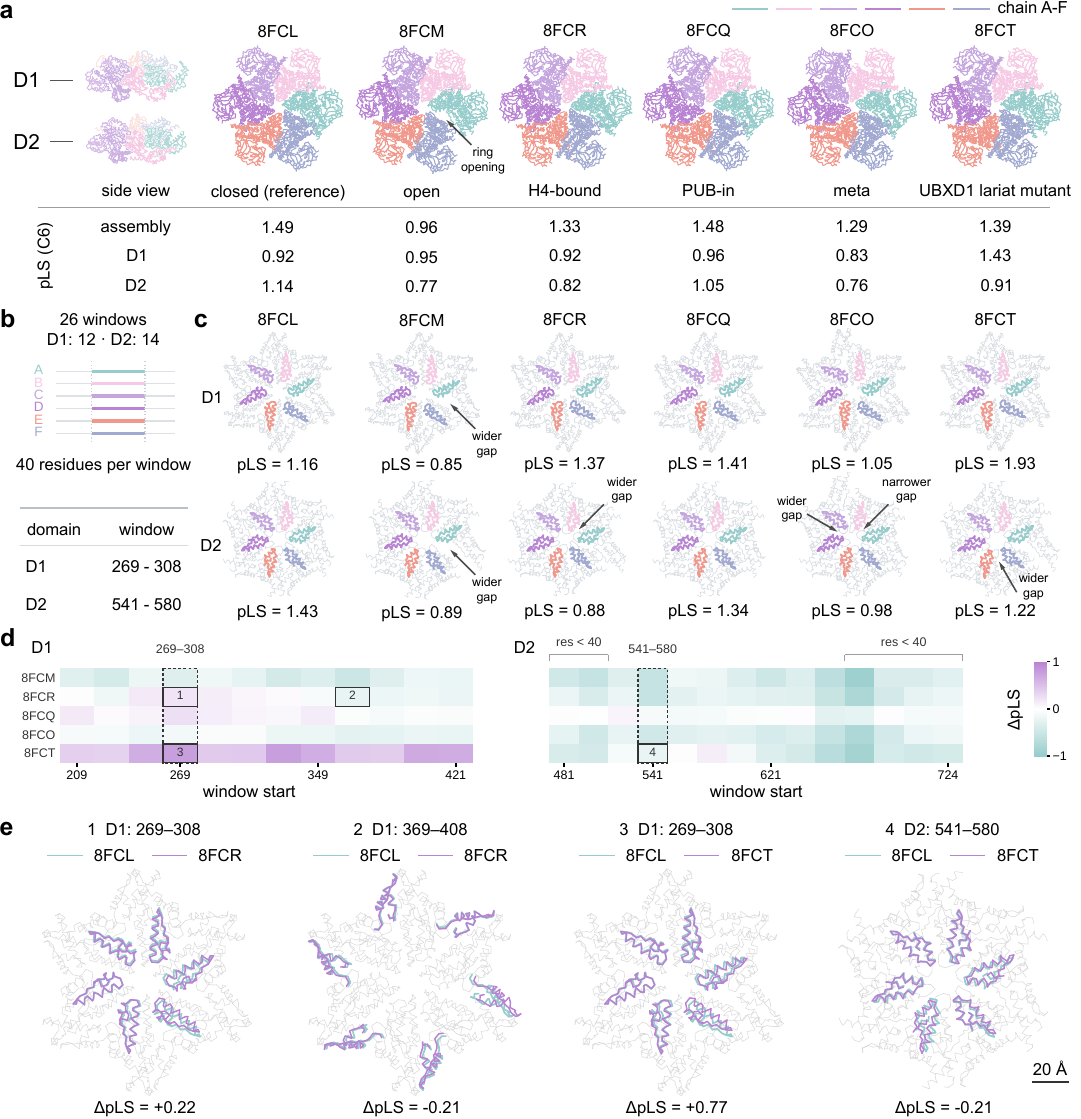}
\par}
\captionof{figure}[Regional symmetry responses across p97 structures.]{\figtitle{Regional symmetry responses across six experimental p97 structures.} Scores use PF at degrees $l=0,\ldots,6$ and common backbone atom identities against a fixed $\mathrm{C}_6$ target. Higher pLS indicates stronger agreement with the $\mathrm{C}_6$ rank-template constraints.
\textbf{a}, Six structures of p97 bound to its adaptor protein UBXD1, reconstructed with $\mathrm{C}_1$ symmetry, with the closed structure (\texttt{8FCL}) as reference\citep{braxton2023p97}. The H4-bound structure has density assigned to the UBXD1 H4 helix at D2. The lariat mutation is in UBXD1. Colours distinguish the six p97 chains.
\textbf{b}, Regional analysis with $12$ D1 and $14$ D2 overlapping $40$-residue windows, each combining corresponding segments from all six chains.
\textbf{c}, The D1 and D2 example windows from \textbf{b} contain the respective pore loops, with the surrounding domains shown in grey. Arrows mark local surface gaps between adjacent chains that widen or narrow relative to the closed reference.
\textbf{d}, Changes in the negative-log leakage score (pLS) for each window in D1 (left) and D2 (right) relative to \texttt{8FCL}, $\Delta\mathrm{pLS}=\mathrm{pLS}_{\mathrm{comparison}}-\mathrm{pLS}_{\mathrm{reference}}$. The H4-bound structure shows both increases and decreases in regional D1 pLS despite its nearly unchanged whole-D1 score in \textbf{a}. The structure associated with the UBXD1 lariat mutant shows increased pLS in all D1 windows and decreased pLS in most D2 windows. The horizontal axis gives window starting residues. Dashed outlines identify the example windows in \textbf{b,c}; brackets labelled res $<40$ mark windows with fewer than $40$ common residues per chain. Numbered cells correspond to \textbf{e}.
\textbf{e}, C$\alpha$-trace superpositions of the four numbered windows, with the reference in teal, the comparison in purple and the surrounding reference domain in grey. Each superposition uses a single rigid fit of the corresponding C$\alpha$ atoms across all six chains. All views share the reference orientation and scale (scale bar, $20$\,\AA). Selection and scoring protocols are given in \cref{sec:appendix-p97-localization}.}
\label{fig:p97}
\par
\endgroup

We further apply SymFP to large protein assemblies, where symmetry governs structural organization and may evolve during functionally relevant conformational rearrangements \citep{braxton2023p97,meyerson2016structural}. We investigate four systems spanning distinct architectures and scales: the AAA+ ATPase p97 ($\sim$4K residues) \citep{braxton2023p97}, influenza haemagglutinin (HA, $\sim$1K) \citep{benton2020structural}, the kainate receptor GluK2 ($\sim$3K) \citep{meyerson2016structural}, and the echovirus 18 (E18) capsid ($>$30K) \citep{buchta2019enterovirus}. We use the PF implementation of SymFP to characterize symmetry variation across structural regions, conformational states, and levels of assembly integrity. Symmetry is quantified by pLS, using backbone atoms for p97, HA, and GluK2 and C$\alpha$ atoms for E18.

We first examine six experimental structures of p97 bound to its adaptor protein UBXD1, using the closed conformation as a reference to assess symmetry changes in the two hexameric rings, D1 and D2, relative to $\mathrm{C}_6$ (\cref{fig:p97}a). To resolve regional differences, we divide each ring into overlapping sequence windows containing corresponding segments from all six chains and compute pLS for each window (\cref{fig:p97}b). Both the open and meta states exhibit lower pLS than the closed reference across all sampled D1 and D2 windows. In contrast, the H4-bound conformation, with density assigned to the UBXD1 H4 helix at D2, has nearly the same whole-D1 score as the closed reference, despite showing both increases and decreases in regional pLS (\cref{fig:p97}c,d). The conformation associated with the UBXD1 lariat mutant exhibits a different pattern, with pLS increasing across all D1 windows but decreasing in most D2 windows. These opposing D1 and D2 responses persist across the tested graph neighbourhoods and window lengths, with shorter windows resolving finer local variations (\cref{edfig:assembly-scale}). Superpositions of selected windows further reveal the structural rearrangements underlying these regional responses (\cref{fig:p97}e). Together, these results show that regional symmetry profiles can reveal local rearrangements and opposing symmetry changes that whole-domain scores may obscure.

We extend the analysis to HA, GluK2 and the E18 capsid, covering different protein architectures and structural states. During the transition from HA state III to IV, the global $\mathrm{C}_3$ pLS increases, whereas the HA1 head and HA2 stem respond in opposite directions, with pLS decreasing in HA1 and increasing in HA2 (\cref{edfig:ha}). In GluK2, the $\mathrm{C}_4$ pLS is generally higher near the transmembrane region than in the upper extracellular regions, both across spatial layers and at the domain level, reflecting differences in subunit arrangement along the receptor axis (\cref{edfig:gluk2}). This pattern persists across the tested layer thicknesses. For the E18 capsid, the degree-six pLS relative to the icosahedral rotation group $\mathrm{I}$ decreases progressively from the intact structure to those lacking one, two or three pentamers (\cref{edfig:e18-capsid}). Together, these examples show that SymFP captures symmetry variation across protein regions and conformational states, as well as changes associated with the integrity of large molecular assemblies.

\section{Discussion}
\label{sec:discussion}

SymFP translates symmetry-imposed constraints on equivariant representations into structural information, providing a unified framework for discrete point-group identification and continuous symmetry analysis. These constraints arise from the algebraic structure of equivariant representations rather than any particular message-passing implementation, establishing a common basis for symmetry analysis across equivariant GNN architectures. The PF implementation constructs representations directly from geometry, enabling SymFP to serve as a readily accessible and reusable tool for structural symmetry characterization.

This formulation allows representation designs and equivariant architectures to be tailored to different systems and analysis objectives. At the analysis level, degree-dependent spectral responses distinguish different forms of structural change, while regional analysis localizes variations that may be obscured by whole-structure scores.

More broadly, SymFP provides a different perspective on symmetry analysis. Existing approaches typically rely on predefined target symmetries and quantify deviations from corresponding reference states. In contrast, SymFP reveals how structural symmetry is encoded in the spectra of equivariant representations, without requiring target-specific geometric references or orientation searches. This perspective provides a general framework for analysing symmetry beyond individual point groups and may extend to broader aspects of symmetry, including space groups, chirality, and helical symmetries. More importantly, it opens opportunities for studying how symmetry constraints reorganize during complex structural transformations, including phase transitions, defect evolution, and amorphization.

\section{Methods}
\label{sec:methods}

\subsection{Preliminaries}
\label{sec:preliminaries}

\subsubsection*{Geometric graphs and point-group actions}

A geometric graph with $N$ nodes is denoted by $\gG(\mH,\gmX;\mA)=\gG(\gV;\gE)$, where $\gV\coloneqq\{1,\dots,N\}$ is the node set, $\mH\coloneqq[\vh_1,\dots,\vh_N]\in\R^{d_v\times N}$ contains node attributes, $\gmX\coloneqq[\gvx_1,\dots,\gvx_N]\in\R^{3\times N}$ contains 3D coordinates, and $\mA\in\R^{N\times N}$ is the adjacency matrix. Both node attributes and coordinates are stored column-wise. The directed edge set $\gE\subseteq\gV\times\gV$ is induced by $\mA$, with $(i,j)\in\gE$ if and only if $\mA_{ij}\neq0$. For undirected neighbourhoods, both $(i,j)$ and $(j,i)$ are included. Edge attributes $\ve_{ij}$ may be assigned to edges when needed.

As translations are removed by centring, we restrict attention to $\mathrm{O}(3)$ actions. For $\Gh\in\mathrm{O}(3)$, we write $\Gh\cdot\gG$ for its action on $\gG$, with coordinates transformed as $\mO_{\Gh}\gmX$, where $\mO_{\Gh}\in\R^{3\times3}$ is the corresponding orthogonal matrix.

\subsubsection*{Compatibility and stabilizers}

\begin{definition}[Graph Compatible with a Finite Point Group]
\label{def:symmetric-graph}
Let $\GH\leq\mathrm{O}(3)$ be a finite subgroup. A geometric graph $\gG(\mH,\gmX;\mA)=\gG(\gV;\gE)$ is \emph{compatible with} $\GH$ if, for every $\Gh\in\GH$ with orthogonal action $\mO_{\Gh}$, there exists a permutation matrix $\mP_{\Gh}\in S_N$ such that
\begin{equation}
    \mO_{\Gh}\gmX=\gmX\mP_{\Gh},
    \qquad
    \mH\mP_{\Gh}=\mH,
    \qquad
    \mP_{\Gh}^{\top}\mA\mP_{\Gh}=\mA,
\end{equation}
and, when edge attributes are present, $\ve_{ij}=\ve_{\pi_{\Gh}(i)\pi_{\Gh}(j)}$ for all $(i,j)\in\gE$, where $\pi_{\Gh}$ is the permutation associated with $\mP_{\Gh}$. We then write $\Gh\cdot\gG=\gG$ and denote the set of graphs compatible with $\GH$ by $\sG(\GH)$.
\end{definition}

The equality $\Gh\cdot\gG=\gG$ is understood up to node relabelling that preserves attributes.

\begin{definition}[Stabilizer]
\label{def:stabilizer}
The stabilizer of a geometric graph $\gG$ is
$\operatorname{Stab}(\gG)\coloneqq\{\Gg\in\mathrm{O}(3):\Gg\cdot\gG=\gG\}$.
It is a subgroup of $\mathrm{O}(3)$ and, when finite, gives the exact point group of $\gG$.
\end{definition}

Exact point-group identification statements below assume a finite stabilizer.

\subsubsection*{Irreducible representations of $\mathrm{O}(3)$}

Every $\Gh\in\mathrm{O}(3)$ has a unique decomposition $\Gh=\Gr\Gm$, where $\Gr\in\mathrm{SO}(3)$ and $\Gm\in\mathrm{C_i}$, with $\mathrm{C_i}=\{\Ge,\Gi\}$ denoting the two-element inversion group. Thus, $\mathrm{O}(3)\cong\mathrm{SO}(3)\times\mathrm{C_i}$. For each degree $l\in\mathbb{N}_0=\{0,1,\ldots\}$ and parity $p\in\{1,-1\}$, define
\begin{equation}
\label{eq:rho}
    \rho^{(l,p)}(\Gh)
    \coloneqq
    \sigma^{(p)}(\Gm)\mD^{(l)}(\Gr),
\end{equation}
where $\sigma^{(p)}(\Ge)=1$, $\sigma^{(p)}(\Gi)=p$, and $\mD^{(l)}(\Gr)\in\R^{(2l+1)\times(2l+1)}$ is the real Wigner-$D$ matrix of degree $l$, following the real spherical-harmonic basis and sign convention of the equivariant-operator library used in our implementation \citep{geiger2022e3nn}. Throughout, we use natural-parity representations with $p_l=(-1)^l$ and write $\rho^{(l)}\coloneqq\rho^{(l,(-1)^l)}$. We denote the degree set $\{0,\ldots,L\}$ by $\Lambda_L$. Degree-$l$ real spherical harmonics transform under $\rho^{(l)}$ and satisfy $Y^{(l)}(-\gvx)=(-1)^lY^{(l)}(\gvx)$ \citep{weiler20183d}.

\subsection{Related work}
\label{sec:related-work}

\subsubsection*{Existing symmetry measures}

Classical continuous symmetry measures (CSMs) provide the standard geometry-level approach to approximate symmetry. In an abstract distance-based formulation, a CSM for a target point group $\GH$ measures the distance from a geometric graph $\gG$ to the nearest $\GH$-compatible realization,
\begin{equation}
    \mathrm{CSM}_{\GH}(\gG)\coloneqq \inf_{\gG^\star\in\sG(\GH)} d(\gG,\gG^\star),
\end{equation}
where the infimum runs over the geometric graphs compatible with $\GH$ in the sense of \cref{def:symmetric-graph}\citep{zabrodsky1992continuous,zabrodsky1993continuous}. The comparison keeps the graph structure and attributes fixed while varying the coordinates under the allowed node correspondences. Subsequent work developed analytical formulas for selected symmetry families\citep{pinsky2008analytical}, accelerated exact computation through structure-preserving permutations\citep{alon2018improved}, specialized the machinery to protein homomers\citep{tuviarad2019improved}, extended it to large molecular structures through approximate algorithms\citep{alon2023continuous}, and packaged the methodology into modern software for quantitative structural analysis\citep{tuviarad2024csm}. Continuous shape measures (CShM) compare a coordination sphere with a reference polyhedron of the same vertex count\citep{llunell2013shape}. More recently, the continuous symmetry operation measure (CSoM) replaces the nearest-symmetric reconstruction with an average of operation-wise deviations over the target group\citep{nielsen2024evaluation,nielsen2025determining}:
\begin{equation}
\label{eq:csom-definition}
    \mathrm{SoM}_{\Gh}(\gG)\coloneqq d(\gG,\Gh\cdot \gG),
    \qquad
    \mathrm{CSoM}_{\GH}(\gG)\coloneqq \frac{1}{|\GH\setminus\{\Ge\}|}\sum_{\Gh\in\GH\setminus\{\Ge\}} \mathrm{SoM}_{\Gh}(\gG).
\end{equation}
The CSoM average in \cref{eq:csom-definition} assumes $\GH\ne\{\Ge\}$. CSoM evaluates target operations in the coordinate frame of the structure, requiring the target orientation to be supplied or determined.

The operation-based view has a harmonic analogue. Replacing the coordinate-space discrepancy by a squared residual between an equivariant harmonic embedding and its transformed copy gives a CSoM-style comparison in representation space. Averaging over the full target group yields twice the energy outside its oriented fixed subspaces.

Engel's harmonic construction uses the complementary retained energy for point-group analysis\citep{engel2021point}. The geometric graph enters through a normalized harmonic embedding $\svv(\gG)\coloneqq(\svvl(\gG))_l$, the spherical-harmonic coefficients of the pooled distribution of bond directions on the sphere, with one coefficient vector at each retained degree and $\sum_l\|\svvl\|_2^2=1$. A spatial realization of the target group then determines the corresponding projector: averaging $\rho^{(l)}$ over an oriented copy $\Gr\GH\Gr^{-1}$ of the target group,
\begin{equation}
    \mP_{\GH}^{(l)}(\Gr)
    \coloneqq
    \frac{1}{|\GH|}\sum_{\Gh\in\GH}\rho^{(l)}(\Gr\Gh\Gr^{-1}),
\end{equation}
yields the orthogonal projector onto the corresponding fixed subspace\citep{linehan2001little}. Write $\alpha_{\GH}^{(l)}(\gG;\Gr)\coloneqq \|\mP_{\GH}^{(l)}(\Gr)\svvl\|_2^2$ for the energy retained in this fixed subspace at degree $l$, and $\alpha_{\GH}$ for the sum over the retained degrees. Averaging the squared harmonic residuals over the full target group, including the identity, gives
\begin{equation}
\label{eq:harmonic-operation-residual}
    R_{\GH}(\gG;\Gr)
    \coloneqq \frac{1}{|\GH|}\sum_{\Gh\in\GH}\sum_l
    \|\svvl-\rho^{(l)}(\Gr\Gh\Gr^{-1})\svvl\|_2^2
    =2(1-\alpha_{\GH}(\gG;\Gr)).
\end{equation}
The derivation is given in \cref{sec:appendix-csom-derivation}.

The trace of $\mP_{\GH}^{(l)}(\Gr)$ equals the dimension of the degree-$l$ fixed subspace, denoted $m_l(\GH)$. The normalization in Engel's construction\citep{engel2021point} compares retained energy against a baseline determined by the target. Under an equal-energy assumption over the harmonic coordinates, this baseline is
\begin{equation}
    \beta_{\GH}^{(l)}
    \coloneqq
    \frac{\tr(\mP_{\GH}^{(l)}(\Gr))}{2l+1},
    \qquad
    \beta_{\GH}
    \coloneqq
    \frac{\sum_l \tr(\mP_{\GH}^{(l)}(\Gr))}{\sum_l (2l+1)}
    =
    \frac{\sum_l m_l(\GH)}{\sum_l (2l+1)}.
\end{equation}
The traces, and hence $\beta_{\GH}$, are invariant under target reorientation. For $\beta_{\GH}<1$, the normalized retained-energy score at a fixed orientation is
\begin{equation}
    \Es_{\GH}(\gG;\Gr)
    \coloneqq
    \frac{\alpha_{\GH}(\gG;\Gr)-\beta_{\GH}}{1-\beta_{\GH}}.
\end{equation}
Following the orientation search adopted in subsequent work by Engel and colleagues\citep{mbah2023early}, we use \emph{Engel score}
\begin{equation}
\label{eq:engel-max-orientation}
    \Es_{\GH}(\gG)\coloneqq\max_{\Gr\in\mathrm{SO}(3)}\Es_{\GH}(\gG;\Gr),
\end{equation}
with one spatial rotation shared across all retained degrees. Larger values indicate greater retained energy.

Alongside these general constructions, a number of methods are specialized to particular settings. A maximum-symmetry search returns the largest point group consistent with a predefined tolerance\citep{pilati1998symmol}, and related procedures assign a point group to a nearly symmetric molecule\citep{largent2012symmetrizer,knowles2022determination}. Others restrict the target instead: continuous measures have been derived for individual group families, notably $\mathrm{C_{nv}}$ and $\mathrm{D_n}$\citep{pinsky2013continuous}, or extended from point sets to vectors, matrices and operators\citep{dryzun2009generalization}.

\subsubsection*{From symmetry-induced constraints to spectral analysis}

The constructions above compare coordinates with a symmetric realization or evaluate a spatial realization of the target operations. SymFP instead starts from the rank constraints that input symmetry imposes on equivariant representation blocks and uses the target's fixed-subspace dimensions to read their singular-value spectra, without constructing a symmetric reference or selecting a target orientation.

Classical analysis identified the combinations of spherical harmonics fixed by individual point groups\citep{meyer1954symmetries}. Cen et al.\citep{cen2024are} and Lin et al.\citep{lin2026reducing} studied how input symmetry can reduce the expressivity of equivariant representations through representation degeneracy and symmetry increase. Related studies have used equivariant networks to search for symmetry-breaking order parameters\citep{smidt2021finding} and examined symmetry breaking through Curie-type constraints on equivariant maps\citep{kaba2023symmetry}.

Operationally, we collect the fixed-subspace dimensions into rank templates and compare them with rank sequences obtained from the singular-value spectra of equivariant representation blocks. Rank-template matching uses the complete rank sequence for discrete point-group identification, while leakage scores provide an orientation-invariant, continuous measure of departures from target-imposed rank constraints.

\subsection{Fixed-subspace derivation of rank templates}
\label{sec:rank-template-derivation}

The rank templates follow from the dimensions of the fixed subspaces imposed by input symmetry on equivariant representations. Detailed proofs are provided in \cref{sec:appendix-theory}.

\begin{restatable}[Fixed-subspace constraint]{theorem}{FixedSubspaceConstraint}
\label{theo:fixed-subspace}
Suppose that $\smVl$ is an $\mathrm{O}(3)$-equivariant representation map satisfying \cref{eq:results-equivariance}. For any $\gG\in\sG(\GH)$, write $\sF^{(l)}(\GH)\coloneqq\{\svv\in\R^{2l+1}:\rho^{(l)}(\Gh)\svv=\svv\ \forall\Gh\in\GH\}$ for the fixed subspace and $\rho^{(l)}(\GH)\coloneqq1/|\GH|\cdot\sum_{\Gh\in\GH}\rho^{(l)}(\Gh)$ for the group average. Then
\begin{equation}
\thmrelabel{eq:fixed-subspace-column}
    \operatorname{colspan}(\smVlgG)
    \subseteq
    \sF^{(l)}(\GH),
\end{equation}
and consequently
\begin{equation}
\thmrelabel{eq:fixed-subspace-rank}
    \operatorname{rank}(\smVlgG)
    \le
    m_l(\GH),
\end{equation}
where $m_l(\GH)$ is the multiplicity of the trivial representation in $\rho^{(l)}|_{\GH}$, that is $m_l(\GH)\coloneqq\dim\sF^{(l)}(\GH)=\operatorname{rank}(\rho^{(l)}(\GH))=\tr(\rho^{(l)}(\GH))$.
\end{restatable}

The fixed-subspace dimension $m_l(\GH)$ can be computed from the character $\chi^{(l)}(\Gh)\coloneqq\tr(\rho^{(l)}(\Gh))$ of the degree-$l$ representation:
\begin{equation}
\label{eq:character-average}
    m_l(\GH)
    =
    \frac{1}{|\GH|}\sum_{\Gh\in\GH}\chi^{(l)}(\Gh).
\end{equation}
Collecting these dimensions over a selected degree set $\Lambda\subset\mathbb{N}_0$ gives the rank template
\begin{equation}
\label{eq:rank-template}
    \mathrm{RankTemplate}_{\Lambda}(\GH)\coloneqq (m_l(\GH))_{l\in\Lambda}.
\end{equation}
Under a change of coordinate frame, the fixed subspace transforms as
\begin{equation}
\label{eq:conjugacy-fixed-space}
    \sF^{(l)}(\Gg\GH\Gg^{-1})
    =
    \rho^{(l)}(\Gg)\,\sF^{(l)}(\GH),
    \qquad
    m_l(\Gg\GH\Gg^{-1})=m_l(\GH),
\end{equation}
for every finite point group $\GH\leq\mathrm{O}(3)$, every $\Gg\in\mathrm{O}(3)$, and every degree $l$. Hence, the rank template is invariant under target reorientation. For the 32 crystallographic point groups, the number of distinct templates depends on the selected degrees.

\begin{restatable}[Rank templates for the 32 crystallographic point groups]{corollary}{CrystallographicSignatureCorollary}
\crefalias{theorem}{corollary}
\label{cor:crystallographic-32}
Let $\sC_{32}$ denote the 32 crystallographic point groups\citep{aroyo2016international}. The map $\GH\mapsto \mathrm{RankTemplate}_{\Lambda_{9}}(\GH)$ is injective on $\sC_{32}$. The shorter degree set $\Lambda_{6}$ gives 30 distinct templates, with only the pairs $\mathrm{D}_6/\mathrm{D_{6h}}$ and $\mathrm{O}/\mathrm{O_h}$ sharing templates.
\end{restatable}

The corresponding multiplicities are listed in \cref{tab:crystallographic-rank-templates}. Rank templates also satisfy the following ordering under subgroup inclusion.

\begin{restatable}[Subgroup monotonicity]{proposition}{SubgroupMonotonicityProposition}
\crefalias{theorem}{proposition}
\label{prop:subgroup-monotonicity}
If $\GH_1\leq \GH_2\leq \mathrm{O}(3)$, then $\sF^{(l)}(\GH_2)\subseteq \sF^{(l)}(\GH_1)$ and $m_l(\GH_2)\leq m_l(\GH_1)$ for every $l$. Consequently, $\mathrm{RankTemplate}_{\Lambda}(\GH_2)\leq \mathrm{RankTemplate}_{\Lambda}(\GH_1)$ componentwise for every finite degree set $\Lambda$.
\end{restatable}

\subsection{Spectral formulation and numerical evaluation}
\label{sec:spectral-formulation}

\subsubsection*{Spectral symmetry fingerprint}

For a finite set of degrees $\Lambda$, consider the representation blocks $\smVlgG\in\mathbb{R}^{(2l+1)\times C}$, each with $d_l\coloneqq\min\{2l+1,C\}$ singular values. Under the default channel count, $d_l=2l+1$. For each $l\in\Lambda$, let
\begin{equation}
\label{eq:sv-spectrum}
    \sigma_1^{(l)}\geq \sigma_2^{(l)}\geq \cdots \geq \sigma_{d_l}^{(l)}\geq 0
\end{equation}
be the ordered singular values of $\smVl$, with the graph argument suppressed when unambiguous and $\sigma_i^{(l)}=0$ for $i>d_l$. Write $\boldsymbol{\sigma}^{(l)}\coloneqq(\sigma_i^{(l)})_{i=1}^{d_l}$. The spectral symmetry fingerprint concatenates these spectra across degrees:
\begin{equation}
\label{eq:symfp}
    \mathrm{SymFP}_{\Lambda}(\gG)\coloneqq (\sigma_i^{(l)})_{l\in\Lambda,\ 1\leq i\leq d_l}.
\end{equation}
SymFP is invariant under global reorientation (\cref{prop:spectral-invariance}). For $\gG\in\sG(\GH)$, the fixed-subspace constraint (\cref{theo:fixed-subspace}) imposes
\begin{equation}
\label{eq:spectral-rank-constraint}
    \operatorname{rank}(\smVl)\leq m_l(\GH), \qquad
    \sigma_i^{(l)}=0 \quad \text{for } m_l(\GH)<i\leq d_l.
\end{equation}

\subsubsection*{Rank-template matching}

We estimate numerical ranks using a graph-dependent threshold that combines absolute and relative tolerances:
\begin{equation}
\label{eq:numerical-rank-threshold}
    \tau(\gG)\coloneqq \max\{\tau_{\mathrm{abs}},\ \tau_{\mathrm{rel}}\,\sigma_1^{(0)}(\gG)\},
\end{equation}
where $\tau_{\mathrm{abs}}$ and $\tau_{\mathrm{rel}}$ are the respective tolerances, with the relative term scaled by the leading degree-zero singular value. The numerical rank at degree $l$ is
\begin{equation}
\label{eq:numerical-rank}
    \widehat{m}_{l,\tau}(\gG)\coloneqq |\{i\in\{1,\dots,d_l\}: \sigma_i^{(l)}>\tau\}|.
\end{equation}
These ranks form the numerical rank sequence
\begin{equation}
\label{eq:numerical-rank-sequence}
    \widehat{\mathrm{RankSeq}}_{\Lambda,\tau}(\gG)\coloneqq (\widehat{m}_{l,\tau})_{l\in\Lambda}.
\end{equation}
Given a candidate family $\sC$, classification minimizes the weighted $\ell_1$ distance to the candidate rank templates:
\begin{equation}
\label{eq:rank-template-matching}
    \widehat{\GH}^{\mathrm{hard}}_{\Lambda,\tau}(\gG)
    \in
    \arg\min_{\GH\in\sC}
    \sum_{l\in\Lambda}\alpha_l|\widehat{m}_{l,\tau}-m_l(\GH)|,
\end{equation}
with uniform weights $\alpha_l=1$. The set of minimizers is invariant under global reorientation, since both the singular values and rank templates are orientation-invariant.

\subsubsection*{Leakage scores}

The degree-resolved leakage score in \cref{eq:results-leakage} is
\begin{equation}
\label{eq:degree-leakage}
    \mathrm{LS}^{(l)}_{\GH}(\gG)
    =\frac{1}{Z_l(\gG)}\sum_{i>m_l(\GH)}(\sigma_i^{(l)})^2,
    \qquad
    Z_l(\gG)=\sum_i(\sigma_i^{(l)})^2+\varepsilon_{\mathrm{norm}}.
\end{equation}
Pooling uniformly over degrees $l\in\Lambda$ gives
\begin{equation}
\label{eq:pooled-leakage}
    \mathrm{LS}_{\Lambda,\GH}(\gG)
    =\frac{1}{|\Lambda|}\sum_{l\in\Lambda}\mathrm{LS}^{(l)}_{\GH}(\gG),
\end{equation}
with the logarithmic score
\begin{equation}
\label{eq:pls}
    \mathrm{pLS}_{\Lambda,\GH}(\gG)
    =-\log_{10}\!\left(\mathrm{LS}_{\Lambda,\GH}(\gG)+\varepsilon_{\log}\right).
\end{equation}
We set $\varepsilon_{\log}=10^{-10}$, yielding a maximum pLS of $10$. A pLS of $3$ corresponds to a mean spectral leakage below $0.1\%$, indicating near-satisfaction of the target rank constraints. However, a high pLS does not necessarily imply the presence of the target symmetry. For example, over degrees $l=0,\ldots,9$, structures with point groups $\mathrm{C}_2$ and $\mathrm{T_h}$ exhibit zero LS relative to $\mathrm{C_s}$ and $\mathrm{D_{6h}}$, respectively, despite lacking these symmetries. This occurs because the ranks of their equivariant representations remain within the corresponding target fixed-subspace dimensions (\cref{tab:crystallographic-rank-templates}). Discrete point-group identification based on rank-template matching is therefore more appropriate for this purpose.

More generally, pooling admits nonnegative weights $\beta_l$ summing to one; all evaluations use uniform weights $\beta_l=1/|\Lambda|$.

\subsubsection*{Normalization}

When $m_l(\GH)=0$, all block energy lies in the spectral tail, causing degree-wise normalized LS to approach unity when the block energy substantially exceeds $\varepsilon_{\mathrm{norm}}$. To avoid this saturation for centrosymmetric targets, the torsion analysis pools only even degrees. Alternative normalizations,
\begin{equation}
\label{eq:alternative-normalization}
    Z_l(\gG)=1,
    \qquad\text{or}\qquad
    Z_l(\gG)=\sum_{k\in\Lambda}\sum_i(\sigma_i^{(k)})^2+\varepsilon_{\mathrm{norm}},
\end{equation}
retain absolute block energies or relative cross-degree energies, respectively. Engel score similarly employs global energy normalization, made explicit by
\begin{equation}
\label{eq:engel-global-normalization}
    \Es_{\GH}(\gG)
    =1-\frac{1}{1-\beta_{\GH}}
    \min_{\Gr\in\mathrm{SO}(3)}
    \frac{\sum_l\|(\mI-\mP_{\GH}^{(l)}(\Gr))\svvl\|_2^2}
         {\sum_l\|\svvl\|_2^2}.
\end{equation}

The subgroup ordering of fixed-subspace dimensions induces the following monotonicity of leakage scores under each of the normalization choices above.

\begin{restatable}[Leakage monotonicity along subgroup chains]{proposition}{LeakageMonotonicityProposition}
\crefalias{theorem}{proposition}
\label{prop:leak-monotonicity}
If $\GH_1\leq \GH_2\leq \mathrm{O}(3)$, then for every geometric graph $\gG$,
\begin{equation}\thmrelabel{eq:leak-monotonicity}
    \mathrm{LS}_{\Lambda,\GH_1}(\gG)\leq \mathrm{LS}_{\Lambda,\GH_2}(\gG),
    \qquad
    \mathrm{pLS}_{\Lambda,\GH_1}(\gG)\geq \mathrm{pLS}_{\Lambda,\GH_2}(\gG).
\end{equation}
\end{restatable}

This ordering need not hold when the pooled degrees depend on the target. Task-specific choices and limiting cases are given in \cref{sec:appendix-shared-settings,sec:appendix-leakage-normalization}.

\subsection{Construction of the parameter-free equivariant GNN}
\label{sec:pf-construction}

In \cref{alg:pf}, $\gvx_c$ denotes the graph centroid, with $\gvx_{ic}\coloneqq\gvx_i-\gvx_c$ and $\gvx_{ij}\coloneqq\gvx_j-\gvx_i$ representing centred node vectors and edge displacement vectors, respectively. Their unit directions are denoted by $\hat{\vx}_{ic}$ and $\hat{\vx}_{ij}$ wherever defined. The radial scales $Z_{\gV}$ and $Z_{\gE}$ are the mean centred-node radius and mean edge length, respectively, with a fixed $\varepsilon_0>0$ used to avoid zero denominators. The functions $\{\rho_n\}$ and $\{\eta_m\}$ are fixed Gaussian radial bases. For zero vectors, we set $Y^{(l)}(\bm{0})=\bm{0}_{2l+1}$ for $l>0$, while retaining the degree-zero scalar. If node attributes are absent, we use $d_v=1$ and $\vh_i\equiv1$; if edge attributes are absent, we set $d_e=0$. In all reported PF evaluations, node attributes $\vh_i$ are scalar atomic numbers, and edge attributes $\ve_{ij}=\|\gvx_j-\gvx_i\|$ are edge lengths in the input coordinate units. Both attributes are scalar, with $d_v=d_e=1$, and neither is rescaled. Implementation constants are listed in \cref{sec:appendix-shared-settings}.

In the edge block, keeping $\vh_i$ and $\vh_j$ in separate channels can retain odd-degree information that would otherwise cancel when opposite edge directions are pooled with symmetric scalar weights \citep{engel2021point}. For paired reverse edges, this cancellation is no longer enforced when the endpoint attributes differ.

For each degree $l$, the three PF blocks satisfy the equivariance relation in \cref{eq:results-equivariance}:
\begin{equation}
    \smVl_{\gV}(\Gg\cdot\gG)
    =
    \rho^{(l)}(\Gg)\smVl_{\gV}(\gG),
    \quad
    \smVl_{\gE}(\Gg\cdot\gG)
    =
    \rho^{(l)}(\Gg)\smVl_{\gE}(\gG),
    \quad
    \smVl(\Gg\cdot\gG)
    =
    \rho^{(l)}(\Gg)\smVl(\gG).
\end{equation}
Equivariance follows because centring commutes with orthogonal transformations, radial weights depend only on norms, and spherical harmonics transform by $\rho^{(l)}$. Every stabilizer operation permutes the nodes and therefore preserves their centroid, so centring removes global translations without changing the symmetry constraints.

\subsection{Theoretical properties and variational interpretation}
\label{sec:theoretical-properties}

\subsubsection*{Generic rank realization and stability}

For a fixed input and degree, an equivariant architecture attains its maximal attainable rank at almost every parameter setting, provided its representation block depends real-analytically on the parameters over a connected open domain (\cref{prop:generic-maximal-rank}). In particular, full span at a single parameter setting guarantees full span almost everywhere, with the columns spanning the fixed subspace and the rank equaling its dimension (\cref{cor:generic-full-span}). If full span holds across the selected degrees, distinct rank templates identify the input's point group within the candidate family, provided the numerical threshold is positive and smaller than every positive singular value. We assess rank attainment across learnable GNN architectures and training regimes, alongside PF, in the controlled benchmark (\cref{tab:synthetic-training-regimes}).

For graph families with fixed attributes and adjacency, the spectral descriptor inherits local Lipschitz continuity from the representation blocks (\cref{theo:local-stability}). Comparison with the target-symmetric subset then bounds spectral departure in terms of distance to that subset, linking the descriptor to a point-group-conditioned continuous symmetry measure (\cref{eq:symfp-csm-bound,cor:leak-csm}).

\subsubsection*{Variational relation to operation-based symmetry measures}

The spectral retained energy admits a variational characterization that connects SymFP to the retained-energy form of operation-based symmetry analysis introduced in \cref{sec:related-work}. Write $\mC^{(l)}\coloneqq \smVl(\smVl)^\top$ and $k\coloneqq m_l(\GH)$. SymFP retains the energy $\gA_{\GH}^{(l),\mathrm{symfp}}\coloneqq \sum_{i=1}^{k}(\sigma_i^{(l)})^2$, whose variational form follows from the Ky Fan maximum principle\citep{fan1949theorem}:
\begin{equation}
\label{eq:spectral-retained-energy-variational}
    \gA_{\GH}^{(l),\mathrm{symfp}}
    =
    \max_{\substack{
        \mP^2=\mP=\mP^\top\\
        \operatorname{rank}(\mP)=k
    }}
    \operatorname{tr}(\mP\mC^{(l)}).
\end{equation}

For the same multi-channel block, a spatial realization of the target group determines a fixed-subspace projector. The retained energy is the trace of this projector against $\mC^{(l)}$. For a single harmonic channel $\smVl=\svvl$, this is the retained-energy term in Engel score at the specified orientation. To relate this projection-based retained energy to SymFP's spectral variational form, we optimize the retained energy at each degree separately:
\begin{equation}
\label{eq:op-retained-energy-main}
    \gA_{\GH}^{(l),\mathrm{op}}
    \coloneqq
    \max_{\Gr\in\mathrm{SO}(3)}
    \operatorname{tr}\!\left(
        \mP_{\GH}^{(l)}(\Gr)\mC^{(l)}
    \right),
\end{equation}
where $\mP_{\GH}^{(l)}(\Gr)$ projects onto the fixed subspace of the spatially realized copy $\Gr\GH\Gr^{-1}$. The two formulations share the variational objective $\operatorname{tr}(\mP\mC^{(l)})$ but differ in their feasible projector sets. SymFP ranges over all rank-$k$ orthogonal projectors, while the operation-based formulation is restricted to projectors onto $\rho^{(l)}(\Gr)\sF^{(l)}(\GH)$ with $\Gr\in\mathrm{SO}(3)$. Since every such projector has rank $k$, we obtain
\begin{equation}
\label{eq:op-symfp-ordering}
    \gA_{\GH}^{(l),\mathrm{op}}
    \le
    \gA_{\GH}^{(l),\mathrm{symfp}}.
\end{equation}
Equality holds in particular for exact-$\GH$-symmetric inputs, for which both optima equal $\|\smVl\|_F^2$. For distorted inputs, optimizing the target orientation selects a reference under the chosen projection criterion. The two formulations offer distinct perspectives: agreement with target operations and target rank constraints. Once the representation blocks are computed, the same singular-value spectra give the SymFP retained energies and leakage scores for every target through its fixed-subspace dimensions. The retained-energy identities and general equality conditions are derived in \cref{sec:appendix-op-relation}.

\subsection{Datasets and evaluation}
\label{sec:methods-data}

\subsubsection*{Synthetic point clouds}

For each of the $32$ crystallographic point groups, we generate $100$ finite point clouds by combining point-group orbits and independently verify their stabilizers. Each group is split into training, validation, and test sets in an $80/10/10$ ratio. Rank tolerances are selected on the validation set and held fixed across GNNs (\cref{sec:appendix-synthetic-protocol}).

\subsubsection*{Controlled deformations}

We induce single-step symmetry breaking by straining a unit octahedron or twisting an eclipsed triangular prism, and progressive symmetry breaking by applying successive linear maps to the octahedron along the subgroup chain in \cref{fig:sym-breaking}b. To compare deformation modes, we use a $30$-point $\mathrm{C}_6$ ring and normalize displacement fields to matched RMSD after removing rigid-body components.

We independently verify the stabilizer of each frame (\cref{sec:appendix-deformation-protocol}) and identify spectral onset using degree-resolved relative tail energy, excluding round-off blocks (\cref{sec:appendix-gnn-responses}). For the HEGNN node-type control, we assign distinct types to the five node orbits, preserving the undeformed symmetry, and hold these assignments fixed across deformations (\cref{sec:appendix-gnn-responses}).

\subsubsection*{Biaryl torsion scans}

We use deposited geometries and energies from \texttt{OpenFF Rowley Biaryl v1.0}\citep{smith2021molssi,qiu2020driving}, comprising $87$ biaryls sampled at $15^\circ$ intervals. All atoms are retained in fully connected graphs. Nontrivial point groups identified by \texttt{pymatgen}\citep{ong2013python} along each scan serve as target symmetries. Excluding molecule--target pairs that preserve the target symmetry throughout the scan yields $94$ pairs from $86$ molecules (\cref{sec:appendix-torsion-protocol}).

For centrosymmetric targets, fixed subspaces vanish at odd natural-parity degrees, causing odd-degree LS to saturate near unity when block energies substantially exceed the regularization scale. We therefore pool degrees $\{0,2,4,6\}$ across all targets. Baseline rankings use Engel score or the negated CSoM deviation, with target orientations optimized for both methods as described in \cref{sec:appendix-torsion-protocol,sec:appendix-orientation-search}. AUCs are averaged first across targets within each molecule and then equally across molecules. For target-symmetry comparisons, extrema are classified by whether they possess the target symmetry, while all conformations remain in each ranking (\cref{sec:appendix-torsion-recall}).

Pretraining uses a fixed split of $69$ training and $17$ validation molecules. Total-energy pretraining minimizes mean absolute error on energies standardized using training-set statistics, while Uni-Mol-style pretraining\citep{zhou2023unimol} combines masked-atom prediction, coordinate recovery, and pair-distance prediction. We compare each pretrained model against its paired initialization and report pLS gap improvements separately for the full cohort and validation molecules (\cref{sec:appendix-torsion-pretraining}).

\subsubsection*{Lanthanide DOTA complexes}

We use seven water-free \ce{[Ln(DOTA)]-} crystal structures (\ce{Y} and \ce{Tb} through \ce{Yb})\citep{manvell2024lndota}, together with water-capped \ce{Dy(DOTA)} and distorted \ce{Tm(DOTA)} structures from the CSoM package\citep{nielsen2025determining}. Metal-centred clusters are reconstructed excluding counterions and lattice solvent. Cross-family comparisons use heavy atoms, whereas illustrative spectral analyses retain hydrogens. For the capping-water trajectory, only the two water protons are rotated about the Dy--O axis, with the coordination cage held fixed. We evaluate degree-resolved LS for crystal-field analysis and pLS pooled over degrees $0,\ldots,6$ for water rotation (\cref{sec:appendix-dota-protocol}).

\subsubsection*{Biological assemblies}

The p97, HA, and GluK2 analyses use backbone atoms with pLS pooled over degrees $0,\ldots,6$. Each assembly or region is independently centred and represented by a nearest-neighbour graph. E18 uses sequence-matched C$\alpha$ atoms and degree-six pLS.

For p97, we retain the six p97 chains from UBXD1-associated reconstructions\citep{braxton2023p97}, match complete backbone atom identities across chains and states, and evaluate assemblies, domains, and corresponding sliding windows against $\mathrm{C}_6$ (\cref{sec:appendix-p97-localization,sec:appendix-regional-scales}).

For HA\citep{benton2020structural}, shared residue windows define HA1, HA2, and their union, each evaluated against $\mathrm{C}_3$ (\cref{sec:appendix-ha-protocol}). GluK2\citep{meyerson2016structural} is evaluated against $\mathrm{C}_4$ by domain and by non-overlapping layers along the axis connecting the transmembrane and amino-terminal C$\alpha$ centroids (\cref{sec:appendix-gluk2-protocol}). For E18\citep{buchta2019enterovirus}, we retain all capsid-protein copies in each deposited assembly and evaluate the assembly against the proper icosahedral group $\mathrm{I}$ (\cref{sec:appendix-e18-protocol}).

\subsubsection*{Computational cost}

Timings include graph construction, representation computation, and spectral scoring, as well as region selection for the p97 window scan and orientation search for fitted baselines. For multiple-target evaluations on Dy-DOTA-water, target-independent computations are reused within each method's call (\cref{sec:appendix-computational-cost}).

\section*{Data availability}

The experimental structures and energy data are obtained from the public sources listed below. The accompanying repository provides the synthetic point clouds and dataset splits, the molecular and biological coordinate selections, and numerical source data for the reported analyses.

Three families of synthetic point clouds appear in
\cref{sec:exp-synthetic,sec:exp-sym-breaking}: the $32$-group
benchmark, the controlled symmetry-breaking families and the matched-amplitude displacement fields.
These datasets are generated deterministically using the constructions specified in Methods and \cref{sec:appendix-synthetic-protocol,sec:appendix-deformation-protocol}.

The torsion study of \cref{sec:exp-torsion} uses the relaxed scans of \texttt{OpenFF Rowley Biaryl
v1.0}, obtained through QCArchive\citep{smith2021molssi}, together with their deposited energies.

The study of lanthanide DOTA complexes in \cref{sec:exp-dota} uses the water-free \ce{Ln(DOTA)} series deposited
as CCDC \texttt{2281329} and \texttt{2281615}--\texttt{2281620}; it also uses the two approximate
analogues and the capping-water trajectory, as redistributed with the CSoM
package\citep{nielsen2025determining}.

The assemblies are Protein Data Bank entries. The p97 regional analysis in \cref{sec:exp-protein-assemblies} uses
\texttt{8FCL}, \texttt{8FCM}, \texttt{8FCR}, \texttt{8FCQ}, \texttt{8FCO} and \texttt{8FCT}.
For
haemagglutinin in \cref{sec:appendix-ha}, the series is \texttt{6Y5G}, \texttt{6Y5H}, \texttt{6Y5I},
\texttt{6Y5J} and \texttt{6Y5K}. The GluK2 domain and spatial-layer analyses use \texttt{5KUF}; the echovirus 18 capsid comparison uses \texttt{6HBH}, \texttt{6HBK}, \texttt{6HHT} and \texttt{6HBL}. Atom selections, complete spectra and input-setting results accompany the analysis code. Computational benchmark records include the timing protocol, hardware settings, per-run measurements and input hashes.

\section*{Code availability}

The SymFP implementation, training code and evaluation scripts are
released at \url{https://github.com/GLAD-RUC/SymFP}. The point-group analysis and space-group
expansion use \texttt{pymatgen}\citep{ong2013python}; implementation details for the equivariant GNNs are provided in the accompanying repository. HEGNN uses the multi-channel extension developed here, which is included in the repository. CSoM is evaluated using the software released by its authors\citep{nielsen2025determining}; Engel score is our implementation of the published construction\citep{engel2021point,mbah2023early}.

\section*{Author contributions}

J.C. proposed and developed the theoretical and algorithmic framework, designed and analysed the experiments, and drafted the manuscript under the supervision of W.H.; A.L. and N.L. contributed throughout these stages; Z.W. and J.-R.W. provided technical support; all authors reviewed and approved the final version.

\section*{Competing interests}

The authors declare no competing interests.

\putbib[Reference]
\clearpage
\clearpage
\section*{Extended Data}
\phantomsection
\addcontentsline{toc}{section}{Extended Data}

\begin{center}
    \includegraphics[width=\textwidth]{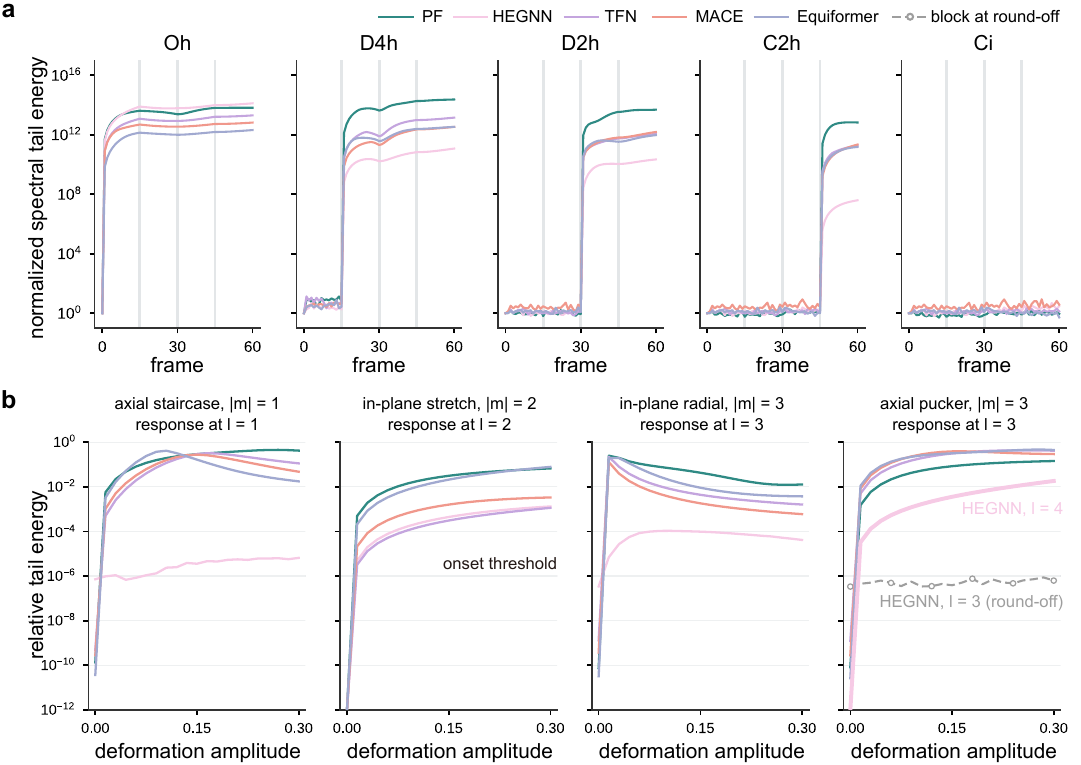}
\end{center}
\extdatalegend{edfig:cross-gnn}{\figtitle{Spectral responses of PF and four learnable equivariant GNNs at random initialization.} \textbf{a}, Progressive symmetry breaking along a subgroup chain. Curves show the spectral tail energy relative to each target group in $\mathrm{O_h}\supset\mathrm{D_{4h}}\supset\mathrm{D_{2h}}\supset\mathrm{C_{2h}}\supset\mathrm{C_i}$, normalized by the corresponding undeformed tail energy. Grey vertical lines indicate the boundaries between deformation segments. \textbf{b}, Deformation-mode comparison. Curves show the relative tail energy at $l=|m|$ as a function of deformation amplitude for four displacement fields with matched RMSD, where $|m|$ denotes the lowest non-zero azimuthal order of each field. Blocks classified as round-off by \cref{eq:roundoff-floor} are shown as grey dashed traces with open markers. For HEGNN, the axial-pucker panel additionally shows the first energy-supported response at $l=4$. The horizontal dashed line indicates the onset threshold in relative tail energy, whereas the round-off criterion is based on absolute block energy. Onset degrees are summarized in \cref{tab:mode-onsets}.}

\clearpage
\begin{center}
    \includegraphics[width=\textwidth]{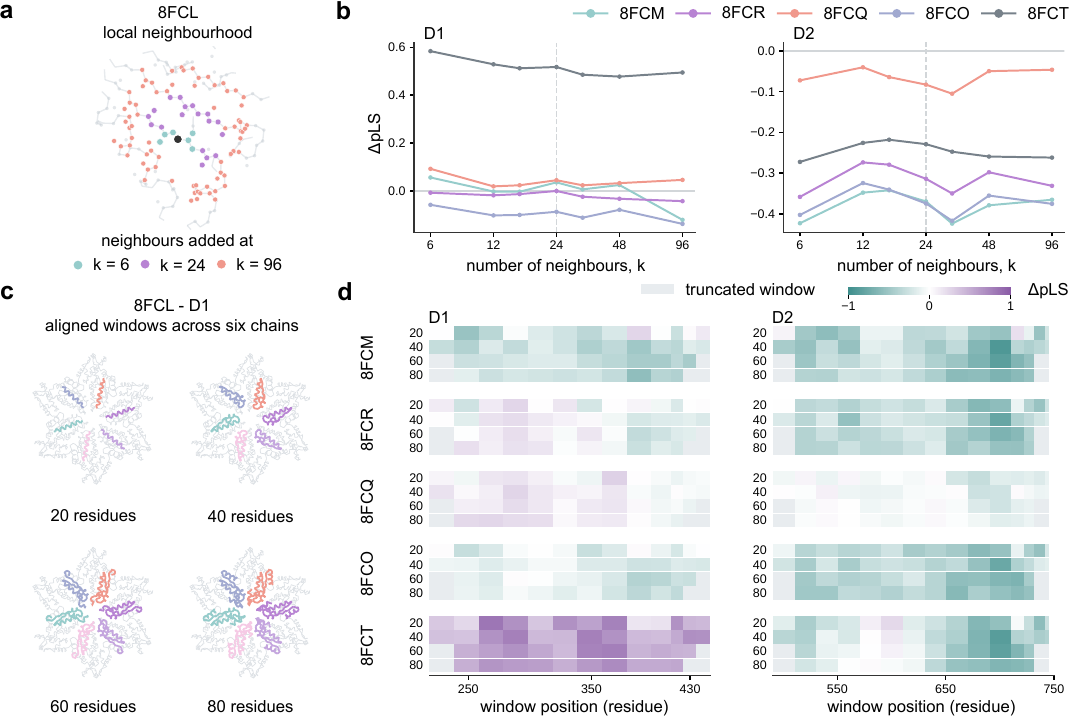}
\end{center}
\extdatalegend{edfig:assembly-scale}{\figtitle{Graph-neighbourhood and window-scale checks for p97.} All scores use the hybrid PF, degrees $0,\ldots,6$ and target $\mathrm{C}_6$. Differences are relative to closed \texttt{8FCL}. \textbf{a}, Local backbone around the C$\alpha$ atom of chain A, residue $269$, in \texttt{8FCL} D1. Colours distinguish neighbours included at $k=6$ and those added at $k=24$ and $96$, selected by three-dimensional distance. \textbf{b}, Whole-domain D1 (left) and D2 (right) changes in the negative-log leakage score (pLS) across neighbourhood sizes for the other five structures in \cref{fig:p97}; dashed lines mark $k=24$. \textbf{c}, Example D1 selections of $20$, $40$, $60$ and $80$ residues per chain, at a common centre. Coloured traces show the six corresponding segments against the complete D1 background, using the same view and scale. \textbf{d}, Regional D1 (left) and D2 (right) pLS changes at $k=24$, with four sequence-window lengths evaluated at common centres. Grey cells mark windows truncated at domain boundaries; each window retains the resolved common residues within its sequence interval. The opposing D1 and D2 responses associated with the UBXD1 lariat mutant persist across the neighbourhood sizes in \textbf{b} and window lengths in \textbf{d}, while shorter windows resolve more local variation. Selection and coverage details are given in \cref{sec:appendix-regional-scales}.}

\clearpage
\begin{center}
    \includegraphics[width=\textwidth]{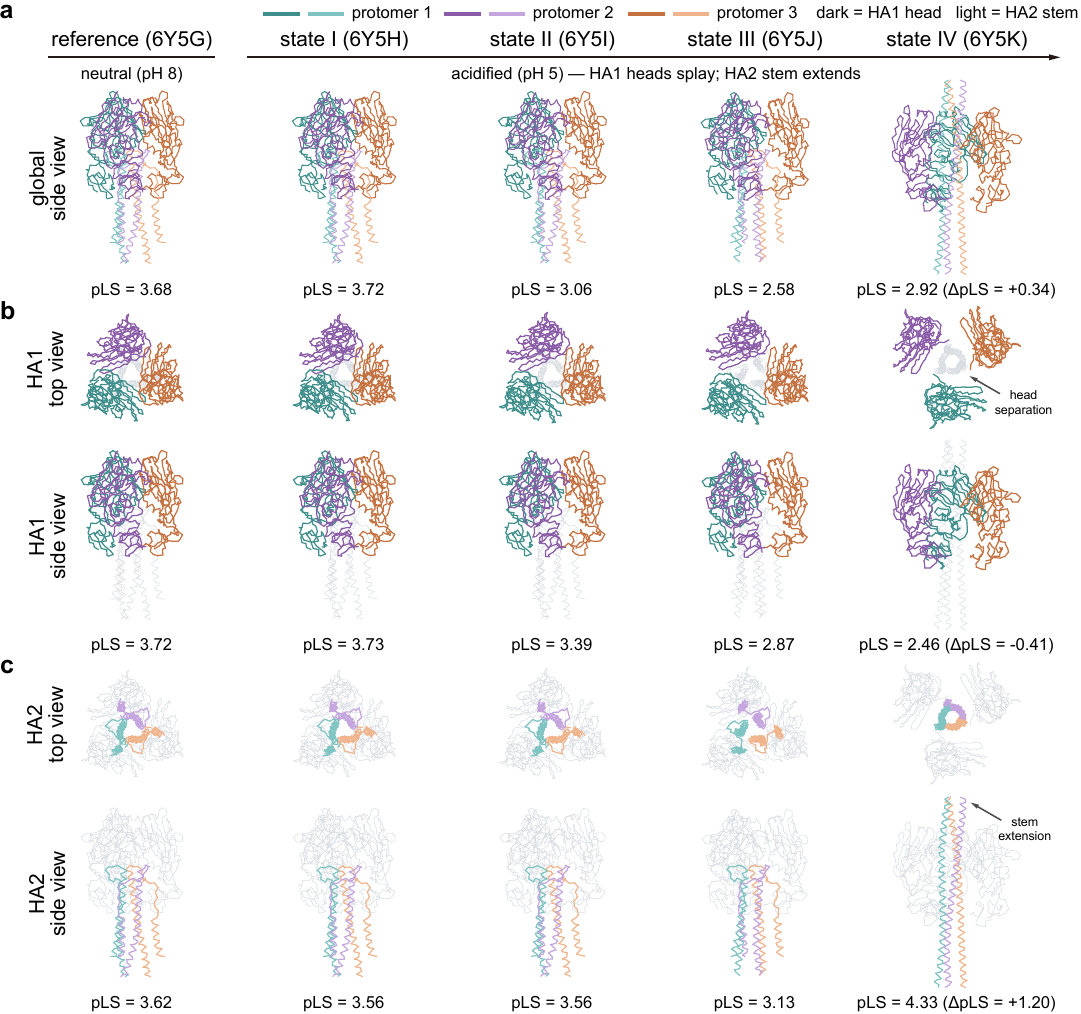}
\end{center}
\extdatalegend{edfig:ha}{\figtitle{Regional symmetry responses in haemagglutinin.} Five experimental structures\citep{benton2020structural}, comprising the neutral reference and acidified states I--IV (\texttt{6Y5G}--\texttt{6Y5K}), shown at a common scale. \textbf{a}, Global side views. \textbf{b,c}, Top and side views highlighting HA1 (\textbf{b}) and HA2 (\textbf{c}), with the other region in grey. Colours identify protomers; darker and lighter shades distinguish HA1 and HA2. Values give the negative-log leakage score (pLS) against $\mathrm{C}_3$ for each selection; $\Delta\mathrm{pLS}$ is state IV minus state III. From state III to IV, global pLS increases, while HA1 pLS decreases and HA2 pLS increases, revealing opposing regional responses. All scores use the hybrid PF, degrees $0,\ldots,6$ and backbone graphs with $k=24$; traces show the corresponding C$\alpha$ atoms. The reconstruction constraints are $\mathrm{C}_3$, $\mathrm{C}_3$, $\mathrm{C}_1$, $\mathrm{C}_1$ and $\mathrm{C}_3$, respectively. Selection details are given in \cref{sec:appendix-ha-protocol}.}

\clearpage
\begin{center}
    \includegraphics[width=\textwidth]{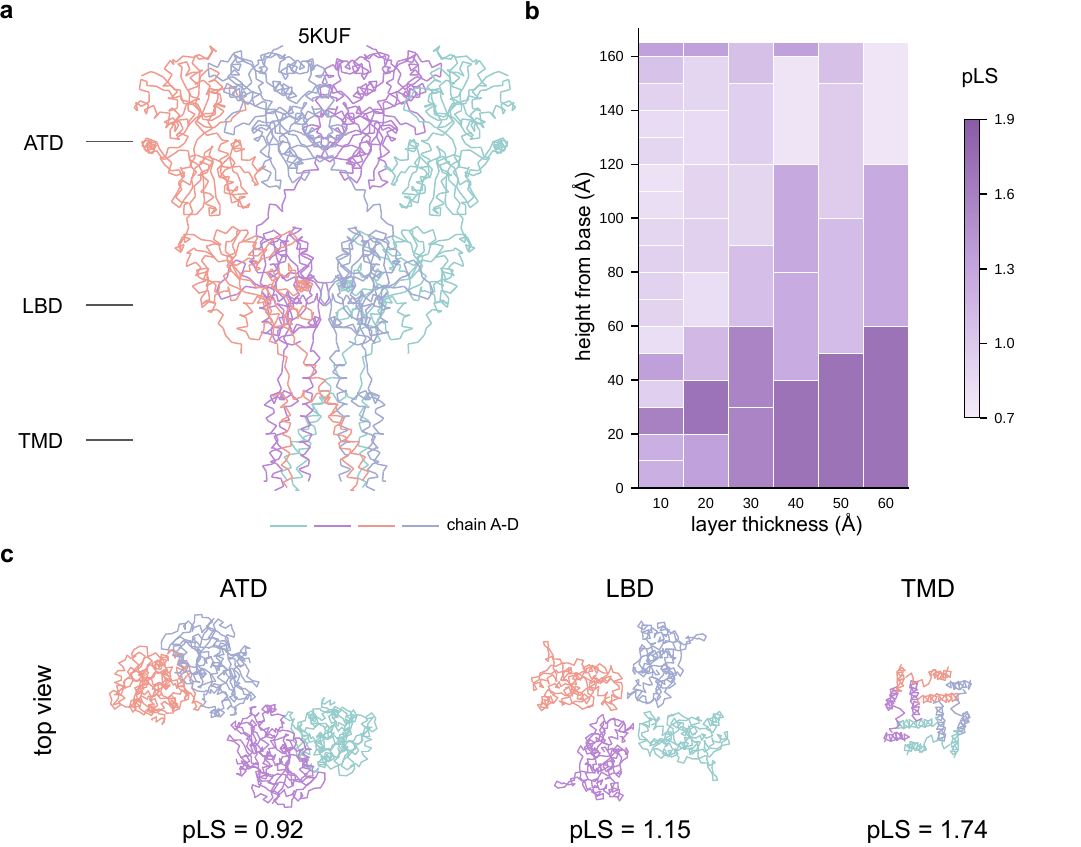}
\end{center}
\extdatalegend{edfig:gluk2}{\figtitle{Spatial symmetry organization in GluK2.} \textbf{a}, Side view of the tetramer (\texttt{5KUF})\citep{meyerson2016structural}, locating the amino-terminal (ATD), ligand-binding (LBD) and transmembrane (TMD) domains. \textbf{b}, Negative-log leakage scores (pLS) against $\mathrm{C}_4$ for non-overlapping layers along the receptor axis, using thicknesses of $10$--$60$\,\AA. Blocks show the actual layer extents, aligned with \textbf{a}, with height measured from the lowest selected C$\alpha$ position and each column's shorter top remainder scored separately. The transmembrane region generally has higher pLS than the upper extracellular regions, a spatial pattern retained across the tested layer thicknesses. \textbf{c}, Top views and $\mathrm{C}_4$ pLS of the three complete domain selections, shown at a common scale. Chains A--D retain the same colours across views. Scores use independently represented backbone selections with the hybrid PF, degrees $0,\ldots,6$ and $k=24$; traces show C$\alpha$ atoms. The reconstruction imposes $\mathrm{C}_2$. Selection details are given in \cref{sec:appendix-gluk2-protocol}.}

\clearpage

\begin{center}
    \includegraphics[width=\textwidth]{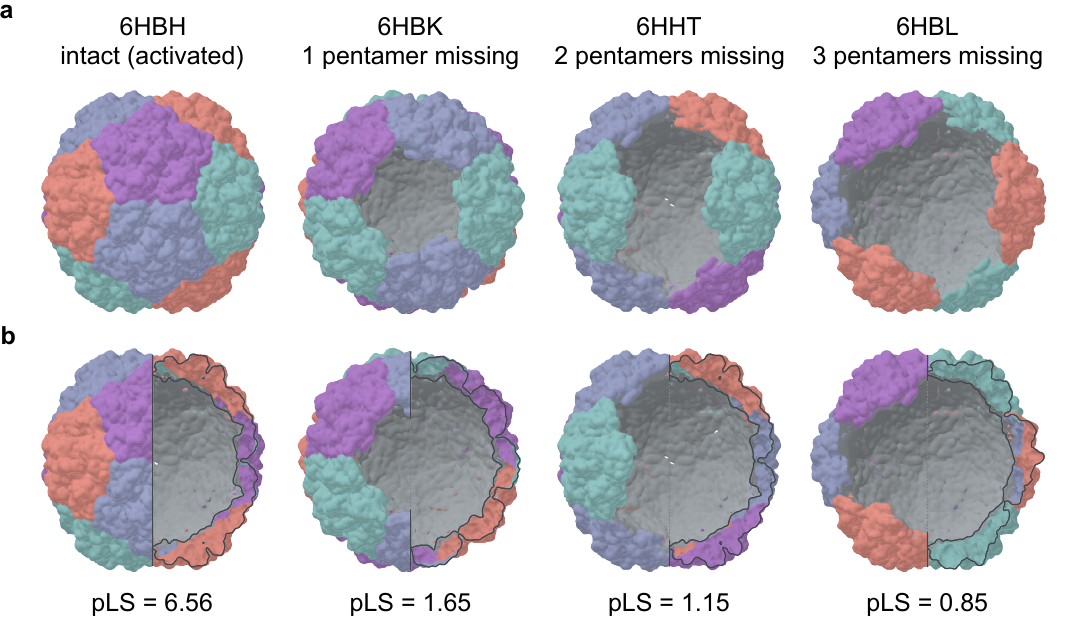}
\end{center}
\extdatalegend{edfig:e18-capsid}{\figtitle{Icosahedral target agreement across echovirus 18 capsid structures.} \textbf{a}, Deposited models of an intact activated E18 capsid and capsids lacking one, two or three pentamers\citep{buchta2019enterovirus}. Surfaces are envelopes of the sequence-matched C$\alpha$ selections used for scoring. Colours distinguish adjacent pentamers, with corresponding positions assigned consistently after structural alignment. \textbf{b}, Cutaway views of the same models, with cavity-facing surfaces shaded grey. All views share a physical scale. The negative-log leakage scores (pLS) are evaluated at degree $l=6$ for the proper icosahedral target group $\mathrm{I}$, using the complete selected coordinates. From the intact capsid to structures lacking one, two or three pentamers, pLS decreases with the number of missing pentamers. Scores use the hybrid PF with $k=24$. The four structural classes were reconstructed with $\mathrm{I}$, $\mathrm{C}_5$, $\mathrm{C}_2$ and $\mathrm{C}_3$ constraints, respectively. Selection details are given in \cref{sec:appendix-e18-protocol}.}

\clearpage

\begin{table}[p]
\centering
\extdatatablelegend{edtab:torsion-symmetry-groups}{Energy-extremum retrieval and dataset composition by target symmetry. \textbf{a}, Mean AUC, recall at a $25\%$ budget, and molecule-level AUC wins/ties/losses (W/T/L) for pLS against each baseline. Bold marks row-wise maxima for each metric. \textbf{b}, Extrema counts per molecule--target pair, with percentages within each extremum type. Target symmetry requires the point group assigned at tolerance $0.1$ to contain the target. Rankings retain all $24$ conformations; averaging follows \cref{sec:appendix-torsion-recall}.}
\smallskip
\fontsize{9}{10.5}\selectfont
\setlength{\tabcolsep}{3pt}
\renewcommand{\arraystretch}{1.12}
\begin{tabular}{>{\raggedright\arraybackslash}p{\dimexpr0.12\linewidth-2\tabcolsep\relax}>{\raggedright\arraybackslash}p{\dimexpr0.14\linewidth-2\tabcolsep\relax}*{6}{>{\raggedleft\arraybackslash}p{\dimexpr0.085\linewidth-2\tabcolsep\relax}}*{2}{>{\centering\arraybackslash}p{\dimexpr0.115\linewidth-2\tabcolsep\relax}}}
\toprule
\multicolumn{10}{l}{\textbf{a}\quad Retrieval performance} \\
\midrule
 & & \multicolumn{3}{c}{AUC} & \multicolumn{3}{c}{Recall at 25\% budget (\%)} & \multicolumn{2}{c}{pLS W/T/L} \\
\cmidrule(lr){3-5}\cmidrule(lr){6-8}\cmidrule(lr){9-10}
Group & Target symmetry & pLS & CSoM & Engel score & pLS & CSoM & Engel score & vs CSoM & vs Engel \\
\midrule
Overall & --- & $\mathbf{0.794}$ & $0.747$ & $0.787$ & $\mathbf{68.8}$ & $66.9$ & $66.3$ & 41/36/9 & 23/53/10 \\
\addlinespace[2pt]
All & w/ & $\mathbf{0.927}$ & $0.920$ & $0.915$ & $\mathbf{99.4}$ & $98.3$ & $95.3$ & 4/81/1 & 13/72/1 \\
All & w/o & $0.593$ & $0.475$ & $\mathbf{0.597}$ & $19.5$ & $16.9$ & $\mathbf{20.0}$ & 40/10/9 & 12/34/13 \\
\addlinespace[2pt]
Maxima & w/ & $\mathbf{0.916}$ & $0.914$ & $0.900$ & $\mathbf{100.0}$ & $99.0$ & $94.4$ & 3/70/7 & 14/65/1 \\
Maxima & w/o & $\mathbf{0.706}$ & $0.601$ & $0.627$ & $\mathbf{38.9}$ & $25.0$ & $33.3$ & 13/6/8 & 11/10/6 \\
\addlinespace[2pt]
Minima & w/ & $\mathbf{0.926}$ & $0.908$ & $0.924$ & $\mathbf{93.8}$ & $93.1$ & $\mathbf{93.8}$ & 9/29/2 & 4/31/5 \\
Minima & w/o & $0.567$ & $0.454$ & $\mathbf{0.591}$ & $17.9$ & $16.5$ & $\mathbf{19.8}$ & 33/13/7 & 7/29/17 \\
\bottomrule
\end{tabular}
\par\medskip
\begin{tabular}{>{\raggedright\arraybackslash}p{\dimexpr0.12\linewidth-2\tabcolsep\relax}>{\raggedright\arraybackslash}p{\dimexpr0.14\linewidth-2\tabcolsep\relax}>{\raggedleft\arraybackslash}p{\dimexpr0.19\linewidth-2\tabcolsep\relax}>{\raggedleft\arraybackslash}p{\dimexpr0.28\linewidth-2\tabcolsep\relax}>{\raggedleft\arraybackslash}p{\dimexpr0.27\linewidth-2\tabcolsep\relax}}
\toprule
\multicolumn{5}{l}{\textbf{b}\quad Dataset composition} \\
\midrule
Group & Target symmetry & Molecules & Molecule--target pairs & Extrema, $n$ (\%) \\
\midrule
Overall & --- & 86 & 94 & 604 (100.0) \\
\addlinespace[2pt]
All & w/ & 86 & 94 & 333 (55.1) \\
All & w/o & 59 & 61 & 271 (44.9) \\
\addlinespace[2pt]
Maxima & w/ & 80 & 88 & 245 (81.1) \\
Maxima & w/o & 27 & 29 & 57 (18.9) \\
\addlinespace[2pt]
Minima & w/ & 40 & 42 & 88 (29.1) \\
Minima & w/o & 53 & 55 & 214 (70.9) \\
\bottomrule
\end{tabular}
\end{table}

\clearpage
\begin{table}[p]
\centering
\extdatatablelegend{edtab:torsion-pretraining}{Effects of pretraining on pLS and energy-extremum retrieval. Changes are relative to paired random initializations. \textbf{a}, Mean maxima-minus-minima pLS gap. \textbf{b}, Mean target-present-minus-absent pLS gap, with target presence defined by point-group containment. Changes in raw pLS and AUC are averaged over targets within molecules, then equally over all $86$ molecules. Improved molecules have increased gaps. Bold marks the largest gap increases and improvement proportions within each panel and pretraining group.}
\smallskip
\fontsize{9}{10.5}\selectfont
\setlength{\tabcolsep}{3pt}
\renewcommand{\arraystretch}{1.10}
\begin{tabular}{>{\raggedright\arraybackslash}p{\dimexpr0.14\linewidth-2\tabcolsep\relax}>{\raggedleft\arraybackslash}p{\dimexpr0.1\linewidth-2\tabcolsep\relax}>{\raggedleft\arraybackslash}p{\dimexpr0.1\linewidth-2\tabcolsep\relax}>{\raggedleft\arraybackslash}p{\dimexpr0.1\linewidth-2\tabcolsep\relax}>{\raggedleft\arraybackslash}p{\dimexpr0.21\linewidth-2\tabcolsep\relax}>{\raggedleft\arraybackslash}p{\dimexpr0.24\linewidth-2\tabcolsep\relax}>{\raggedleft\arraybackslash}p{\dimexpr0.11\linewidth-2\tabcolsep\relax}}
\toprule
\multicolumn{7}{l}{\textbf{a}\quad Energy extrema} \\
\midrule
GNN & \multicolumn{3}{c}{Change in pLS} & \multicolumn{2}{c}{Improved molecules} & $\Delta$AUC \\
\cmidrule(lr){2-4}\cmidrule(lr){5-6}
 & Maxima & Minima & Gap & All ($n=86$) & Validation ($n=17$) & \\
\midrule
\multicolumn{7}{l}{\textbf{Total-energy pretraining}} \\
\addlinespace[2pt]
HEGNN & $-0.4718$ & $-0.7146$ & $+0.2428$ & 54/86 (62.8\%) & 10/17 (58.8\%) & $+0.0061$ \\
TFN & $-0.4048$ & $-0.4192$ & $+0.0143$ & 45/86 (52.3\%) & 9/17 (52.9\%) & $-0.0088$ \\
MACE & $-1.5250$ & $-1.6448$ & $+0.1198$ & 56/86 (65.1\%) & 9/17 (52.9\%) & $+0.0006$ \\
Equiformer & $-0.8038$ & $-1.2372$ & $\mathbf{+0.4334}$ & \textbf{67/86 (77.9\%)} & \textbf{15/17 (88.2\%)} & $-0.0069$ \\
\midrule
\multicolumn{7}{l}{\textbf{Uni-Mol-style pretraining}} \\
\addlinespace[2pt]
HEGNN & $-1.9679$ & $-1.9988$ & $+0.0308$ & 49/86 (57.0\%) & 9/17 (52.9\%) & $+0.0057$ \\
TFN & $-0.7933$ & $-0.9071$ & $+0.1137$ & 62/86 (72.1\%) & 10/17 (58.8\%) & $-0.0076$ \\
MACE & $-2.0026$ & $-2.1744$ & $+0.1718$ & 60/86 (69.8\%) & 10/17 (58.8\%) & $+0.0099$ \\
Equiformer & $-0.0465$ & $-0.5154$ & $\mathbf{+0.4689}$ & \textbf{66/86 (76.7\%)} & \textbf{15/17 (88.2\%)} & $+0.0008$ \\
\bottomrule
\end{tabular}
\par\medskip
\begin{tabular}{>{\raggedright\arraybackslash}p{\dimexpr0.14\linewidth-2\tabcolsep\relax}>{\raggedleft\arraybackslash}p{\dimexpr0.15\linewidth-2\tabcolsep\relax}>{\raggedleft\arraybackslash}p{\dimexpr0.15\linewidth-2\tabcolsep\relax}>{\raggedleft\arraybackslash}p{\dimexpr0.11\linewidth-2\tabcolsep\relax}>{\raggedleft\arraybackslash}p{\dimexpr0.21\linewidth-2\tabcolsep\relax}>{\raggedleft\arraybackslash}p{\dimexpr0.24\linewidth-2\tabcolsep\relax}}
\toprule
\multicolumn{6}{l}{\textbf{b}\quad Target symmetry} \\
\midrule
GNN & \multicolumn{3}{c}{Change in pLS} & \multicolumn{2}{c}{Improved molecules} \\
\cmidrule(lr){2-4}\cmidrule(lr){5-6}
 & Target present & Target absent & Gap & All ($n=86$) & Validation ($n=17$) \\
\midrule
\multicolumn{6}{l}{\textbf{Total-energy pretraining}} \\
\addlinespace[2pt]
HEGNN & $-0.3695$ & $-0.9970$ & $\mathbf{+0.6275}$ & \textbf{81/86 (94.2\%)} & \textbf{17/17 (100.0\%)} \\
TFN & $-0.3463$ & $-0.4705$ & $+0.1243$ & 61/86 (70.9\%) & 11/17 (64.7\%) \\
MACE & $-1.4720$ & $-1.7944$ & $+0.3224$ & 71/86 (82.6\%) & 15/17 (88.2\%) \\
Equiformer & $-0.8573$ & $-1.2538$ & $+0.3964$ & 67/86 (77.9\%) & 14/17 (82.4\%) \\
\midrule
\multicolumn{6}{l}{\textbf{Uni-Mol-style pretraining}} \\
\addlinespace[2pt]
HEGNN & $-1.6033$ & $-2.6896$ & $\mathbf{+1.0864}$ & \textbf{81/86 (94.2\%)} & \textbf{17/17 (100.0\%)} \\
TFN & $-0.7853$ & $-0.9514$ & $+0.1661$ & 64/86 (74.4\%) & 14/17 (82.4\%) \\
MACE & $-1.9376$ & $-2.3959$ & $+0.4583$ & 76/86 (88.4\%) & 15/17 (88.2\%) \\
Equiformer & $-0.1030$ & $-0.5324$ & $+0.4294$ & 69/86 (80.2\%) & 14/17 (82.4\%) \\
\bottomrule
\end{tabular}

\end{table}

\clearpage
\begin{table}[p]
\centering
\extdatatablelegend{edtab:computational-cost}{Computational cost of structural symmetry analysis. \textbf{a}, Multi-target CPU timings for Dy-DOTA-water. \textbf{b}, CPU comparison on biological structures. \textbf{c}, SymFP CPU/GPU timings for whole structures and the p97 regional scan. CPU measurements use one thread. Times are medians except for single-trial biological coordinate fitting; $\dagger$ denotes a $7200$\,s timeout. Engel BFGS optimizes retained energy; BFGS + CSoM adds a coordinate-fitted start. Protocols and repetition counts are given in \cref{sec:appendix-computational-cost}.}
\smallskip
\footnotesize
\setlength{\tabcolsep}{5pt}
\renewcommand{\arraystretch}{1.12}
\begin{tabular}{>{\raggedright\arraybackslash}p{\dimexpr0.40\linewidth-2\tabcolsep\relax}*{4}{>{\raggedleft\arraybackslash}p{\dimexpr0.15\linewidth-2\tabcolsep\relax}}}
\toprule
\multicolumn{5}{l}{\textbf{a}\quad Dy-DOTA-water (30 heavy atoms): CPU total time (ms)} \\
\multicolumn{5}{p{\dimexpr\linewidth-2\tabcolsep\relax}}{Multiple targets share computations within each call; baseline timings include orientation search.} \\
\midrule
Method & 1 target & 2 targets & 4 targets & 7 targets \\
 & $\mathrm{C}_2$ & $\mathrm{C}_2,\mathrm{C}_4$ & $\mathrm{C}_2,\mathrm{C}_4,$ $\mathrm{C}_6,\mathrm{D}_2$ & $\mathrm{C}_2,\mathrm{C}_4,\mathrm{C}_6,$ $\mathrm{D}_2,\mathrm{D}_4,\mathrm{D}_6,\mathrm{O}$ \\
\midrule
SymFP & 6.46 & 6.59 & 6.87 & 7.21 \\
CSoM & 1575.40 & 5605.46 & 15901.44 & 96382.65 \\
Engel, grid search & 611.77 & 619.24 & 620.43 & 627.62 \\
Engel, BFGS & 297.45 & 499.16 & 975.75 & 1861.96 \\
Engel, BFGS + CSoM & 338.02 & 586.11 & 1454.12 & 7211.64 \\
\bottomrule
\end{tabular}
\par\smallskip
\begin{tabular}{>{\raggedright\arraybackslash}p{\dimexpr0.52\linewidth-2\tabcolsep\relax}>{\raggedleft\arraybackslash}p{\dimexpr0.20\linewidth-2\tabcolsep\relax}>{\raggedleft\arraybackslash}p{\dimexpr0.28\linewidth-2\tabcolsep\relax}}
\toprule
\multicolumn{3}{l}{\textbf{b}\quad Structural-analysis CPU total time (s)} \\
\midrule
Method & HA global ($\mathrm{C}_3$) & p97 common backbone ($\mathrm{C}_6$) \\
\midrule
SymFP & 1.631 & 9.75 \\
CSoM & $\geq7200^{\dagger}$ & $\geq7200^{\dagger}$ \\
Engel, grid search & 16.175 & 244.378 \\
Engel, BFGS & 15.633 & 242.963 \\
Engel, BFGS + CSoM & 907.391 & $\geq7200^{\dagger}$ \\
\bottomrule
\end{tabular}
\par\smallskip
\begin{tabular}{>{\raggedright\arraybackslash}p{\dimexpr0.40\linewidth-2\tabcolsep\relax}*{4}{>{\raggedleft\arraybackslash}p{\dimexpr0.15\linewidth-2\tabcolsep\relax}}}
\toprule
\multicolumn{5}{l}{\textbf{c}\quad SymFP total time (s)} \\
\multicolumn{5}{p{\dimexpr\linewidth-2\tabcolsep\relax}}{Atom and directed-edge ranges are per window. GPU totals include graph construction and data transfer.} \\
\midrule
Selection & Atoms & Edges & CPU & GPU \\
\midrule
HA global & 4,248 & 101,952 & 1.631 & 0.229 \\
p97 common backbone & 17,088 & 410,113 & 9.75 & 4.14 \\
p97 regional scan (156 windows) & 480--960 & 11,520--23,041 & 47.66 & 4.16 \\
\bottomrule
\end{tabular}
\end{table}

\clearpage
\end{bibunit}

\appendix
\begin{bibunit}[unsrtnat]
\setappendix

\section{Theoretical analysis}
\label{sec:appendix-theory}

The following results establish the point-group rank templates, spectral invariance and subgroup ordering. We then relate full-span attainability to generic rank realization and quantify spectral departures under coordinate perturbations within a fixed graph family.

\subsection{Point-group rank templates}
\label{sec:appendix-rank-templates}

\restateintheappendix{\FixedSubspaceConstraint}
\begin{proof}
For $\gG\in\sG(\GH)$, equivariance gives $\smVl(\Gh\cdot\gG)=\rho^{(l)}(\Gh)\smVlgG=\smVlgG$ for every $\Gh\in\GH$. Define the group average $\rho^{(l)}(\GH)\coloneqq 1/|\GH|\cdot\sum_{\Gh\in\GH}\rho^{(l)}(\Gh)$. Averaging the equivariance identity yields
\begin{equation}
\label{eq:group-average-invariance}
    \smVlgG=\rho^{(l)}(\GH)\smVlgG.
\end{equation}
Hence every column of $\smVlgG$ lies in $\operatorname{Im}(\rho^{(l)}(\GH))$, so
\begin{equation}
    \operatorname{colspan}(\smVlgG)
    \subseteq
    \operatorname{Im}(\rho^{(l)}(\GH)).
\end{equation}

It remains to identify this image with the fixed subspace. If $\svv^{(l)}=\rho^{(l)}(\GH)\svu$ for some $\svu\in\R^{2l+1}$, then for any $\Gh_0\in\GH$,
\begin{equation}
    \rho^{(l)}(\Gh_0)\svv^{(l)}
    =
    \frac{1}{|\GH|}\sum_{\Gh\in\GH}\rho^{(l)}(\Gh_0\Gh)\svu
    =
    \rho^{(l)}(\GH)\svu
    =
    \svv^{(l)},
\end{equation}
since $\Gh_0\GH=\GH$. Thus $\svv^{(l)}\in\sF^{(l)}(\GH)$. Conversely, if $\svv^{(l)}\in\sF^{(l)}(\GH)$, then
\begin{equation}
    \rho^{(l)}(\GH)\svv^{(l)}
    =
    \frac{1}{|\GH|}\sum_{\Gh\in\GH}\rho^{(l)}(\Gh)\svv^{(l)}
    =
    \svv^{(l)},
\end{equation}
so $\svv^{(l)}\in\operatorname{Im}(\rho^{(l)}(\GH))$.

Therefore $\operatorname{Im}(\rho^{(l)}(\GH))=\sF^{(l)}(\GH)$. The same averaging argument yields
\begin{equation}
    (\rho^{(l)}(\GH))^2=\rho^{(l)}(\GH),
\end{equation}
so $\rho^{(l)}(\GH)$ is an idempotent projection. Moreover, each $\rho^{(l)}(\Gh)$ is orthogonal, hence
\begin{equation}
    (\rho^{(l)}(\GH))^\top
    =
    \frac{1}{|\GH|}\sum_{\Gh\in\GH}\rho^{(l)}(\Gh)^\top
    =
    \frac{1}{|\GH|}\sum_{\Gh\in\GH}\rho^{(l)}(\Gh^{-1})
    =
    \rho^{(l)}(\GH),
\end{equation}
because $\GH$ is closed under inversion. Thus $\rho^{(l)}(\GH)$ is self-adjoint as well, so it is the orthogonal projector onto $\sF^{(l)}(\GH)$. Hence all its eigenvalues are $0$ or $1$, and therefore
\begin{equation}
    \operatorname{rank}(\rho^{(l)}(\GH))
    =
    \tr(\rho^{(l)}(\GH))
    =
    \dim \sF^{(l)}(\GH).
\end{equation}
The column-space inclusion and the rank inequality in \cref{eq:fixed-subspace-column,eq:fixed-subspace-rank} follow immediately.
\end{proof}

If a finite candidate family $\sC$ has pairwise distinct rank templates over all natural-parity degrees, finitely many degrees suffice to separate it. Indeed, choose one distinguishing degree for each pair of groups; the finite union of these degrees gives a separating set $\Lambda$.

Let $\sC_{32}$ denote the 32 crystallographic point groups in Schoenflies notation\citep{aroyo2016international}:
\begin{equation}
\sC_{32}
=
\left\{
\begin{aligned}
&\mathrm{C}_1,\mathrm{C_i},\mathrm{C}_2,\mathrm{C_s},\mathrm{C_{2h}},\mathrm{D}_2,\mathrm{C_{2v}},\mathrm{D_{2h}},\mathrm{C}_4,\mathrm{S}_4,\mathrm{C_{4h}},\\
&\mathrm{D}_4,\mathrm{C_{4v}},\mathrm{D_{2d}},\mathrm{D_{4h}},\mathrm{C}_3,\mathrm{S}_6,\mathrm{D}_3,\mathrm{C_{3v}},\mathrm{D_{3d}},\mathrm{C}_6,\mathrm{C_{3h}},\\
&\mathrm{C_{6h}},\mathrm{D}_6,\mathrm{C_{6v}},\mathrm{D_{3h}},\mathrm{D_{6h}},\mathrm{T},\mathrm{T_h},\mathrm{O},\mathrm{T_d},\mathrm{O_h}
\end{aligned}
\right\}.
\end{equation}

\restateintheappendix{\CrystallographicSignatureCorollary}
\begin{proof}
For every natural-parity degree $l$, the quantity $m_l(\GH)$ is the multiplicity of the trivial representation in $\rho^{(l)}|_{\GH}$, equivalently the character average $\tr(\rho^{(l)}(\GH))$. Evaluating these multiplicities for the 32 crystallographic point groups gives the natural-parity rows of \cref{tab:crystallographic-rank-templates}. Restricting to degrees up to $9$ leaves the rows pairwise distinct, which proves injectivity for $\Lambda_{9}$; restricting to degrees up to $6$ gives 30 distinct rows, with the two coincidences stated, between $\mathrm{D}_6$ and $\mathrm{D_{6h}}$ and between $\mathrm{O}$ and $\mathrm{O_h}$.
\end{proof}

\begin{table}[!htbp]
\centering
\fontsize{9}{11}\selectfont
\setlength{\tabcolsep}{4pt}
\renewcommand{\arraystretch}{1.15}
\captionsetup{font=normalsize,margin=0pt}
\caption{Fixed-subspace dimensions of crystallographic point groups. Values use natural-parity representations with $l=0,\ldots,11$. The final column gives the crystal-field parameter count $n_{\mathrm{CF}}=\sum_{k\in\{2,4,6\}}m_k$ (\cref{sec:exp-dota}).}
\label{tab:crystallographic-rank-templates}
\begin{tabular}{>{\raggedright\arraybackslash}p{\dimexpr0.12\linewidth-2\tabcolsep\relax}>{\raggedright\arraybackslash}p{\dimexpr0.18\linewidth-2\tabcolsep\relax}>{\centering\arraybackslash}p{\dimexpr0.0525\linewidth-2\tabcolsep\relax}>{\centering\arraybackslash}p{\dimexpr0.0525\linewidth-2\tabcolsep\relax}>{\centering\arraybackslash}p{\dimexpr0.0525\linewidth-2\tabcolsep\relax}>{\centering\arraybackslash}p{\dimexpr0.0525\linewidth-2\tabcolsep\relax}>{\centering\arraybackslash}p{\dimexpr0.0525\linewidth-2\tabcolsep\relax}>{\centering\arraybackslash}p{\dimexpr0.0525\linewidth-2\tabcolsep\relax}>{\centering\arraybackslash}p{\dimexpr0.0525\linewidth-2\tabcolsep\relax}>{\centering\arraybackslash}p{\dimexpr0.0525\linewidth-2\tabcolsep\relax}>{\centering\arraybackslash}p{\dimexpr0.0525\linewidth-2\tabcolsep\relax}>{\centering\arraybackslash}p{\dimexpr0.0525\linewidth-2\tabcolsep\relax}>{\centering\arraybackslash}p{\dimexpr0.0525\linewidth-2\tabcolsep\relax}>{\centering\arraybackslash}p{\dimexpr0.0525\linewidth-2\tabcolsep\relax}>{\centering\arraybackslash}p{\dimexpr0.07\linewidth-2\tabcolsep\relax}}
    \toprule
    \multirow{2}{*}{Schoenflies} & \multirow{2}{*}{Hermann--Mauguin} & \multicolumn{12}{c}{Fixed-subspace dimension $m_l$ (natural parity)} & \multirow{2}{*}{$n_{\mathrm{CF}}$} \\
    \cmidrule(lr){3-14}
    & & $0$ & $1$ & $2$ & $3$ & $4$ & $5$ & $6$ & $7$ & $8$ & $9$ & $10$ & $11$ & \\
    \midrule
    $\mathrm{C}_1$ & $1$ & 1 & 3 & 5 & 7 & 9 & 11 & 13 & 15 & 17 & 19 & 21 & 23 & 27 \\
    $\mathrm{C_i}$ & $-1$ & 1 & 0 & 5 & 0 & 9 & 0 & 13 & 0 & 17 & 0 & 21 & 0 & 27 \\
    $\mathrm{C}_2$ & $2$ & 1 & 1 & 3 & 3 & 5 & 5 & 7 & 7 & 9 & 9 & 11 & 11 & 15 \\
    $\mathrm{C_s}$ & $m$ & 1 & 2 & 3 & 4 & 5 & 6 & 7 & 8 & 9 & 10 & 11 & 12 & 15 \\
    $\mathrm{C_{2h}}$ & $2/m$ & 1 & 0 & 3 & 0 & 5 & 0 & 7 & 0 & 9 & 0 & 11 & 0 & 15 \\
    $\mathrm{D}_2$ & $222$ & 1 & 0 & 2 & 1 & 3 & 2 & 4 & 3 & 5 & 4 & 6 & 5 & 9 \\
    $\mathrm{C_{2v}}$ & $mm2$ & 1 & 1 & 2 & 2 & 3 & 3 & 4 & 4 & 5 & 5 & 6 & 6 & 9 \\
    $\mathrm{D_{2h}}$ & $mmm$ & 1 & 0 & 2 & 0 & 3 & 0 & 4 & 0 & 5 & 0 & 6 & 0 & 9 \\
    $\mathrm{C}_4$ & $4$ & 1 & 1 & 1 & 1 & 3 & 3 & 3 & 3 & 5 & 5 & 5 & 5 & 7 \\
    $\mathrm{S}_4$ & $-4$ & 1 & 0 & 1 & 2 & 3 & 2 & 3 & 4 & 5 & 4 & 5 & 6 & 7 \\
    $\mathrm{C_{4h}}$ & $4/m$ & 1 & 0 & 1 & 0 & 3 & 0 & 3 & 0 & 5 & 0 & 5 & 0 & 7 \\
    $\mathrm{D}_4$ & $422$ & 1 & 0 & 1 & 0 & 2 & 1 & 2 & 1 & 3 & 2 & 3 & 2 & 5 \\
    $\mathrm{C_{4v}}$ & $4mm$ & 1 & 1 & 1 & 1 & 2 & 2 & 2 & 2 & 3 & 3 & 3 & 3 & 5 \\
    $\mathrm{D_{2d}}$ & $-42m$ & 1 & 0 & 1 & 1 & 2 & 1 & 2 & 2 & 3 & 2 & 3 & 3 & 5 \\
    $\mathrm{D_{4h}}$ & $4/mmm$ & 1 & 0 & 1 & 0 & 2 & 0 & 2 & 0 & 3 & 0 & 3 & 0 & 5 \\
    $\mathrm{C}_3$ & $3$ & 1 & 1 & 1 & 3 & 3 & 3 & 5 & 5 & 5 & 7 & 7 & 7 & 9 \\
    $\mathrm{S}_6$ & $-3$ & 1 & 0 & 1 & 0 & 3 & 0 & 5 & 0 & 5 & 0 & 7 & 0 & 9 \\
    $\mathrm{D}_3$ & $32$ & 1 & 0 & 1 & 1 & 2 & 1 & 3 & 2 & 3 & 3 & 4 & 3 & 6 \\
    $\mathrm{C_{3v}}$ & $3m$ & 1 & 1 & 1 & 2 & 2 & 2 & 3 & 3 & 3 & 4 & 4 & 4 & 6 \\
    $\mathrm{D_{3d}}$ & $-3m$ & 1 & 0 & 1 & 0 & 2 & 0 & 3 & 0 & 3 & 0 & 4 & 0 & 6 \\
    $\mathrm{C}_6$ & $6$ & 1 & 1 & 1 & 1 & 1 & 1 & 3 & 3 & 3 & 3 & 3 & 3 & 5 \\
    $\mathrm{C_{3h}}$ & $-6$ & 1 & 0 & 1 & 2 & 1 & 2 & 3 & 2 & 3 & 4 & 3 & 4 & 5 \\
    $\mathrm{C_{6h}}$ & $6/m$ & 1 & 0 & 1 & 0 & 1 & 0 & 3 & 0 & 3 & 0 & 3 & 0 & 5 \\
    $\mathrm{D}_6$ & $622$ & 1 & 0 & 1 & 0 & 1 & 0 & 2 & 1 & 2 & 1 & 2 & 1 & 4 \\
    $\mathrm{C_{6v}}$ & $6mm$ & 1 & 1 & 1 & 1 & 1 & 1 & 2 & 2 & 2 & 2 & 2 & 2 & 4 \\
    $\mathrm{D_{3h}}$ & $-6m2$ & 1 & 0 & 1 & 1 & 1 & 1 & 2 & 1 & 2 & 2 & 2 & 2 & 4 \\
    $\mathrm{D_{6h}}$ & $6/mmm$ & 1 & 0 & 1 & 0 & 1 & 0 & 2 & 0 & 2 & 0 & 2 & 0 & 4 \\
    $\mathrm{T}$ & $23$ & 1 & 0 & 0 & 1 & 1 & 0 & 2 & 1 & 1 & 2 & 2 & 1 & 3 \\
    $\mathrm{T_h}$ & $m\text{-}3$ & 1 & 0 & 0 & 0 & 1 & 0 & 2 & 0 & 1 & 0 & 2 & 0 & 3 \\
    $\mathrm{O}$ & $432$ & 1 & 0 & 0 & 0 & 1 & 0 & 1 & 0 & 1 & 1 & 1 & 0 & 2 \\
    $\mathrm{T_d}$ & $-43m$ & 1 & 0 & 0 & 1 & 1 & 0 & 1 & 1 & 1 & 1 & 1 & 1 & 2 \\
    $\mathrm{O_h}$ & $m\text{-}3m$ & 1 & 0 & 0 & 0 & 1 & 0 & 1 & 0 & 1 & 0 & 1 & 0 & 2 \\
    \bottomrule
\end{tabular}
\end{table}

\restateintheappendix{\SubgroupMonotonicityProposition}
\begin{proof}
If $\svv^{(l)}\in \sF^{(l)}(\GH_2)$, then $\rho^{(l)}(\Gh)\svv^{(l)}=\svv^{(l)}$ for every $\Gh\in\GH_2$, and hence in particular for every $\Gh\in\GH_1$. Thus $\svv^{(l)}\in\sF^{(l)}(\GH_1)$, which proves the subspace inclusion. The dimension and rank-template inequalities follow immediately.
\end{proof}

\subsection{Spectral invariance and leakage monotonicity}
\label{sec:appendix-spectral-properties}

\restateintheappendix{\LeakageMonotonicityProposition}
\begin{proof}
Fix a geometric graph $\gG$ and suppress the explicit argument $(\gG)$ in the derived leakage quantities.

If $\GH_1\leq \GH_2$, then \cref{prop:subgroup-monotonicity} gives
\begin{equation}
    m_l(\GH_2)\leq m_l(\GH_1)
    \qquad \text{for every } l\in\Lambda.
\end{equation}
Hence the tail index set in \cref{eq:degree-leakage} for $\GH_1$ is contained in the corresponding tail index set for $\GH_2$, so
\begin{equation}
    \sum_{i>m_l(\GH_1)}(\sigma_i^{(l)})^2
    \leq
    \sum_{i>m_l(\GH_2)}(\sigma_i^{(l)})^2.
\end{equation}
The denominator in \cref{eq:degree-leakage} is the same for both target groups; therefore
\begin{equation}
    \mathrm{LS}^{(l)}_{\GH_1}
    \leq
    \mathrm{LS}^{(l)}_{\GH_2}.
\end{equation}
Averaging over $l\in\Lambda$ as in \cref{eq:pooled-leakage} yields the first inequality in \cref{eq:leak-monotonicity}. Since $x\mapsto -\log_{10}(x+\varepsilon_{\log})$ is strictly decreasing on $[0,\infty)$, the second inequality for $\mathrm{pLS}$ follows immediately.
\end{proof}

\begin{proposition}[Orthogonal invariance of SymFP]
\label{prop:spectral-invariance}
Assume that for every $l\in\Lambda$, every geometric graph $\gG$, and every $\Gg\in \mathrm{O}(3)$,
\begin{equation}\label{eq:representation-equivariance}
    \smVl(\Gg\cdot\gG)
    =
    \rho^{(l)}(\Gg)\smVlgG.
\end{equation}
Then, for every $l\in\Lambda$,
\begin{equation}\label{eq:spectrum-invariance}
    \boldsymbol{\sigma}^{(l)}(\Gg\cdot\gG)
    =
    \boldsymbol{\sigma}^{(l)}(\gG).
\end{equation}
Consequently, $\mathrm{SymFP}_{\Lambda}$, $\widehat{\mathrm{RankSeq}}_{\Lambda,\tau}$, $\mathrm{LS}_{\Lambda,\GH}$, and $\mathrm{pLS}_{\Lambda,\GH}$ are all invariant under the global $\mathrm{O}(3)$ action.
\end{proposition}
\begin{proof}
Assume \cref{eq:representation-equivariance}. Since $\rho^{(l)}(\Gg)$ is orthogonal, left multiplication by $\rho^{(l)}(\Gg)$ does not change the singular values of $\smVlgG$. Therefore \cref{eq:spectrum-invariance} holds for every $l\in\Lambda$. The remaining listed quantities are deterministic functions of these singular values, the invariant threshold in \cref{eq:numerical-rank-threshold}, and fixed template multiplicities $m_l(\GH)$, so they inherit the same global $\mathrm{O}(3)$ invariance.
\end{proof}

Right multiplication of a representation block by an orthogonal channel matrix also preserves its singular values. A general non-isometric channel transformation can change the spectrum and its leakage scores, since these use the Euclidean metric of the chosen channel space.

\paragraph{Limiting cases of degree-resolved leakage.}
\label{sec:appendix-leakage-normalization}
Two limiting cases clarify the behavior of degree-resolved leakage. When $m_l(\GH)=2l+1$, the spectral tail is empty and $\mathrm{LS}^{(l)}_{\GH}=0$ regardless of normalization. Such degrees impose no rank constraint and contribute nothing to the pooled score. An example is the natural-parity scalar representation at $l=0$, for which $m_0(\GH)=1$ for every point group. Conversely, when $m_l(\GH)=0$, the fixed subspace is trivial and all block energy lies in the spectral tail. Under the default block-energy normalization, $\mathrm{LS}^{(l)}_{\GH}$ approaches unity as the block energy becomes large relative to $\varepsilon_{\mathrm{norm}}$.

\subsection{Attainability and generic realization of equivariant representations}
\label{sec:appendix-generic-rank}

For fixed node count, centred geometric graphs modulo node relabelling form a metrizable space under the natural quotient topology, with a continuous $\mathrm O(3)$ action. The following extension-and-averaging argument\citep{mostow1957equivariant} establishes pointwise attainability.

\begin{proposition}[Pointwise attainability of the fixed subspace]
\label{prop:pointwise-fixed-subspace}
Let a compact Hausdorff group $\mathfrak K$ act continuously on a metric space $\mathbb G$, with continuous output representation $\rho^{(l)}:\mathfrak K\to\mathrm{GL}(\R^{2l+1})$. For each $\gG$ with full stabilizer $\GH$, the continuous equivariant maps satisfy
\begin{equation}
\{f(\gG):f\in C_{\mathfrak K}(\mathbb G,\R^{2l+1})\}
=\sF^{(l)}(\GH).
\end{equation}
\end{proposition}
\begin{proof}
Equivariance gives the forward inclusion. For $\svv\in\sF^{(l)}(\GH)$, define $\varphi(\Gg\cdot\gG)=\rho^{(l)}(\Gg)\svv$ on the orbit $\mathcal O_{\gG}$. Since $\svv$ is fixed by $\GH$ and $\mathcal O_{\gG}\cong\mathfrak K/\GH$, this map is well-defined and continuous. The orbit is compact and closed, so coordinatewise Tietze extension gives a continuous $F:\mathbb G\to\R^{2l+1}$ extending $\varphi$. With normalized Haar measure $\nu$, set
\begin{equation}
f(\gG')=\int_{\mathfrak K}\rho^{(l)}(\Gg)^{-1}F(\Gg\cdot\gG')\,\mathrm d\nu(\Gg).
\end{equation}
Compactness and joint continuity imply continuity of $f$. Substituting $\mathfrak a=\Gg\mathfrak k$ and using right invariance of $\nu$ gives $f(\mathfrak k\cdot\gG')=\rho^{(l)}(\mathfrak k)f(\gG')$. On the original orbit, $F(\Gg\cdot\gG)=\rho^{(l)}(\Gg)\svv$, hence $f(\gG)=\svv$.
\end{proof}

\begin{corollary}[Rank attainment by universal equivariant networks]
\label{cor:universal-rank-attainment}
Under these assumptions, suppose an equivariant network family with $C$ output channels approximates all continuous equivariant matrix maps uniformly on compact sets. For each $\gG$ with full stabilizer $\GH$ and $C\geq m_l(\GH)$, some network satisfies
\begin{equation}
\operatorname{rank}\smVl_{\vtheta}(\gG)=m_l(\GH).
\end{equation}
The network and its architecture may depend on $\gG$.
\end{corollary}
\begin{proof}
Set $m=m_l(\GH)$; the case $m=0$ is immediate. By \cref{prop:pointwise-fixed-subspace}, realize an orthonormal basis of $\sF^{(l)}(\GH)$ as columns of a continuous equivariant map $\smVl_*$, padding with zero columns. Then $\sigma_m(\smVl_*(\gG))=1$. Approximation within $1/2$ in operator norm gives $\sigma_m(\smVl_{\vtheta}(\gG))>1/2$ by singular-value perturbation. Thus the rank is at least $m$, and equivariance bounds it by $m$.
\end{proof}

\paragraph{From attainability to generic realization.}
For a fixed architecture, input and degree, one full-span parameter setting suffices to establish full span almost everywhere under real-analytic parameter dependence on a connected open domain. The following results formalize this consequence using the zero-set theorem for real-analytic functions\citep{mityagin2020zero}.

Fix a geometric graph $\gG$ with finite stabilizer $\GH=\operatorname{Stab}(\gG)$, a degree $l$, and an equivariant architecture. Let $\smVl_{\vtheta}(\gG)\in\R^{(2l+1)\times C}$ denote its representation block, parameterized by $\vtheta\in\Omega\subset\R^P$. Assume that $\Omega$ is nonempty, open, and connected, and that the block entries depend real-analytically on $\vtheta$. Define the maximal attainable rank
\begin{equation}
r_l^*(\gG)\coloneqq\max_{\vtheta\in\Omega}\operatorname{rank}\smVl_{\vtheta}(\gG).
\end{equation}
By \cref{theo:fixed-subspace}, every column belongs to $\sF^{(l)}(\GH)$, hence
\begin{equation}
\label{eq:attainable-rank-constraint}
    \operatorname{rank}\smVl_{\vtheta}(\gG)
    \leq r_l^*(\gG)
    \leq \min\{C,m_l(\GH)\}.
\end{equation}

\begin{proposition}[Generic maximal-rank realization]
\label{prop:generic-maximal-rank}
Under these assumptions, $\operatorname{rank}\smVl_{\vtheta}(\gG)=r_l^*(\gG)$ for Lebesgue-almost every $\vtheta\in\Omega$. Thus, for any probability distribution $\mu$ absolutely continuous with respect to Lebesgue measure on $\Omega$,
\begin{equation}
\label{eq:generic-maximal-rank}
    \Pr_{\vtheta\sim\mu}\!\left[
    \operatorname{rank}\smVl_{\vtheta}(\gG)=r_l^*(\gG)
    \right]=1.
\end{equation}
\end{proposition}
\begin{proof}
Let $r=r_l^*(\gG)$. The case $r=0$ is immediate. For $r>0$, choose $\vtheta_*\in\Omega$ attaining rank $r$ and row and column index sets $\sI,\sJ$ of size $r$ such that
\begin{equation}
g(\vtheta)\coloneqq\det\bigl(\smVl_{\vtheta}(\gG)[\sI,\sJ]\bigr)
\end{equation}
satisfies $g(\vtheta_*)\neq0$. Since $g$ is real analytic and not identically zero, its zero set has Lebesgue measure zero\citep{mityagin2020zero}. Rank deficiency relative to $r$ implies $g(\vtheta)=0$, while maximality precludes larger ranks. The probability statement follows by absolute continuity.
\end{proof}

\begin{corollary}[Attainable full span is generic]
\label{cor:generic-full-span}
Under the same assumptions, suppose $C\geq m_l(\GH)$ and full span is attained at some $\vtheta_*\in\Omega$, i.e., $\operatorname{rank}\smVl_{\vtheta_*}(\gG)=m_l(\GH)$. Then
\begin{equation}
\label{eq:generic-full-span}
    \operatorname{colspan}\bigl(\smVl_{\vtheta}(\gG)\bigr)
    =\sF^{(l)}(\GH)
    \qquad\text{for $\mu$-almost every $\vtheta\in\Omega$.}
\end{equation}
\end{corollary}
\begin{proof}
The attainable-rank constraint gives $r_l^*(\gG)=m_l(\GH)$. The claim follows from \cref{prop:generic-maximal-rank} and the column-space inclusion.
\end{proof}

\begin{corollary}[Simultaneous realization on a finite collection]
\label{cor:generic-finite-collection}
Fix an architecture and parameter domain $\Omega$, and let $\sK=\{(\gG_i,l):1\leq i\leq N,\ l\in\Lambda_i\}$ be finite. If the analytic assumptions hold and full span is attainable for every pair, then, with probability one under $\mu$,
\begin{equation}
\operatorname{colspan}\bigl(\smVl_{\vtheta}(\gG_i)\bigr)
=\sF^{(l)}(\operatorname{Stab}(\gG_i))
\end{equation}
simultaneously for all $(\gG_i,l)\in\sK$.
\end{corollary}
\begin{proof}
Each pair has a null exceptional set by \cref{cor:generic-full-span}; their finite union remains null.
\end{proof}

The full-span witnesses may differ across pairs, yet simultaneous realization holds generically. The result requires neither independent channels nor Gaussian representations and applies to any fixed finite collection within an architecture satisfying the analytic assumptions.

\paragraph{Scope for the evaluated GNNs.}
We consider pre-SVD representation blocks at fixed inputs in exact arithmetic, holding initialization-fixed parameters and running statistics constant. HEGNN is analytic in its free parameters, whereas radial ReLU in TFN/MACE and guarded attention in Equiformer restrict generic full-span realization to connected analytic regions where full span is attainable. Replacing radial ReLU with SiLU yields analytic TFN/MACE variants without sacrificing attainability. Indeed, scaling the first affine map by $t$ and the second weight matrix by $t^{-1}$ gives $t^{-1}\operatorname{SiLU}(tz)\to\operatorname{ReLU}(z)$ as $t\to\infty$. The resulting blocks converge to their ReLU counterparts, preserving every nonzero minor of order $m_l(\GH)$ for sufficiently large finite $t$. Thus, \cref{cor:generic-full-span} applies to these analytic variants over the unrestricted free-parameter space. PF is deterministic in this setting.

\paragraph{Numerical evidence of full-span attainability.}
The rank-template match rate measures agreement between numerical rank sequences and the true point-group templates across all retained degrees. In \cref{tab:synthetic-initialization-seeds}, TFN, MACE, and Equiformer recover the complete rank sequences for all $320$ clean test inputs at $\Lambda_6$ across three initializations, with TFN and MACE also achieving perfect recovery at $\Lambda_9$. These results provide numerical evidence of full-span attainability in the evaluated configurations, although rank recovery remains sensitive to the positive singular values and the numerical threshold.

\subsection{Stability and continuous symmetry measures}
\label{sec:appendix-stability}

Let $\sU$ be a family of geometric graphs with identical node attributes, edge attributes, and adjacency, differing only in their coordinates. Equip $\sU$ with a permutation-invariant metric $\operatorname{dist}$.

\begin{theorem}[Local stability of SymFP on a fixed graph family]
\label{theo:local-stability}
Fix a finite degree set $\Lambda$, a finite family of point groups $\sC$, and a graph family $\sU$ as above. Suppose that, for each $l\in\Lambda$, there exists $L_l>0$ such that
\begin{equation}
\label{eq:representation-lipschitz}
    \|\smVl(\gG)-\smVl(\gG')\|_F\leq L_l\operatorname{dist}(\gG,\gG')
\end{equation}
for all $\gG,\gG'\in\sU$. Then
\begin{equation}
\label{eq:singular-value-stability}
    \|\boldsymbol{\sigma}^{(l)}(\gG)-\boldsymbol{\sigma}^{(l)}(\gG')\|_2
    \leq L_l\operatorname{dist}(\gG,\gG')
\end{equation}
for every $l\in\Lambda$. Consequently,
\begin{equation}
\label{eq:symfp-stability}
    \|\mathrm{SymFP}_{\Lambda}(\gG)-\mathrm{SymFP}_{\Lambda}(\gG')\|_\infty
    \leq C_{\Lambda}\operatorname{dist}(\gG,\gG'),
\end{equation}
where $C_{\Lambda}\coloneqq\max_{l\in\Lambda}L_l$.
\end{theorem}
\begin{proof}
Mirsky's inequality gives
\begin{equation}
    \|\boldsymbol{\sigma}^{(l)}(\gG)-\boldsymbol{\sigma}^{(l)}(\gG')\|_2
    \leq \|\smVl(\gG)-\smVl(\gG')\|_F,
\end{equation}
which, together with \cref{eq:representation-lipschitz}, proves \cref{eq:singular-value-stability}. Taking the maximum over all singular-value coordinates yields \cref{eq:symfp-stability}.
\end{proof}

\begin{definition}[Point-group-conditioned continuous symmetry measure]
\label{def:graph-csm}
For a point group $\GH\leq\mathrm{O}(3)$ with $\sU\cap\sG(\GH)\neq\emptyset$, define
\begin{equation}
\label{eq:graph-csm}
    \mathrm{CSM}_{\GH}(\gG)\coloneqq
    \inf_{\gG^\star\in\sU\cap\sG(\GH)}
    \operatorname{dist}(\gG,\gG^\star),
    \qquad \gG\in\sU.
\end{equation}
\end{definition}

This measure quantifies distance to the $\GH$-compatible subset within $\sU$, following the nearest-symmetric-configuration construction of \cref{sec:related-work}. By \cref{theo:local-stability}, taking the infimum over this subset gives
\begin{equation}
\label{eq:symfp-csm-bound}
    \inf_{\gG^\star\in\sU\cap\sG(\GH)}
    \|\mathrm{SymFP}_{\Lambda}(\gG)-\mathrm{SymFP}_{\Lambda}(\gG^\star)\|_\infty
    \leq C_{\Lambda}\mathrm{CSM}_{\GH}(\gG).
\end{equation}
The fixed-subspace constraint further yields a quadratic bound on spectral leakage.

\begin{corollary}[Leakage-score control via CSM]
\label{cor:leak-csm}
Under the assumptions of \cref{theo:local-stability}, consider the default block-energy normalization and nonnegative pooling weights $\beta_l$ with $\sum_{l\in\Lambda}\beta_l=1$. For any $\GH\in\sC$ satisfying $\sU\cap\sG(\GH)\neq\emptyset$ and any $\gG\in\sU$,
\begin{equation}
\label{eq:leak-csm}
    \mathrm{LS}_{\Lambda,\GH}(\gG)
    \leq C^{\mathrm{LS}}_{\Lambda}(\gG)\,
    \mathrm{CSM}_{\GH}(\gG)^2,
\end{equation}
where
\begin{equation}
\label{eq:leak-csm-constant}
    C^{\mathrm{LS}}_{\Lambda}(\gG)
    \coloneqq
    \sum_{l\in\Lambda}\beta_l
    \frac{L_l^2}{\|\smVl(\gG)\|_F^2+\varepsilon_{\mathrm{norm}}}.
\end{equation}
Consequently,
\begin{equation}
\label{eq:pls-csm}
    \mathrm{pLS}_{\Lambda,\GH}(\gG)
    \geq -\log_{10}\!\left(
    C^{\mathrm{LS}}_{\Lambda}(\gG)\,
    \mathrm{CSM}_{\GH}(\gG)^2+\varepsilon_{\log}
    \right).
\end{equation}
\end{corollary}
\begin{proof}
For any $\gG^\star\in\sU\cap\sG(\GH)$, the fixed-subspace constraint (\cref{eq:spectral-rank-constraint}) gives $\sigma_i^{(l)}(\gG^\star)=0$ for $i>m_l(\GH)$. Hence, by \cref{eq:singular-value-stability},
\begin{equation}
    \begin{aligned}
    \sum_{i>m_l(\GH)}(\sigma_i^{(l)}(\gG))^2
    &\leq
    \|\boldsymbol{\sigma}^{(l)}(\gG)
    -\boldsymbol{\sigma}^{(l)}(\gG^\star)\|_2^2\\
    &\leq L_l^2\operatorname{dist}(\gG,\gG^\star)^2.
    \end{aligned}
\end{equation}
Dividing by the block-energy normalization and pooling with weights $\beta_l$ yields
\begin{equation}
    \mathrm{LS}_{\Lambda,\GH}(\gG)
    \leq C^{\mathrm{LS}}_{\Lambda}(\gG)
    \operatorname{dist}(\gG,\gG^\star)^2.
\end{equation}
Taking the infimum over $\gG^\star\in\sU\cap\sG(\GH)$ proves \cref{eq:leak-csm}. The pLS bound follows from the monotonicity of $-\log_{10}$.
\end{proof}

The coefficient $C^{\mathrm{LS}}_{\Lambda}(\gG)$ reflects the block Lipschitz constants and energies, with $\varepsilon_{\mathrm{norm}}$ controlling low-energy blocks. Thus, spectral leakage is bounded quadratically by the distance to the target-symmetric subset.

\section{Relation to existing symmetry measures}
\label{sec:appendix-measures}

The energy retained by SymFP has a variational characterization determined by the target fixed-subspace dimension. The same trace objective underlies operation-based retained energy, obtained from the average operation residual in harmonic representation space. We derive these relations to support \cref{sec:theoretical-properties}.

\subsection{The operation average in representation space}
\label{sec:appendix-csom-derivation}

We derive the retained-energy identity in \cref{eq:harmonic-operation-residual} from a CSoM-style operation average in harmonic representation space.

Following \cref{sec:related-work}, let $\svv(\gG)=(\svvl(\gG))_l$ be a normalized harmonic embedding, with $\svvl(\gG)\in\R^{2l+1}$ transforming equivariantly:
\begin{equation}
    \svvl(\Gh\cdot\gG) =\rho^{(l)}(\Gh)\svvl(\gG).
\end{equation}
For a fixed graph, we suppress the argument $\gG$ and assume $\sum_l\|\svvl\|_2^2=1$, with sums taken over the retained degrees. For an oriented target group $\Gr\GH\Gr^{-1}$, define the average representation residual
\begin{equation}
    R_{\GH}(\gG;\Gr) \coloneqq \frac{1}{|\GH|}\sum_{\Gh\in\GH}\sum_l \|\svvl-\rho^{(l)}(\Gr\Gh\Gr^{-1})\svvl\|_2^2.
\end{equation}
The group average defines the orthogonal projector onto the rotated fixed subspace:
\begin{equation}
    \mP_{\GH}^{(l)}(\Gr) \coloneqq \frac{1}{|\GH|}\sum_{\Gh\in\GH} \rho^{(l)}(\Gr\Gh\Gr^{-1}).
\end{equation}
Expanding the squared norms and using the orthogonality of the representation matrices yields
\begin{equation}
    R_{\GH}(\gG;\Gr) =2\sum_l( \|\svvl\|_2^2-(\svvl)^\top\mP_{\GH}^{(l)}(\Gr)\svvl ) =2\sum_l\|(\mI_{2l+1}-\mP_{\GH}^{(l)}(\Gr))\svvl\|_2^2.
\end{equation}
Define the retained energy at degree $l$ by
\begin{equation}
    \alpha_{\GH}^{(l)}(\gG;\Gr) \coloneqq \|\mP_{\GH}^{(l)}(\Gr)\svvl\|_2^2 =(\svvl)^\top\mP_{\GH}^{(l)}(\Gr)\svvl,
\end{equation}
and let $\alpha_{\GH}(\gG;\Gr)\coloneqq\sum_l\alpha_{\GH}^{(l)}(\gG;\Gr)$. By normalization,
\begin{equation}
    R_{\GH}(\gG;\Gr) =2\sum_l( \|\svvl\|_2^2-\alpha_{\GH}^{(l)}(\gG;\Gr) ) =2(1-\alpha_{\GH}(\gG;\Gr)).
\end{equation}
Thus, minimizing the operation-averaged residual in harmonic representation space is equivalent to maximizing retained energy in the target fixed subspaces.

\subsection{Spectral and orientation-conditioned retained energy}
\label{sec:appendix-op-relation}

For the degree-$l$ equivariant representation, let $\mC^{(l)}\coloneqq\smVl(\smVl)^\top$ and $k\coloneqq m_l(\GH)$. SymFP retains the spectral energy $\gA_{\GH}^{(l),\mathrm{symfp}}=\sum_{i=1}^{k}(\sigma_i^{(l)})^2$, whose variational characterization in \cref{eq:spectral-retained-energy-variational} maximizes $\operatorname{tr}(\mP\mC^{(l)})$ over all rank-$k$ orthogonal projectors. Operation-based retained energy optimizes the same trace objective over projectors induced by spatial orientations of the target group.

\paragraph{Reduction of the operation average to a retained energy.}
Let $\mP_{\GH}^{(l)}$ be the orthogonal projector onto the fixed subspace $\sF^{(l)}(\GH)$. By \cref{eq:conjugacy-fixed-space}, under a proper spatial rotation $\Gr\in \mathrm{SO}(3)$ the same target group in the rotated frame has projector
\begin{equation}
\label{eq:op-rotated-projector}
    \mP_{\GH}^{(l)}(\Gr) \coloneqq \rho^{(l)}(\Gr) \mP_{\GH}^{(l)} \rho^{(l)}(\Gr)^\top = \frac{1}{|\GH|}\sum_{\Gh\in\GH} \rho^{(l)}(\Gr\Gh\Gr^{-1}).
\end{equation}
Consider the average operation residual at degree $l$ and orientation $\Gr$,
\begin{equation}
\label{eq:op-operation-residual}
    R_{\GH}^{(l)}(\Gr) \coloneqq \frac{1}{|\GH|}\sum_{\Gh\in\GH} \|\smVl-\rho^{(l)}(\Gr\Gh\Gr^{-1})\smVl\|_F^2.
\end{equation}
Using the orthogonality of each matrix $\rho^{(l)}(\Gr\Gh\Gr^{-1})$, the group-average identity in \cref{eq:op-rotated-projector}, and the fact that $\mP_{\GH}^{(l)}(\Gr)$ is an orthogonal projector, we obtain
\begin{equation}
\label{eq:op-operation-energy}
    R_{\GH}^{(l)}(\Gr) = 2\|(\mI_{2l+1}-\mP_{\GH}^{(l)}(\Gr))\smVl\|_F^2 = 2( \|\smVl\|_F^2-\operatorname{tr}(\mP_{\GH}^{(l)}(\Gr)\mC^{(l)}) ).
\end{equation}
At each orientation, the residual is twice the difference between the total energy at degree $l$ and the retained energy,
\begin{equation}
\label{eq:op-retained-energy}
    \gA_{\GH}^{(l)}(\Gr) \coloneqq \|\mP_{\GH}^{(l)}(\Gr)\smVl\|_F^2 = \operatorname{tr}(\mP_{\GH}^{(l)}(\Gr)\mC^{(l)}),
\end{equation}
which extends the relation between operation averages and retained energy to all channels of the equivariant representation. For Engel's single harmonic channel, $C=1$, $\smVl=\svvl$ and $\mC^{(l)}=\svvl(\svvl)^\top$, so \cref{eq:op-operation-residual,eq:op-retained-energy} reduce to the single-channel operation residual and retained energy of the previous section. The latter is $\alpha_{\GH}^{(l)}(\gG;\Gr)$; summing over the retained degrees at a common orientation $\Gr$ gives the pooled retained energy\citep{engel2021point}. Maximizing the normalized score over this shared orientation gives \cref{eq:engel-max-orientation}\citep{mbah2023early}.

To relate this projection-based retained energy to SymFP's spectral variational form degree by degree, we maximize over spatial orientations separately at each degree, giving the quantity $\gA_{\GH}^{(l),\mathrm{op}}\coloneqq \max_{\Gr\in\mathrm{SO}(3)} \gA_{\GH}^{(l)}(\Gr)$ used in \cref{sec:theoretical-properties}.

\paragraph{Feasible-set inclusion.}
SymFP optimizes over all rank-$k$ orthogonal projectors, whereas the operation-based formulation restricts the search to projectors onto $\rho^{(l)}(\Gr)\sF^{(l)}(\GH)$ for $\Gr\in\mathrm{SO}(3)$. Every such projector has rank $k$, so the operation-based feasible set is contained in that of \cref{eq:spectral-retained-energy-variational}. Maximizing over $\Gr$ yields \cref{eq:op-symfp-ordering}. The corresponding leakage inequality follows by subtracting both sides from $\|\smVl\|_F^2$ and normalizing.

\paragraph{Conditions for equality.}
Equality holds if a top-$k$ left-singular subspace of $\smVl$ coincides with $\rho^{(l)}(\Gr)\sF^{(l)}(\GH)$ for some $\Gr\in\mathrm{SO}(3)$. In particular, if $\Gr_0^{-1}\cdot\gG\in\sG(\GH)$, \cref{theo:fixed-subspace} ensures that the column space of $\smVl$ lies in $\rho^{(l)}(\Gr_0)\sF^{(l)}(\GH)$, so both formulations retain the full energy $\|\smVl\|_F^2$. Equality also holds at $l=1$, where rotations act transitively on subspaces of each admissible dimension in $\R^3$.

Symmetry-equivalent orientations yield identical fixed-subspace projectors and retained energies. SymFP obtains the maximum over all rank-$k$ projectors directly from the singular values, without selecting an orientation.

\paragraph{Subgroup ordering of approximate-symmetry quantities.}
For $\GH_1\leq\GH_2$ and a fixed orientation $\Gr$, the inclusion $\rho^{(l)}(\Gr)\sF^{(l)}(\GH_2)\subseteq\rho^{(l)}(\Gr)\sF^{(l)}(\GH_1)$ implies monotonicity of the raw retained energies along subgroup chains. SymFP captures this ordering through the fixed-subspace dimensions, yielding the leakage monotonicity in \cref{prop:leak-monotonicity} under common normalization and pooling. In contrast, classical CSoM normalizes sums of deviations over target-dependent operation sets, while Engel score involves the target-dependent factor $\beta_{\GH}$. Consequently, neither normalized score is guaranteed to preserve subgroup ordering, even at a fixed orientation.

\section{Datasets and evaluation settings}
\label{sec:appendix-experiment-details}

\subsection{Shared representation and scoring settings}
\label{sec:appendix-shared-settings}

Unless otherwise specified, evaluations use natural-parity degrees $\Lambda_6=\{0,\ldots,6\}$, normalization regularizer $\varepsilon_{\mathrm{norm}}=10^{-10}$, and logarithmic regularizer $\varepsilon_{\log}=10^{-10}$ (\cref{eq:degree-leakage,eq:pls}). Learnable GNNs use a fixed random initialization per configuration, except under explicit training or repeated-initialization protocols. Leakage scores are pooled uniformly across degrees (\cref{sec:spectral-formulation}). The p97, haemagglutinin, GluK2, and capping-water analyses use $\Lambda_6$; torsion scans use $\{0,2,4,6\}$, and E18 uses degree six alone. Crystal-field and controlled-deformation settings are specified separately below.

\paragraph{PF settings.}
PF uses hybrid blocks with $16$ Gaussian radial bases per branch (\cref{alg:pf}). Their centres are uniformly spaced on $[0,1]$, with width $\sigma=1/16$. Node and edge attributes are atomic numbers and lengths, respectively, giving $d_v=d_e=1$. Single-species deformation inputs and E18 C$\alpha$ inputs use $\vh_i=6$. Attributes remain unscaled; $Z_{\gV}$ and $Z_{\gE}$ normalize only the radial bases. Node, edge, and hybrid blocks have $16$, $48$, and $64$ channels, respectively.

Radial denominators are stabilized with $\varepsilon_0=10^{-10}$. Nodes satisfying $\|\gvx_{ic}\|\leq\delta(Z_{\gV}+\varepsilon_0)$, with $\delta=10^{-5}$, are treated as coincident with the centroid: their spherical-harmonic components vanish for $l>0$, while the scalar component is retained. The same convention applies to coincident edge endpoints.

\subsection{Synthetic point-group benchmark}
\label{sec:appendix-synthetic-protocol}

We generate $100$ finite point clouds for each of the $32$ crystallographic point groups, with $80/10/10$ training/validation/test splits within each group. Each cloud combines orbits of generic seed points, using fewer orbits for higher-order groups. We retain only clouds whose Euclidean point group is exactly the target, excluding accidental higher symmetries.

We compare PF, HEGNN, TFN, MACE, and Equiformer. At $\Lambda_6$, learnable GNNs use $16$ channels per degree; at $\Lambda_9$, they use $24$ channels per degree at random initialization, while PF uses $24$ radial bases per branch. HEGNN employs a multi-channel extension of the original architecture\citep{cen2024are} developed in this work.

Structures are classified by nearest rank-template matching under the $\ell_1$ rule (\cref{eq:rank-template-matching}), with ties broken by the ordering in \cref{tab:crystallographic-rank-templates}. Identification accuracy measures correct group assignments, whereas rank-template match rate requires agreement with the true group's template at every retained degree. Shared templates count as matches for either group but permit only one identification label. At $\Lambda_6$, two indistinguishable pairs cap identification accuracy at $30/32$ on the balanced benchmark.

\paragraph{Training regimes on the synthetic benchmark.}
The four learnable GNNs are evaluated at random initialization, after coordinate-denoising pretraining, and after supervised classification training. Both training regimes use the training split. For denoising, isotropic Gaussian noise is added to the coordinates, and the networks learn to predict displacements restoring the clean structures. The noise standard deviation is sampled from $[0.02,0.10]$ times the median nearest-neighbour distance during training and fixed at $0.05$ times that distance for validation. Checkpoints are selected by minimum validation coordinate MAE.

For supervised training, pooled representations of clean structures are used to predict the $32$ point-group labels. Models are trained with cross-entropy loss, with checkpoints selected by maximum validation accuracy. We perform one run per architecture in each training regime, keeping the architectures fixed across regimes.

\begin{table}[!htbp]
\centering
\fontsize{9}{11}\selectfont
\setlength{\tabcolsep}{4pt}
\renewcommand{\arraystretch}{1.15}
\captionsetup{font=normalsize,margin=0pt}
\caption{Discrete point-group identification across training regimes. Evaluations use natural-parity degrees $l=0,\ldots,6$ and relative rank tolerance $\tau_{\mathrm{rel}}=3\times10^{-5}$.}
\label{tab:synthetic-training-regimes}
\begin{tabular}{>{\raggedright\arraybackslash}p{\dimexpr0.22\linewidth-2\tabcolsep\relax}>{\raggedright\arraybackslash}p{\dimexpr0.2\linewidth-2\tabcolsep\relax}>{\centering\arraybackslash}p{\dimexpr0.3\linewidth-2\tabcolsep\relax}>{\centering\arraybackslash}p{\dimexpr0.28\linewidth-2\tabcolsep\relax}}
    \toprule
    Training regime & GNN & Identification accuracy (\%) & Template match rate (\%) \\
    \midrule
    ---
     & PF & 93.75 & 97.19 \\
    \midrule
    \multirow{4}{*}{Random init.}
     & HEGNN\citep{cen2024are} & 93.75 & 92.50 \\
     & TFN\citep{thomas2018tensor} & 93.75 & 100.00 \\
     & MACE\citep{batatia2022mace} & 93.75 & 100.00 \\
     & Equiformer\citep{liao2023equiformer} & 93.75 & 100.00 \\
    \midrule
    \multirow{4}{*}{Pretrained}
     & HEGNN\citep{cen2024are} & 93.75 & 91.88 \\
     & TFN\citep{thomas2018tensor} & 93.75 & 100.00 \\
     & MACE\citep{batatia2022mace} & 93.75 & 100.00 \\
     & Equiformer\citep{liao2023equiformer} & 93.75 & 100.00 \\
    \midrule
    \multirow{4}{*}{Supervised}
     & HEGNN\citep{cen2024are} & 93.44 & 94.38 \\
     & TFN\citep{thomas2018tensor} & 93.75 & 100.00 \\
     & MACE\citep{batatia2022mace} & 93.75 & 100.00 \\
     & Equiformer\citep{liao2023equiformer} & 93.75 & 100.00 \\
    \bottomrule
\end{tabular}
\end{table}

\begin{table}[!htbp]
\centering
\fontsize{9}{11}\selectfont
\setlength{\tabcolsep}{4pt}
\renewcommand{\arraystretch}{1.15}
\captionsetup{font=normalsize,margin=0pt}
\caption{Discrete point-group identification using degrees $l=0,\ldots,9$. Evaluations use natural-parity representations and validation-selected relative rank tolerance $\tau_{\mathrm{rel}}=10^{-6}$ (\cref{sec:appendix-synthetic-protocol}).}
\label{tab:synthetic-degree-nine}
\begin{tabular}{>{\raggedright\arraybackslash}p{\dimexpr0.22\linewidth-2\tabcolsep\relax}>{\centering\arraybackslash}p{\dimexpr0.43\linewidth-2\tabcolsep\relax}>{\centering\arraybackslash}p{\dimexpr0.35\linewidth-2\tabcolsep\relax}}
    \toprule
    GNN & Identification accuracy (\%) & Template match rate (\%) \\
    \midrule
    PF & 99.38 & 93.75 \\
    HEGNN\citep{cen2024are} & 99.06 & 92.19 \\
    TFN\citep{thomas2018tensor} & 100.00 & 100.00 \\
    MACE\citep{batatia2022mace} & 100.00 & 100.00 \\
    Equiformer\citep{liao2023equiformer} & 99.38 & 99.38 \\
    \bottomrule
\end{tabular}
\end{table}

\paragraph{Rank threshold.}
Numerical ranks use $\tau_{\mathrm{abs}}=0$ and a relative tolerance scaled by the leading degree-zero singular value $\sigma_1^{(0)}$ (\cref{eq:numerical-rank-threshold}), providing a common threshold across degrees, including those with $m_l(\GH)=0$. We calibrate the tolerance on clean validation structures by first maximizing the minimum identification accuracy across configurations, then the mean rank-template match rate. Across the thirteen $\Lambda_6$ configurations, this yields $\tau_{\mathrm{rel}}=3\times10^{-5}$ (\cref{tab:rank-tolerance-calibration}), which is held fixed for testing (\cref{tab:synthetic-training-regimes,tab:rank-tolerance-sweep}).

For $\Lambda_9$, we apply the same criterion across five GNNs, breaking any remaining ties in favour of the smaller tolerance. The search covers $\{1,3\}\times10^j$ for $j=-8,\ldots,-4$, together with $10^{-3}$. Calibration on $320$ validation structures selects $\tau_{\mathrm{rel}}=10^{-6}$, which is then applied to the $320$ test structures. Each learnable GNN retains identical initialized weights across validation and testing (\cref{tab:synthetic-degree-nine}).

\paragraph{Random-initialization repeats.}
We evaluate all four learnable GNNs across three independent random initializations for each degree set (\cref{tab:synthetic-initialization-seeds}), holding data splits and architectures fixed. The validation-calibrated tolerances remain $3\times10^{-5}$ for $\Lambda_6$ and $10^{-6}$ for $\Lambda_9$. Results in \cref{fig:synthetic-benchmark,tab:synthetic-training-regimes,tab:synthetic-degree-nine} use a consistent reference initialization for each architecture.

\begin{table}[!htbp]
\centering
\fontsize{9}{11}\selectfont
\setlength{\tabcolsep}{3pt}
\renewcommand{\arraystretch}{1.15}
\captionsetup{font=normalsize,margin=0pt}
\caption{Discrete point-group identification across random initializations.}
\label{tab:synthetic-initialization-seeds}
\begin{tabular}{>{\centering\arraybackslash}p{\dimexpr0.05\linewidth-2\tabcolsep\relax}>{\raggedright\arraybackslash}p{\dimexpr0.15\linewidth-2\tabcolsep\relax}>{\centering\arraybackslash}p{\dimexpr0.08\linewidth-2\tabcolsep\relax}>{\centering\arraybackslash}p{\dimexpr0.08\linewidth-2\tabcolsep\relax}>{\centering\arraybackslash}p{\dimexpr0.08\linewidth-2\tabcolsep\relax}>{\centering\arraybackslash}p{\dimexpr0.16\linewidth-2\tabcolsep\relax}>{\centering\arraybackslash}p{\dimexpr0.08\linewidth-2\tabcolsep\relax}>{\centering\arraybackslash}p{\dimexpr0.08\linewidth-2\tabcolsep\relax}>{\centering\arraybackslash}p{\dimexpr0.08\linewidth-2\tabcolsep\relax}>{\centering\arraybackslash}p{\dimexpr0.16\linewidth-2\tabcolsep\relax}}
\toprule
\multirow{2}{*}{$L$} & \multirow{2}{*}{GNN} & \multicolumn{4}{c}{Identification accuracy (\%)} & \multicolumn{4}{c}{Template match rate (\%)} \\
\cmidrule(lr){3-6}\cmidrule(lr){7-10}
 & & Seed 42 & Seed 43 & Seed 44 & Mean $\pm$ s.d. & Seed 42 & Seed 43 & Seed 44 & Mean $\pm$ s.d. \\
\midrule
\multirow{4}{*}{6} & HEGNN & 93.75 & 93.44 & 93.13 & $93.44 \pm 0.31$ & 92.50 & 92.50 & 91.25 & $92.08 \pm 0.72$ \\
 & TFN & 93.75 & 93.75 & 93.75 & $93.75 \pm 0.00$ & 100.00 & 100.00 & 100.00 & $100.00 \pm 0.00$ \\
 & MACE & 93.75 & 93.75 & 93.75 & $93.75 \pm 0.00$ & 100.00 & 100.00 & 100.00 & $100.00 \pm 0.00$ \\
 & Equiformer & 93.75 & 93.75 & 93.75 & $93.75 \pm 0.00$ & 100.00 & 100.00 & 100.00 & $100.00 \pm 0.00$ \\
\midrule
\multirow{4}{*}{9} & HEGNN & 99.06 & 99.06 & 99.06 & $99.06 \pm 0.00$ & 92.19 & 93.13 & 92.81 & $92.71 \pm 0.48$ \\
 & TFN & 100.00 & 100.00 & 100.00 & $100.00 \pm 0.00$ & 100.00 & 100.00 & 100.00 & $100.00 \pm 0.00$ \\
 & MACE & 100.00 & 100.00 & 100.00 & $100.00 \pm 0.00$ & 100.00 & 100.00 & 100.00 & $100.00 \pm 0.00$ \\
 & Equiformer & 99.38 & 99.38 & 99.38 & $99.38 \pm 0.00$ & 99.38 & 99.06 & 99.06 & $99.17 \pm 0.18$ \\
\bottomrule
\end{tabular}
\end{table}

\begin{table}[!htbp]
\centering
\fontsize{9}{11}\selectfont
\setlength{\tabcolsep}{4pt}
\renewcommand{\arraystretch}{1.15}
\captionsetup{font=normalsize,margin=0pt}
\caption{Sensitivity to rank tolerance. Group and Match denote point-group identification accuracy and rank-template match rate (\%). Bold marks the validation-selected tolerance.}
\label{tab:rank-tolerance-sweep}
\begin{tabular}{>{\raggedright\arraybackslash}p{\dimexpr0.12\linewidth-2\tabcolsep\relax}>{\centering\arraybackslash}p{\dimexpr0.088\linewidth-2\tabcolsep\relax}>{\centering\arraybackslash}p{\dimexpr0.088\linewidth-2\tabcolsep\relax}>{\centering\arraybackslash}p{\dimexpr0.088\linewidth-2\tabcolsep\relax}>{\centering\arraybackslash}p{\dimexpr0.088\linewidth-2\tabcolsep\relax}>{\centering\arraybackslash}p{\dimexpr0.088\linewidth-2\tabcolsep\relax}>{\centering\arraybackslash}p{\dimexpr0.088\linewidth-2\tabcolsep\relax}>{\centering\arraybackslash}p{\dimexpr0.088\linewidth-2\tabcolsep\relax}>{\centering\arraybackslash}p{\dimexpr0.088\linewidth-2\tabcolsep\relax}>{\centering\arraybackslash}p{\dimexpr0.088\linewidth-2\tabcolsep\relax}>{\centering\arraybackslash}p{\dimexpr0.088\linewidth-2\tabcolsep\relax}}
    \toprule
    \multirow{2}{*}{$\tau_{\mathrm{rel}}$} & \multicolumn{2}{c}{PF} & \multicolumn{2}{c}{HEGNN\citep{cen2024are}} & \multicolumn{2}{c}{TFN\citep{thomas2018tensor}} & \multicolumn{2}{c}{MACE\citep{batatia2022mace}} & \multicolumn{2}{c}{Equiformer\citep{liao2023equiformer}} \\
    \cmidrule(lr){2-3} \cmidrule(lr){4-5} \cmidrule(lr){6-7} \cmidrule(lr){8-9} \cmidrule(lr){10-11}
    & \multicolumn{1}{c}{Group} & \multicolumn{1}{c}{Match} & \multicolumn{1}{c}{Group} & \multicolumn{1}{c}{Match} & \multicolumn{1}{c}{Group} & \multicolumn{1}{c}{Match} & \multicolumn{1}{c}{Group} & \multicolumn{1}{c}{Match} & \multicolumn{1}{c}{Group} & \multicolumn{1}{c}{Match} \\
    \midrule
    \multicolumn{11}{c}{\textbf{PF and random init.}} \\
    \midrule
    $1\times10^{-7}$ & 87.50 & 77.81 & 93.75 & 98.75 & 93.75 & 99.69 & 93.75 & 100.00 & 28.44 & 15.62 \\
    $3\times10^{-7}$ & 93.75 & 93.44 & 93.75 & 99.06 & 93.75 & 100.00 & 93.75 & 100.00 & 86.56 & 84.38 \\
    $1\times10^{-6}$ & 93.75 & 97.19 & 93.75 & 97.81 & 93.75 & 100.00 & 93.75 & 100.00 & 93.75 & 99.69 \\
    $3\times10^{-6}$ & 93.75 & 97.19 & 93.75 & 96.88 & 93.75 & 100.00 & 93.75 & 100.00 & 93.75 & 99.69 \\
    $1\times10^{-5}$ & 93.75 & 97.19 & 93.75 & 94.06 & 93.75 & 100.00 & 93.75 & 100.00 & 93.75 & 100.00 \\
    $\mathbf{3\times10^{-5}}$ & \textbf{93.75} & \textbf{97.19} & \textbf{93.75} & \textbf{92.50} & \textbf{93.75} & \textbf{100.00} & \textbf{93.75} & \textbf{100.00} & \textbf{93.75} & \textbf{100.00} \\
    $1\times10^{-4}$ & 93.75 & 97.19 & 90.62 & 87.19 & 93.75 & 99.06 & 93.75 & 98.12 & 93.75 & 100.00 \\
    $3\times10^{-4}$ & 93.75 & 96.88 & 81.25 & 70.94 & 93.12 & 95.31 & 93.44 & 94.06 & 93.75 & 100.00 \\
    $1\times10^{-3}$ & 93.75 & 95.62 & 46.88 & 36.25 & 91.88 & 89.38 & 89.69 & 84.06 & 93.75 & 100.00 \\
    \midrule
    \multicolumn{11}{c}{\textbf{Pretrained}} \\
    \midrule
    $1\times10^{-7}$ & --- & --- & 93.75 & 98.75 & 92.81 & 88.75 & 24.06 & 15.31 & 25.31 & 15.31 \\
    $3\times10^{-7}$ & --- & --- & 93.75 & 99.06 & 93.75 & 99.69 & 84.38 & 78.44 & 88.44 & 90.00 \\
    $1\times10^{-6}$ & --- & --- & 93.75 & 97.81 & 93.75 & 100.00 & 93.75 & 99.69 & 93.75 & 99.38 \\
    $3\times10^{-6}$ & --- & --- & 93.75 & 96.56 & 93.75 & 100.00 & 93.75 & 100.00 & 93.75 & 99.69 \\
    $1\times10^{-5}$ & --- & --- & 93.75 & 93.12 & 93.75 & 100.00 & 93.75 & 100.00 & 93.75 & 100.00 \\
    $\mathbf{3\times10^{-5}}$ & --- & --- & \textbf{93.75} & \textbf{91.88} & \textbf{93.75} & \textbf{100.00} & \textbf{93.75} & \textbf{100.00} & \textbf{93.75} & \textbf{100.00} \\
    $1\times10^{-4}$ & --- & --- & 89.69 & 85.00 & 93.75 & 100.00 & 93.75 & 100.00 & 93.75 & 100.00 \\
    $3\times10^{-4}$ & --- & --- & 78.75 & 64.38 & 93.75 & 98.75 & 93.75 & 100.00 & 93.75 & 100.00 \\
    $1\times10^{-3}$ & --- & --- & 39.69 & 28.75 & 92.50 & 92.19 & 93.75 & 100.00 & 93.75 & 100.00 \\
    \midrule
    \multicolumn{11}{c}{\textbf{Supervised}} \\
    \midrule
    $1\times10^{-7}$ & --- & --- & 3.12 & 3.12 & 92.81 & 97.19 & 52.81 & 39.69 & 38.44 & 21.56 \\
    $3\times10^{-7}$ & --- & --- & 4.69 & 2.81 & 93.44 & 99.06 & 91.88 & 97.19 & 89.69 & 91.88 \\
    $1\times10^{-6}$ & --- & --- & 39.69 & 20.00 & 93.75 & 99.69 & 93.44 & 99.38 & 93.12 & 98.75 \\
    $3\times10^{-6}$ & --- & --- & 86.88 & 82.19 & 93.75 & 99.69 & 93.75 & 99.69 & 93.75 & 99.69 \\
    $1\times10^{-5}$ & --- & --- & 92.81 & 93.75 & 93.75 & 100.00 & 93.75 & 100.00 & 93.75 & 100.00 \\
    $\mathbf{3\times10^{-5}}$ & --- & --- & \textbf{93.44} & \textbf{94.38} & \textbf{93.75} & \textbf{100.00} & \textbf{93.75} & \textbf{100.00} & \textbf{93.75} & \textbf{100.00} \\
    $1\times10^{-4}$ & --- & --- & 92.81 & 91.88 & 93.75 & 99.69 & 93.75 & 100.00 & 93.75 & 100.00 \\
    $3\times10^{-4}$ & --- & --- & 89.06 & 86.88 & 93.75 & 96.88 & 93.75 & 100.00 & 93.75 & 100.00 \\
    $1\times10^{-3}$ & --- & --- & 83.12 & 77.81 & 92.50 & 92.19 & 93.75 & 100.00 & 93.75 & 100.00 \\
    \bottomrule
\end{tabular}
\end{table}

\paragraph{PF ablation.}
The PF ablation uses $\Lambda_6$ with relative rank tolerance $3\times10^{-5}$ (\cref{fig:synthetic-benchmark}c, \cref{tab:pf-ablation}). We compare node, edge, and hybrid branches using $N=16$ radial bases with width $\sigma=1/16$. For the hybrid block, we then vary the basis count over $N\in\{8,16,32\}$ with $\sigma=1/N$, followed by a width sweep at fixed $N=16$ over $\sigma\in\{\Delta/2,\Delta,2\Delta\}$, where $\Delta=1/(N-1)$ denotes the radial-centre spacing on $[0,1]$.

\begin{table}[!htbp]
\centering
\fontsize{9}{11}\selectfont
\setlength{\tabcolsep}{4pt}
\renewcommand{\arraystretch}{1.15}
\captionsetup{font=normalsize,margin=0pt}
\caption{PF ablation on synthetic structures. $N$ is the number of radial bases per branch, $\sigma$ their Gaussian width, and $\Delta=1/(N-1)$ their centre spacing on $[0,1]$. Bold marks the default hybrid configuration.}
\label{tab:pf-ablation}
\begin{tabular}{>{\raggedright\arraybackslash}p{\dimexpr0.20\linewidth-2\tabcolsep\relax}>{\centering\arraybackslash}p{\dimexpr0.10\linewidth-2\tabcolsep\relax}>{\centering\arraybackslash}p{\dimexpr0.04\linewidth-2\tabcolsep\relax}>{\centering\arraybackslash}p{\dimexpr0.10\linewidth-2\tabcolsep\relax}>{\centering\arraybackslash}p{\dimexpr0.28\linewidth-2\tabcolsep\relax}>{\centering\arraybackslash}p{\dimexpr0.28\linewidth-2\tabcolsep\relax}}
    \toprule
    Variant & Branch & $N$ & $\sigma$ & Identification accuracy (\%) & Template match rate (\%) \\
    \midrule
    node default            & node   & 16 & $0.0625$  & 57.50  & 52.81 \\
    edge default            & edge   & 16 & $0.0625$  & 92.81  & 94.69 \\
    \textbf{hybrid default} & \textbf{hybrid} & \textbf{16} & \textbf{$0.0625$} & \textbf{93.75} & \textbf{97.19} \\
    hybrid small $N$        & hybrid & 8  & $0.1250$  & 93.75 & 96.25 \\
    hybrid large $N$        & hybrid & 32 & $0.03125$ & 93.75 & 97.19 \\
    hybrid $\sigma=\Delta/2$ & hybrid & 16 & $0.0333$ & 93.75 & 96.88 \\
    hybrid $\sigma=\Delta$   & hybrid & 16 & $0.0667$ & 93.75 & 97.19 \\
    hybrid $\sigma=2\Delta$  & hybrid & 16 & $0.1333$ & 93.75 & 97.19 \\
    \bottomrule
\end{tabular}
\end{table}

\begin{table}[!htbp]
\centering
\fontsize{9}{11}\selectfont
\setlength{\tabcolsep}{4pt}
\renewcommand{\arraystretch}{1.15}
\captionsetup{font=normalsize,margin=0pt}
\caption{Validation-based rank-tolerance selection. Group and Match denote point-group identification accuracy and rank-template match rate (\%); Worst and Mean are taken across GNN and training-regime configurations.}
\label{tab:rank-tolerance-calibration}
\begin{tabular}{>{\raggedright\arraybackslash}p{\dimexpr0.16\linewidth-2\tabcolsep\relax}>{\centering\arraybackslash}p{\dimexpr0.28\linewidth-2\tabcolsep\relax}>{\centering\arraybackslash}p{\dimexpr0.28\linewidth-2\tabcolsep\relax}>{\centering\arraybackslash}p{\dimexpr0.28\linewidth-2\tabcolsep\relax}}
    \toprule
    $\tau_{\mathrm{rel}}$ & Worst Group (\%) & Mean Group (\%) & Mean Match (\%) \\
    \midrule
    $10^{-8}$          & 3.12  & 30.99 & 18.82 \\
    $3\times10^{-8}$   & 3.12  & 44.52 & 37.14 \\
    $10^{-7}$          & 3.12  & 62.64 & 59.40 \\
    $3\times10^{-7}$   & 4.06  & 84.47 & 87.40 \\
    $10^{-6}$          & 41.25 & 89.38 & 92.88 \\
    $3\times10^{-6}$   & 90.00 & 93.27 & 98.08 \\
    $10^{-5}$          & 93.12 & 93.68 & 98.58 \\
    $\mathbf{3\times10^{-5}}$ & \textbf{93.44} & \textbf{93.70} & \textbf{98.20} \\
    $10^{-4}$          & 90.94 & 93.32 & 96.85 \\
    $3\times10^{-4}$   & 76.25 & 91.06 & 92.19 \\
    $10^{-3}$          & 35.94 & 84.33 & 84.40 \\
    \bottomrule
\end{tabular}
\end{table}

\subsection{Controlled deformation benchmarks}
\label{sec:appendix-deformation-protocol}

We evaluate PF on single-species point clouds, isolating geometric symmetry. Single-step deformations begin with a unit octahedron or an eclipsed triangular prism. Stretching the octahedron along $z$ induces $\mathrm{O_h}\to\mathrm{D_{4h}}$, while straining along $[111]$ induces $\mathrm{O_h}\to\mathrm{D_{3d}}$. Twisting the prism's upper triangle breaks mirror symmetry, yielding $\mathrm{D_{3h}}\to\mathrm{D}_3$. Each deformation uses $41$ quadratically spaced amplitudes, reaching $0.30$ times the base radius for strains and $30^\circ$ for twisting.

Progressive symmetry breaking applies successive linear maps to the octahedron along $\mathrm{O_h}\to\mathrm{D_{4h}}\to\mathrm{D_{2h}}\to\mathrm{C_{2h}}\to\mathrm{C_i}$. The trajectory comprises $61$ frames, with $15$ steps per segment and exact subgroup symmetries at the boundaries. All maps preserve inversion. Point-group analysis independently verifies each frame's symmetry.

In \cref{fig:sym-breaking}a, singular values exceeding $10^{-4}$ are considered active. We display the smallest initially active, largest initially inactive, and all newly active singular values. In \cref{fig:sym-breaking}b, the total spectral tail energy relative to $\GH$ is $\sum_{l\in\Lambda_6}\sum_{i>m_l(\GH)}(\sigma_i^{(l)})^2$.

For matched-amplitude comparisons, we construct a $30$-point ring with exact $\mathrm{C}_6$ symmetry from six copies of a chiral five-point subunit, arranged at $60^\circ$ intervals about $z$ with radius $2.0$. Four displacement fields probe distinct symmetry-breaking modes. Axial staircase assigns subunit $k$ a displacement along $z$ proportional to $k$, mixing azimuthal orders and reducing the symmetry to $\mathrm{C}_1$. In-plane stretch applies $(x,-y,0)$ to each point $(x,y,z)$, producing azimuthal order $m=2$ and symmetry $\mathrm{C}_2$. In-plane radial displacement alternates outward and inward shifts between subunits, producing $m=3$ and $\mathrm{C}_3$. Axial pucker alternates subunit displacements along $\pm z$, likewise producing $m=3$ and $\mathrm{C}_3$.

To match deformation amplitudes, we project out the six rigid-body modes (three translations and three infinitesimal rotations) and normalize each displacement field to unit per-atom RMSD. Scaling by $\alpha$ then produces RMSD $\alpha$ relative to the initial structure. Each mode uses $21$ uniformly spaced amplitudes from $0$ to $0.30$, giving $84$ frames. Fourier analysis across subunits uses co-rotating displacement vectors, with lowest nonzero orders $|m|=1,2,3,3$ for staircase, stretch, in-plane radial, and axial pucker, respectively. Point-group analysis verifies the initial and deformed structures.

Matched-amplitude evaluations use fully connected graphs over the $30$ points with target $\mathrm{C}_6$. We measure degree-resolved relative tail energy (\cref{eq:degree-leakage}), RMSD from the initial structure, and $\mathrm{C}_6$ self-match RMSD. The latter compares each frame with its $60^\circ$ rotation about the initial ring axis under the subunit correspondence $k\mapsto k+1$. Spectral onset is the smallest degree $l$ at which relative tail energy exceeds $10^{-6}$.

\subsection{Biaryl torsion benchmark}
\label{sec:appendix-torsion-protocol}

The torsion benchmark considers nine candidate point groups: $\mathrm{C_s}$, $\mathrm{C}_2$, $\mathrm{C_{2v}}$, $\mathrm{D}_2$, $\mathrm{D_{2d}}$, $\mathrm{D_{2h}}$, $\mathrm{C_i}$, $\mathrm{C_{2h}}$, and $\mathrm{S}_4$. PF represents each conformation as a fully connected graph over all atoms. We analyse all $24$ conformations per scan using \texttt{pymatgen} with tolerance $0.1$, taking the distinct nontrivial point groups encountered as candidate targets. Of the $87$ scans, $86$ yield $96$ molecule--target pairs. Excluding two pairs whose target symmetry is present throughout the scan leaves $94$ pairs from $86$ molecules.

For each conformation--target pair, CSoM minimizes the mean coordinate deviation over non-identity group operations with uniform atom weights. We sample $10$ axes and $2$ angles, refine all $20$ initial orientations using Powell optimization, and retain the minimum deviation. Engel maximizes retained energy through a multistart orientation search that includes the CSoM-optimal orientation (\cref{sec:appendix-orientation-search}).

\subsubsection{Energy-extremum retrieval}
\label{sec:appendix-torsion-recall}

All methods share the same conformations, target groups, and energy labels. SymFP ranks conformations by even-degree pLS, Engel by its orientation-maximized score, and CSoM by negated deviation. On the periodic $15^\circ$ torsion grid, strict minima and maxima are conformations whose energies are respectively lower or higher than both neighbours.

For each molecule--target pair and selection budget $k=0,\ldots,24$, we retrieve the $k$ highest-ranked conformations. Recall is the fraction of energy extrema retrieved, with exact score ties resolved by their expected hit count under uniform tie-breaking. Energy labels determine the extrema but not the rankings. Trapezoidal integration of recall against $k/24$ gives the recall--budget AUC, with random-ranking expectation $0.5$.

We average AUCs over targets within each molecule and then equally over molecules, separately for minima, maxima, and paired method differences. Percentile $95\%$ confidence intervals use $5{,}000$ molecule-level bootstrap resamples, paired for method comparisons and without multiple-comparison adjustment. In paired-difference displays, per-molecule AUC differences above $10^{-12}$ count as wins, those below $-10^{-12}$ as losses, and the remainder as numerical ties. The example in \cref{fig:torsion}b uses biphenyl record $21272387$ with target $\mathrm{D_{2h}}$.

For \cref{edtab:torsion-symmetry-groups}, extrema are stratified by energy type and target-symmetry presence, defined by whether the assigned point group contains the target, including equality. Rankings retain all $24$ conformations, while recall counts only extrema in the specified class. Pairs without eligible extrema are excluded, and results are averaged over eligible targets within molecules and then equally over molecules.

\paragraph{Sensitivity to degree selection.}
For centrosymmetric targets, odd natural-parity degrees have trivial fixed subspaces, causing degree-resolved LS to saturate near unity when block energies substantially exceed the regularization scale. We therefore use a common even-degree set in the main analysis. This makes $\mathrm{D}_2$ and $\mathrm{D_{2h}}$ indistinguishable by SymFP, as their even-degree fixed-subspace dimensions coincide. As a sensitivity check, pooling over $\Lambda_6$ with all other settings unchanged yields a mean recall--budget AUC of $0.771$, compared with $0.794$ for even-degree pooling.

\subsubsection{Engel orientation search}
\label{sec:appendix-orientation-search}

We evaluate Engel score (\cref{eq:engel-max-orientation}) by minimizing the pooled projection residual over a common spatial rotation. For bond-direction coefficients $\svvl$ at degrees $l=1,\ldots,6$, the objective is
\begin{equation}
\label{eq:engel-orientation-loss}
L_{\GH}^{\mathrm{E}}(\Gr)
=\frac{\sum_{l=1}^{6}\|(\mI-\mP_{\GH}^{(l)}(\Gr))\svvl\|_2^2}
{\sum_{l=1}^{6}\|\svvl\|_2^2},
\end{equation}
yielding the score $1-L_{\GH}^{\mathrm{E}}/(1-\beta_{\GH})$.

Rotations are parameterized by Y--X--Y Euler angles. BFGS refinement uses analytic gradients derived from representation generators, a gradient tolerance of $10^{-10}$, and at most $200$ iterations per run. We evaluate the identity and $128$ Haar-distributed rotations, then refine the best four among the identity and first $32$ samples, the best twelve among all $129$ orientations, and the CSoM-fitted orientation. The minimum finite residual across sampled and refined orientations determines the score.

\subsubsection{Learnable GNNs and torsion pretraining}
\label{sec:appendix-torsion-pretraining}

We evaluate HEGNN, TFN, MACE, and Equiformer on complete directed graphs without self-edges, retaining all atoms. Node features encode elements, edges carry interatomic distances, and coordinates are centred without rescaling. Within each architecture, random, energy-pretrained, and Uni-Mol-style variants share the same initialization. The molecule-level split comprises $69$ training molecules ($1{,}656$ conformations) and $17$ validation molecules ($408$ conformations).

For energy pretraining, mean-pooled scalar features predict deposited total energies in Hartree, standardized by the training-set mean and population standard deviation. The GNN and prediction head are jointly optimized using mean absolute error.

Uni-Mol-style pretraining selects approximately $15\%$ of atoms per conformation, with at least one atom selected. Among selected atoms, $90\%$ are masked, $5\%$ are replaced by an element sampled uniformly from the training-set elements, and $5\%$ remain unchanged. Masked and replaced atoms receive independent coordinate perturbations drawn uniformly from $[-1,1]$\,\AA\ per component, after which edge distances are recomputed. Training corruption is resampled each epoch, whereas validation corruption is fixed across epochs and architectures.

The pretraining loss combines masked-atom cross-entropy, coordinate SmoothL1, and pair-distance SmoothL1 with weights $1$, $5$, and $10$, respectively. Both SmoothL1 terms use $\beta=1$. Losses are computed at selected atoms, with pair-distance predictions spanning each selected atom and all atoms in its conformation. Distance targets are standardized using training-set statistics. Energy labels are not used.

Checkpoints are selected by the lowest validation energy MAE or fixed-corruption validation loss, respectively. Prediction heads are discarded, and spectral evaluation uses clean conformations with the shared even-degree pLS protocol. For \cref{edtab:torsion-pretraining}, we report raw pLS and two contrasts: the maxima--minima gap, defined as mean pLS at energy maxima minus mean pLS at energy minima, and the target-symmetry gap, defined as mean pLS with the target symmetry present minus that with it absent. Target presence follows the point-group assignments at tolerance $0.1$, including equality.

Gaps are averaged over targets within each molecule. Relative to paired random initializations, increases exceeding $10^{-10}$ in maxima--minima gaps or $10^{-6}$ in target-symmetry gaps count as improvements. Changes in mean pLS, gap, and retrieval AUC are averaged equally across all $86$ molecules. Improvement counts are reported for the full cohort and separately for the $17$ validation molecules.

\subsection{DOTA complexes}
\label{sec:appendix-dota-protocol}

The water-free \ce{Ln(DOTA)} series\cite{manvell2024lndota} comprises seven CCDC depositions: \texttt{2281620} (\ce{Y}), \texttt{2281617} (\ce{Tb}), \texttt{2281615} (\ce{Dy}), \texttt{2281619} (\ce{Ho}), \texttt{2281616} (\ce{Er}), \texttt{2281618} (\ce{Tm}), and \texttt{2281329} (\ce{Yb}). All crystallize in space group $P4bm$, with the metal on the fourfold axis. Metal-centred clusters are reconstructed using space-group operations and bond tracing, with a $2.9\,\text{\AA}$ metal--donor cutoff and a covalent cutoff of $1.30$ times the sum of Cordero radii for other bonds. Counterions and lattice solvent are excluded, yielding $53$-atom \ce{[Ln(DOTA)]-} units of composition \ce{LnO8N4C16H24}.

The approximate-symmetry structures \ce{[Dy(DOTA)(H2O)]-} and \ce{Tm(DOTA)} are taken from the CSoM package\cite{nielsen2025determining}. Their heavy-atom structures contain $30$ and $29$ atoms, respectively, with the metal centred at the origin. The normal to the least-squares plane through the four coordinated macrocycle nitrogens defines the approximate fourfold axis. Because these structures lack hydrogens, the tetragonal series is likewise evaluated without hydrogens in \cref{fig:lndota-cf}f.

We evaluate each structure against $\mathrm{C}_4$, $\mathrm{S}_4$, $\mathrm{C_{4v}}$, $\mathrm{D_{4h}}$, $\mathrm{D}_4$, $\mathrm{D_{2d}}$, $\mathrm{C_{2v}}$, $\mathrm{C}_2$, and $\mathrm{C}_1$. Crystal-field analysis reports degree-resolved leakage $\mathrm{LS}^{(k)}_{\GH}$ at degree $l=k$ for crystal-field rank $k$ (\cref{eq:degree-leakage}).

The package also provides a water-rotation trajectory for \ce{[Dy(DOTA)(H2O)]-}. Its $360$ frames rotate the capping water about the \ce{Dy-O} axis in $1^\circ$ increments, with the coordination cage fixed. For each score series, we subtract its mean and identify the dominant harmonic among the first $24$ nonzero angular harmonics (\cref{tab:dota-periodicity}). Engel score is evaluated at every frame for the four shared targets using the orientation search in \cref{sec:appendix-orientation-search}.

\begin{table}[!htbp]
\centering
\fontsize{9}{11}\selectfont
\setlength{\tabcolsep}{4pt}
\renewcommand{\arraystretch}{1.15}
\captionsetup{font=normalsize,margin=0pt}
\caption{Dominant harmonics of capping-water rotation. Entries give the largest-amplitude harmonic among $n=1,\ldots,24$ after mean subtraction. Stars mark noise-sensitive components of near-constant CSoM profiles; a dash denotes an unavailable target.}
\label{tab:dota-periodicity}
\begin{tabular}{>{\raggedright\arraybackslash}p{\dimexpr0.25\linewidth-2\tabcolsep\relax}>{\centering\arraybackslash}p{\dimexpr0.25\linewidth-2\tabcolsep\relax}>{\centering\arraybackslash}p{\dimexpr0.25\linewidth-2\tabcolsep\relax}>{\centering\arraybackslash}p{\dimexpr0.25\linewidth-2\tabcolsep\relax}}
\toprule
Target & SymFP & CSoM & Engel \\
\midrule
$\mathrm{C}_4$ & $4\alpha$ & $16\alpha^{*}$ & $4\alpha$ \\
$\mathrm{C_{4v}}$ & $4\alpha$ & $8\alpha$ & $4\alpha$ \\
$\mathrm{C}_2$ & $4\alpha$ & $21\alpha^{*}$ & $4\alpha$ \\
$\mathrm{C_{2v}}$ & $4\alpha$ & $4\alpha$ & $4\alpha$ \\
$\mathrm{C_s}$ & $8\alpha$ & $2\alpha$ & --- \\
\bottomrule
\end{tabular}
\end{table}

\FloatBarrier

\subsection{Biological assemblies}
\label{sec:appendix-assemblies-protocol}

\subsubsection{Matched regional analysis of p97}
\label{sec:appendix-p97-localization}

We analyse six p97--UBXD1 structures from Braxton et al.\citep{braxton2023p97}: \texttt{8FCL}, \texttt{8FCM}, \texttt{8FCR}, \texttt{8FCQ}, \texttt{8FCO}, and \texttt{8FCT}, all reconstructed with $\mathrm{C}_1$ symmetry. The closed state \texttt{8FCL} serves as reference. Each assembly contains six p97 chains identified by accession \texttt{P55072}. Atoms are matched across chains and structures by residue number, insertion code, residue identity, atom name, and element, retaining only residues with complete N, C$\alpha$, C, and O backbones. This yields $712$ residues per chain and $17{,}088$ atoms per assembly. D1 spans residues $209$--$460$ and D2 residues $481$--$763$; each region is independently centred, represented, and scored.

We scan D1 and D2 using $40$-residue windows with a stride of $20$, adding a domain-end-aligned window where necessary. The resulting $12$ D1 and $14$ D2 windows require at least $50\%$ residue coverage and ten complete residues per chain. All D1 and eight D2 windows retain $40$ common residues per chain; the six partially covered D2 windows are indicated in \cref{fig:p97}d. Graphs use $k=24$ nearest neighbours, including distance ties within absolute and relative tolerances of $10^{-7}$ and $10^{-6}$. Node and edge radial scales are computed separately for each region.

Local gap annotations in \cref{fig:p97}c quantify the minimum separation between exposed van der Waals surfaces of adjacent windows, using deposited heavy atoms, including side chains. With Biopython atomic radii, changes exceeding $0.5$\,\AA{} are annotated when their direction is consistent at both $960$ and $240$ surface points per atom.

\subsubsection{Graph-neighbourhood and window-scale checks}
\label{sec:appendix-regional-scales}

For the six matched p97 structures, we assess whole-domain D1 and D2 scores across neighbourhood sizes $k\in\{6,12,16,24,32,48,96\}$, retaining distance ties (\cref{edfig:assembly-scale}).

For window-scale comparisons, we combine window centres from $40$- and $80$-residue scans with stride $20$ and domain-end alignment. At each centre, we construct windows of $20$, $40$, $60$, and $80$ residues, clipped to the domain and restricted to complete-backbone residues matched across all six chains. Each selection is independently centred and scored at $k=24$, with the closed-reference score subtracted at the same centre and width. Window lengths refer to sequence intervals before unresolved residues are excluded. Unlike the main scan, this analysis retains all centres regardless of coverage, including partially resolved D2 windows with as few as three common residues per chain.

\subsubsection{Haemagglutinin}
\label{sec:appendix-ha-protocol}

We analyse the X-31 haemagglutinin ectodomain series\cite{benton2020structural}: the pH-$8$ reference \texttt{6Y5G} and four successive pH-$5$ states I--IV (\texttt{6Y5H}, \texttt{6Y5I}, \texttt{6Y5J}, \texttt{6Y5K}). Chains are identified by accession \texttt{P03437}, with HA1 chains A, C, and E and HA2 chains B, D, and F forming three protomers.

The reference and states I--IV were reconstructed under $\mathrm{C}_3$, $\mathrm{C}_3$, $\mathrm{C}_1$, $\mathrm{C}_1$, and $\mathrm{C}_3$ symmetry constraints, respectively. Fixed residue windows define HA1 (residues $45$--$310$), HA2 ($38$--$125$), and their union. These selections contain $4{,}248$ backbone atoms per state, except state III ($4{,}236$). Restricting states III and IV to the common HA2 range $38$--$124$ preserves the pLS increase ($1.196$ versus $1.203$ with the original windows).

Each scope is independently centred and evaluated against $\mathrm{C}_3$ using a fixed $24$-nearest-neighbour graph, without the distance-tie inclusion used for p97. For each state, the self-matching axis is normal to the plane through the three protomer centroids, computed from the combined HA1--HA2 C$\alpha$ selection. HA1 and HA2 share this axis.

\subsubsection{GluK2}
\label{sec:appendix-gluk2-protocol}

We use biological assembly 1 of the desensitized GluK2 tetramer \texttt{5KUF}\citep{meyerson2016structural}, reconstructed with $\mathrm{C}_2$ symmetry. In author residue numbering, the amino-terminal domain (ATD) spans residues $1$--$384$; the ligand-binding domain (LBD), $398$--$513$ and $636$--$775$; and the transmembrane domain (TMD), resolved helix cores $521$--$554$, $599$--$629$, and $785$--$809$. Boundaries follow the source study and UniProt \texttt{P42260}, mapped to author numbering. Retaining only residues with complete N, C$\alpha$, C, and O atoms matched across all four chains yields $6{,}128$, $4{,}096$, and $1{,}360$ atoms in ATD, LBD, and TMD, respectively.

Each domain is independently centred and scored against $\mathrm{C}_4$ using a $24$-nearest-neighbour graph. For spatial profiling, we define the receptor axis from the TMD to the ATD C$\alpha$ centroid and partition the structure into non-overlapping layers of width $10$, $20$, $30$, $40$, $50$, or $60$\,\AA, starting at the lowest selected C$\alpha$ and retaining any terminal remainder. Residues are assigned by C$\alpha$ height within each chain, retaining all four backbone atoms. Layers are lower-inclusive and upper-exclusive, except the final layer, which includes the highest C$\alpha$. Each layer is independently centred and represented; plotted heights are measured from the lowest selected C$\alpha$.

\subsubsection{Echovirus 18}
\label{sec:appendix-e18-protocol}

We analyse biological assembly 1 of \texttt{6HBH}, \texttt{6HBK}, \texttt{6HHT}, and \texttt{6HBL}\citep{buchta2019enterovirus}, representing the intact activated capsid and capsids lacking one, two, and three pentamers, respectively. The four classes were reconstructed under $\mathrm{I}$, $\mathrm{C}_5$, $\mathrm{C}_2$, and $\mathrm{C}_3$ symmetry constraints. Capsid-protein sequences are aligned to \texttt{6HBH}, retaining reference positions and residue identities shared across all copies and models within each protein. All physically present copies are included, yielding $40{,}800$, $37{,}400$, $34{,}000$, and $30{,}600$ C$\alpha$ atoms, respectively. E18 uses a $24$-nearest-neighbour graph and degree-six pLS against the proper icosahedral group $\mathrm{I}$, whose degree-six fixed subspace is one-dimensional.

\subsection{Computational cost}
\label{sec:appendix-computational-cost}

\paragraph{Controlled computational benchmark.}
Timings are measured on a workstation with an Intel Xeon Gold 5318Y CPU and an NVIDIA RTX 4090 GPU, using a single CPU thread for both CPU and GPU measurements. Inputs begin as coordinates and atomic attributes in host memory. SymFP timings include centring, graph construction, host-to-device transfers where applicable, representation computation, SVD of each degree block, and target scoring, with final scores returned to the host. GPU timing boundaries are synchronized. Model initialization, group tables, and one-time code generation are excluded.

Complete-input benchmarks use the $17{,}088$ matched backbone atoms of closed p97 (\texttt{8FCL}) and the $4{,}248$ atoms of the HA global selection (\texttt{6Y5G}), following their respective graph and centring conventions. Regional throughput measures complete scans of all $26$ windows across six p97 structures ($156$ evaluations per repetition), including window selection and representation reconstruction. Each region contains $480$--$960$ atoms.

CSoM uses the official implementation, with either the deposited target orientation or the multistart fitting procedure in \cref{sec:appendix-torsion-protocol}. Engel retains all directed pair contributions.

For search-free evaluations, a calibration call determines the repetition count: calls of at most $2$\,s receive three warm-ups and ten measured repetitions; longer calls receive one warm-up and three repetitions. Medians are reported. Biological coordinate-fitting runs use a single trial without search warm-up, subject to a $7200$\,s limit. Configurations terminate at the first timeout, with incomplete runs reported as censored.

\paragraph{Multiple target groups.}
We benchmark the $30$-heavy-atom Dy-DOTA-water structure against the first $1$, $2$, $4$, or $7$ targets in the ordered sequence $\mathrm{C}_2$, $\mathrm{C}_4$, $\mathrm{C}_6$, $\mathrm{D}_2$, $\mathrm{D}_4$, $\mathrm{D}_6$, $\mathrm{O}$. Coordinates retain the deposited metal-centred frame, and PF uses a complete graph.

Within each timed call, SymFP shares representations and spectra across targets, while Engel reuses its harmonic descriptor. CSoM shares centring and, for fixed orientations, deviations under common operation matrices. Coordinate fitting and BFGS optimize targets separately, whereas grid search reuses rotated descriptors. All caching is confined to individual calls, with shared computations counted once. CSoM uses the Powell procedure in \cref{sec:appendix-torsion-protocol}. DOTA CSoM timings are medians of three repetitions following a warm-up on the seven-target set, using one pinned CPU thread.

\paragraph{Engel search timing.}
\label{sec:appendix-engel-search-timing}

Panels a and b of \cref{edtab:computational-cost} compare Engel grid search with BFGS refinement, with and without CSoM initialization. Both BFGS variants follow \cref{sec:appendix-orientation-search}, differing only in the additional coordinate-fitted start and its refinement. Timings include coordinate preprocessing, all-pair descriptor construction, optional coordinate fitting, retained-energy optimization, and final scoring. Group tables and fixed numerical kernels are prepared beforehand.

The direct grid search samples Y--X--Y Euler angles at $15^\circ$ intervals, yielding $24\times13\times24=7{,}488$ triples, including repeated orientations at the poles, without local refinement. Timings include rotation construction, descriptor rotation, and projection, with rotated descriptors reused across targets. The grid and BFGS timings reflect their respective search protocols rather than matched optimization accuracy.

Biological grid-search and BFGS runs without CSoM initialization, together with all DOTA Engel searches, are reported as medians of three repetitions on one pinned CPU thread. Biological runs are prewarmed on the smaller DOTA input, while DOTA runs receive their own warm-ups. Biological BFGS runs with CSoM initialization use single trials.

\section{Supporting studies}
\label{sec:appendix-studies}

\subsection{Spectral responses across equivariant GNNs}
\label{sec:appendix-gnn-responses}

We compare PF with four randomly initialized learnable GNNs on the controlled deformations and crystallographic complexes in \cref{sec:exp-sym-breaking,sec:exp-dota}.

\paragraph{Degree-resolved responses across GNNs.}
Along the subgroup chain, all five GNNs detect each symmetry-breaking transition at the first deformed frame (\cref{edfig:cross-gnn}a). Because all deformations preserve inversion, leakage relative to $\mathrm{C_i}$ remains at the numerical floor.

For matched-amplitude deformations, spectral onset is the lowest degree whose relative tail energy exceeds $10^{-6}$ at maximum amplitude, excluding round-off blocks. A block is classified as round-off if
\begin{equation}
\label{eq:roundoff-floor}
    E_l < N\,\epsilon_{\mathrm{fp32}}^{2}\,E_{\max},
\end{equation}
where $E_l$ is its energy, $E_{\max}$ the maximum block energy for the same model and structure, and $\epsilon_{\mathrm{fp32}}$ the single-precision machine epsilon. The node count $N=30$ accounts for accumulated round-off in pooling.

PF, TFN, MACE, and Equiformer recover onset degrees $l=1,2,3,3$ for axial staircase, in-plane stretch, in-plane radial, and axial pucker, respectively (\cref{tab:mode-onsets}). For HEGNN, the degree-three axial-pucker block remains at round-off throughout the sweep, shifting the first energy-supported response to $l=4$ (\cref{edfig:cross-gnn}b).

\begin{table}[!htbp]
\centering
\fontsize{9}{11}\selectfont
\setlength{\tabcolsep}{2pt}
\renewcommand{\arraystretch}{1.15}
\captionsetup{font=normalsize,margin=0pt}
\caption{\textbf{Onset degrees across deformation modes.} Onsets exclude round-off blocks (\cref{eq:roundoff-floor}). $\dagger$ The $l=3$ block remains at round-off; the first energy-supported response occurs at $l=4$.}
\label{tab:mode-onsets}
\begin{tabular}{>{\raggedright\arraybackslash}p{\dimexpr0.14\linewidth-2\tabcolsep\relax}>{\centering\arraybackslash}p{\dimexpr0.21\linewidth-2\tabcolsep\relax}>{\centering\arraybackslash}p{\dimexpr0.23\linewidth-2\tabcolsep\relax}>{\centering\arraybackslash}p{\dimexpr0.21\linewidth-2\tabcolsep\relax}>{\centering\arraybackslash}p{\dimexpr0.21\linewidth-2\tabcolsep\relax}}
    \toprule
    GNN & Axial staircase ($|m|=1$) & In-plane stretch ($|m|=2$) & In-plane radial ($|m|=3$) & Axial pucker ($|m|=3$) \\
    \midrule
    PF                        & 1 & 2 & 3 & 3 \\
    HEGNN\citep{cen2024are}              & 1 & 2 & 3 & $4^{\dagger}$ \\
    TFN\citep{thomas2018tensor}           & 1 & 2 & 3 & 3 \\
    MACE\citep{batatia2022mace}           & 1 & 2 & 3 & 3 \\
    Equiformer\citep{liao2023equiformer}  & 1 & 2 & 3 & 3 \\
    \bottomrule
\end{tabular}
\end{table}

To assess the effect of node attributes, we assign distinct types to the five node orbits, constant within each $\mathrm{C}_6$ orbit and fixed across deformations. This preserves the underlying symmetries and restores HEGNN's axial-pucker onset to $l=3$, yielding the expected onset sequence $l=1,2,3,3$ across all five GNNs.

\paragraph{Crystal-field responses across GNNs.}
For the seven crystallographic $\mathrm{C}_4$ clusters, all five GNNs exhibit numerical-floor leakage against $\mathrm{C}_4$ and identical even-degree scores for $\mathrm{C}_4$ and $\mathrm{S}_4$. The five-parameter targets show increased leakage at $k=4,6$, whereas $k=2$ remains at the floor because each target admits one independent parameter at that degree.

\FloatBarrier

\subsection{Graph-neighbourhood and window-scale checks for p97}
\label{sec:appendix-p97-scale-checks}

We assess sensitivity to graph neighbourhood size and window width using the matched p97 structures (\cref{sec:appendix-regional-scales}). Relative to the closed reference, the UBXD1 lariat mutant consistently exhibits higher D1 and lower D2 whole-domain pLS across all tested neighbourhood sizes. Its elevated D1 and predominantly reduced D2 regional pLS persist across all four window widths (\cref{edfig:assembly-scale}). The other four structures likewise show lower whole-domain D2 pLS throughout the neighbourhood sweep. Smaller D1 differences are more sensitive to neighbourhood size: the open structure shifts from a slight increase at low $k$ to a decrease at high $k$. Narrower windows resolve finer spatial variation, including sign changes in near-zero differences.

\subsection{Domain-resolved responses in haemagglutinin}
\label{sec:appendix-ha}

During the low-pH transition from HA state III to IV\citep{benton2020structural}, $\mathrm{C}_3$ pLS decreases in HA1 but increases in HA2 and the combined HA1--HA2 region (\cref{edfig:ha}). Structural views reveal outward splaying of the heads and extension of the stem.

A $120^\circ$ cyclic self-match about the fitted threefold axis provides a complementary coordinate-space assessment (\cref{tab:ha-selfmatch}). From state III to IV, the self-match residual increases in HA1 and decreases slightly in HA2, consistent with their respective pLS changes. Within each state, mean pairwise C$\alpha$ RMSD after Kabsch alignment of protomers remains below $0.04$\,\AA{}, indicating closely matched internal conformations across protomers.

\begin{table}[htbp]
\centering
\fontsize{9}{11}\selectfont
\setlength{\tabcolsep}{4pt}
\renewcommand{\arraystretch}{1.15}
\captionsetup{font=normalsize,margin=0pt}
\caption{Domain-resolved self-match residuals in haemagglutinin. Residuals (\AA) use a cyclic $120^\circ$ rotation about the shared protomer-centroid axis.}
\label{tab:ha-selfmatch}
\begin{tabular}{>{\raggedright\arraybackslash}p{\dimexpr0.22\linewidth-2\tabcolsep\relax}>{\centering\arraybackslash}p{\dimexpr0.156\linewidth-2\tabcolsep\relax}>{\centering\arraybackslash}p{\dimexpr0.156\linewidth-2\tabcolsep\relax}>{\centering\arraybackslash}p{\dimexpr0.156\linewidth-2\tabcolsep\relax}>{\centering\arraybackslash}p{\dimexpr0.156\linewidth-2\tabcolsep\relax}>{\centering\arraybackslash}p{\dimexpr0.156\linewidth-2\tabcolsep\relax}}
    \toprule
    Domain & neutral pH & state I & state II & state III & state IV \\
    \midrule
    HA1 head & $0.26$ & $0.26$ & $0.54$ & $0.85$ & $1.05$ \\
    HA2 stem & $0.23$ & $0.22$ & $0.54$ & $0.51$ & $0.49$ \\
    \bottomrule
\end{tabular}
\end{table}

\FloatBarrier

\subsection{Spatial symmetry organization in GluK2}
\label{sec:appendix-gluk2}

GluK2 exhibits a twofold arrangement in ATD and approximately fourfold arrangements in LBD and TMD\citep{meyerson2016structural}. Consistently, $\mathrm{C}_4$ pLS is highest in TMD, followed by LBD and ATD (\cref{edfig:gluk2}c). Across layer thicknesses of $10$--$60$\,\AA, TMD-proximal layers generally exhibit higher pLS than upper extracellular layers, with local variations depending on layer boundaries and thickness (\cref{edfig:gluk2}b; protocol in \cref{sec:appendix-gluk2-protocol}).

\subsection{Icosahedral target agreement in E18 capsids}
\label{sec:appendix-e18}

Against the proper icosahedral group $\mathrm{I}$, degree-six pLS decreases from $6.56$ in the intact activated E18 capsid to $1.65$, $1.15$, and $0.85$ in structures lacking one, two, and three pentamers, respectively (\cref{edfig:e18-capsid}). Each class was reconstructed independently (\cref{sec:appendix-e18-protocol}).

\putbib[Reference]
\end{bibunit}

\end{document}